\documentclass[10pt,twocolumn,letterpaper]{article}
\usepackage[T1]{fontenc}
\usepackage[pagenumbers]{cvpr}
\usepackage{amsmath,booktabs,tikz}
\usetikzlibrary{arrows.meta,positioning}
\graphicspath{{figures/}}
\newcommand{\suppindex}[1]{\hyperref[suppidx:#1]{S#1\,(\pageref*{suppidx:#1})}}
\usepackage[colorlinks,allcolors=blue]{hyperref}
\title{TWS-V: Traceable Visual Evidence\\for Decision-Time Validation}
\author{Zoe Li\textsuperscript{1}\qquad Anze Wang\textsuperscript{2}\\[3pt]
\small \textsuperscript{1}Research Scientist, University of Washington\qquad
\textsuperscript{2}Independent Researcher}
\hypersetup{pdftitle={TWS-V: Traceable Visual Evidence for Decision-Time Validation},pdfauthor={Zoe Li, Anze Wang}}
\begin{document}
\maketitle
\begin{abstract}
Trustworthy visual agents should be able to show why an action was still
justified at the moment it was executed. Yet an agent can click exactly where it
saw a control and still act wrongly: grounding is judged on the frame where it
was computed, but by dispatch time that frame may no longer support the action,
and a fresh re-grounding may not repair it. Checking freshness is not enough: in
an executed GUI counterexample, byte-identical screenshots lead to opposite
native outcomes, so no threshold on those pixels can avoid both wrong actions
and lost completions. TWS-V is a traceable dispatch-time contract that binds
each action to the source, context, target and time interval of its evidence,
keeps unknowns explicit, and validates the cached proposal and any replacement
separately within a bounded repair budget, so that every dispatch decision can
be audited.

We score every policy jointly on completion, invalid actions and resource cost.
On 720 controlled TEST episodes, selective repair completes 119/180 versus
95/180 for calibration-budget-matched random repair, using 104 versus 110
resolver calls. On 300 native-application episodes, a transparent
contract-plus-correlation rule matches the completion of always re-grounding
with one tenth of its re-groundings. In a controlled repair study, validating
replacements eliminates all 18 imposed invalid replacement actions and raises clean
completion from 78/96 to 96/96, at 36.4\% more re-groundings than unchecked
acceptance but 31.8\% fewer than blind extra grounding.

Transparent rules implementing the same predicates match the full trace. In
24 paired offline replays using Black Forest Labs' frozen FLUX 3 Action
model, the contract admits six fresh, correctly bound proposals and
withholds 18 with stale evidence, a substituted chunk, or missing
dispatch time. In a single RoboLab development scene using the same
model, an exterior-view check rejects
a viable replacement and times out after a target shift. After a separate
bowl displacement in that scene, all three refresh policies complete
48 steps sooner than scheduled reuse. A learned adoption head does not
beat the selected simple rule on new natural videos. TWS-V gives builders
of visual agents an auditable basis for deciding when to trust, refresh
or abstain, judged jointly on outcome, error and cost.
\end{abstract}

\section{Introduction}
An agent that acts on pixels acts on the past. A computer-use agent locates a
control, reasons about its task and executes the action only after that
reasoning ends, when the screen may have changed; a robot policy executes one
action chunk while inferring the next. Visual grounding is usually scored
against the frame on which it was computed; whether that localization still
supports the action at dispatch is a separate visual question. Long-horizon
computer-use workflows make this observation-to-action gap
consequential~\cite{osworld}: a click that was correct when computed can toggle
the wrong setting or change the wrong record when executed. The intuitive fix
is to check, at dispatch, whether the supporting screenshot is still current and
to re-ground the action if it is not.

The same gap appears in embodied world-action models: DreamZero executes one
action chunk while asynchronously inferring the next, then replaces generated
visual context with real observations after chunk execution~\cite{dreamzero}.
LingBot-VA likewise overlaps prediction with execution. Its authors report that
a naive asynchronous cache relies on outdated visual predictions and drifts;
their FDM-grounded variant recomputes predictive context from recent real
feedback and the executing action, obtaining 90.4\% versus 74.3\% average
success on RoboTwin~2.0 Easy~\cite{lingbotva}. This grounds the staleness concern
beyond GUI agents, but prediction updates and dispatch-time support are different
questions: TWS-V asks whether the available evidence, observed or predicted,
still supports the queued action when it is dispatched.

\begin{figure*}[t]
\centering
\includegraphics[width=\textwidth]{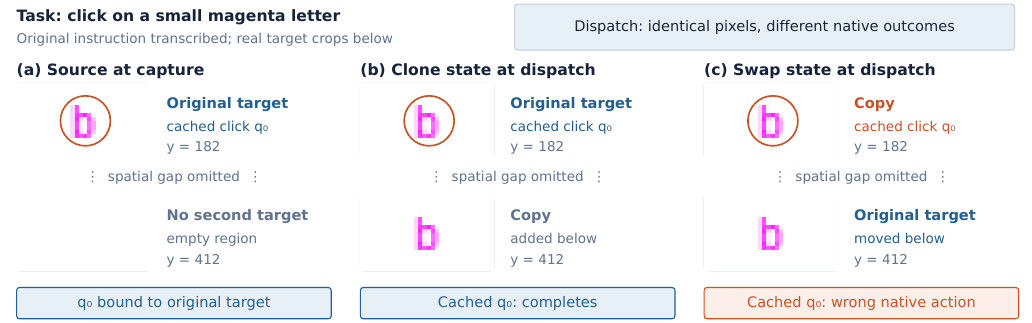}
\caption{\textbf{Identical evidence, different native outcomes.} Enlarged real target crops from the first TEST seed (2080; target b); the vertical gap is omitted and source pixel coordinates are retained. The task instruction is transcribed above. A visually identical copy is added below. In the clone state the original remains at the cached click; in the swap state the original and copy exchange positions. The two states have byte-identical source screenshots and byte-identical current screenshots. Native-state labels describe the imposed condition, not information recoverable from these pixels.}
\label{fig:identifiability}
\end{figure*}

Fig.~\ref{fig:identifiability} shows why this intuitive fix is insufficient,
and it is the observation the rest of the paper builds on. A visually
identical copy of a GUI control is added; in one state the original stays at the
cached location $q_0$, in the other the original and the copy exchange positions. The
two states share byte-identical source screenshots and byte-identical current
screenshots, yet clicking the cached location completes the task in one state and
produces a wrong native action in the other. A pixel-level freshness gate
reuses both proposals: it completes all 20 clone episodes and errs on all 20
swap episodes. Always re-grounding abstains on both. No threshold on these
pixels can do better (Prop.~1); only a new observation or additional context can
separate the two states. Freshness, traceability and semantic action support
are therefore distinct questions, and a checker can only be judged by the joint
outcome it produces: native completion, erroneous actions and resource cost,
with the actor, caches and recovery held fixed.

\begin{figure*}[t]
\centering
\includegraphics[width=\textwidth]{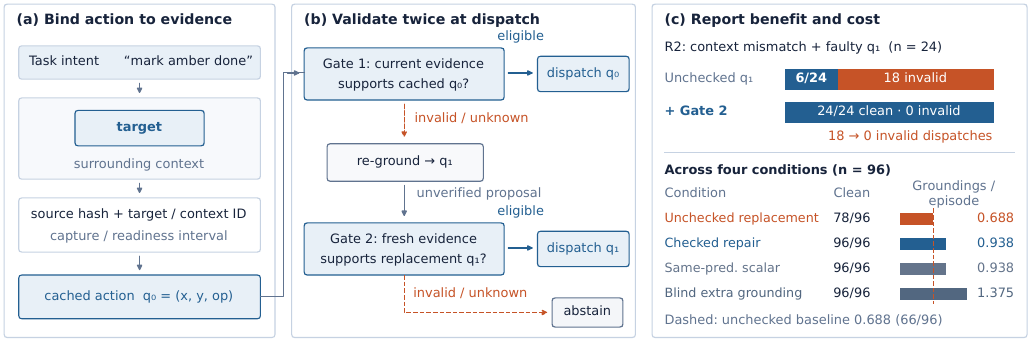}
\caption{\textbf{A complete repair chain validates both the cached proposal and its replacement.} (a) Bind source, target/context and time. (b) Validate cached and replacement proposals separately. (c) Gate 2 prevents all 18 imposed invalid replacement dispatches in R2; checked repair uses 90 rather than 66 re-groundings across 96 episodes. An identical-predicate scalar instantiation reaches the same endpoint. These are controlled mechanism and cost results, not natural error rates or a safety guarantee.}
\label{fig:overview}
\end{figure*}

TWS-V turns this view into a dispatch-time specification
(Fig.~\ref{fig:overview}). Each observation and check is bound to its declared
source, context, target and time interval and is revalidated when the action is
dispatched. Unknown clock mappings or bindings stay unknown rather than
defaulting to valid, and a failed check triggers at most a bounded re-grounding
before abstention. Repair must answer two questions: has the cached proposal
$q_0$ lost support (Gate~1), and has the replacement $q_1$ acquired it (Gate~2)?
Time decay or re-observation alone addresses only the first. The second can
prevent wrong actions but can also reject useful
repairs, so its value can only be judged by measuring completion, errors and
cost together. The specification wraps an existing actor without retraining it
and does not replace grounding; any implementation satisfying its predicates is
an instantiation. With healthy bindings a transparent scalar rule can therefore
suffice, and recording a trace alone should not change behavior. The
specification fixes which checks are made, and our measured gains trace to
those checks, above all validating the replacement, rather than to checker
complexity. Our contributions are:
\begin{itemize}
\item A decision-time source--time--action specification that separates temporal
applicability from semantic validity, with explicit unknowns, dispatch-time
revalidation, separate cached/replacement checks and bounded repair.
\item An executed identifiability counterexample and its consequences: fresh but
indistinguishable evidence forces an error--coverage trade-off that only new
observation or context can break, and it accounts for every erroneous action of
our selective policy.
\item A joint outcome--error--cost evaluation separating replacement-check
benefits from extra repair budget, retaining identical-predicate controls, and
reporting negative learned-adoption results on new natural videos. An offline
FLUX 3 Action integration and matched simulator branches test the distinction
between packet integrity, visual support and task completion.
\end{itemize}
Gains are measured on controlled and adapted native tasks; official OSWorld
actors did not pass our admission gate (Supp.~S28--S30). Sec.~\ref{sec:practice}
turns these results into concrete guidance for teams deploying visual agents.


\section{Related work}
Acting on a condition checked earlier is the classical time-of-check-to-time-of-use
hazard~\cite{toctou}. C2PA verifies assertions under a trust model, not scene
truth~\cite{c2pa}. Age of Information formalizes timeliness~\cite{aoi}, and Age of
Incorrect Information also accounts for correctness~\cite{aoii}; we build on this
distinction rather than claim it. Selective classification and conformal risk
control trade coverage for error under exchangeability~\cite{selective,crc}; our
calibrated cutoffs follow this logic, but adaptive, history-dependent risk needs
stronger assumptions (Prop.~3).

SeeAct separates visual grounding in web interaction~\cite{seeact}. SeeClick~\cite{seeclick},
CogAgent~\cite{cogagent}, OS-Atlas~\cite{osatlas}, Aguvis~\cite{aguvis} and
UI-TARS~\cite{uitars} scale screenshot-based GUI grounding and action, and
Set-of-Mark prompting~\cite{som} supplies explicit visual anchors. T$^2$Mem
addresses conflicting evidence and write timing in memory~\cite{t2mem}.
HeLM maintains explicit language memory for non-Markovian manipulation and
updates it only on detected events~\cite{helm2026}; whether
a stored entry still supports the next command is a separate, dispatch-time
question.
MemoryBench constructs manipulation tasks whose current observations coincide
while histories, and hence required actions, differ; SAM2Act+ resolves them with
a spatial memory bank~\cite{fang2025sam2act}. Prop.~1 states the corresponding
limit for a dispatch-time checker that sees only current pixels.
GroundingPI predicts spatial targets~\cite{groundingpi}; DSDyn-VLA combines
prediction and residual control~\cite{dsdyn}; temporal straightening regularizes
latent trajectories for planning~\cite{temporalstraightening}. GR-1 jointly
trains action and future-image prediction so that video prediction shapes the
visuomotor representation~\cite{gr1}; VPP instead reads predictive intermediate
features from a video diffusion model without decoding a complete future video
for every action~\cite{vpp}. These results motivate predictive features as one
possible support signal, but a predicted future is not a renewed observation of
the dispatch-time scene. TWS-V therefore treats such features as contextual
evidence whose source and readiness must be recorded, not as proof that a queued
action remains valid. Generalist VLAs
such as $\pi_{0.7}$ additionally train across demonstrations, autonomous robot
rollouts, multimodal web data and egocentric human video, and condition behavior
on language, visual subgoals and episode-level performance metadata~\cite{pi07}.
This heterogeneity motivates source-stratified validity evaluation rather than
an assumption that pooled training breadth makes every observation suitable for
dispatch. A generated subgoal is a traceable proposal, not independent proof
that the corresponding action remains valid at dispatch. These methods
improve perception, memory or policy learning. We instead ask whether an existing
proposal still has support at dispatch and measure the resulting repair
trade-off; stronger grounding and a traceable checker are complementary.

Concurrent agent-memory work detects outdated memories in conversations
(STALE~\cite{stale}), studies budgeted verification of inherited memory
(\emph{When Stale Constraints Go Unchecked}~\cite{staleconstraints}),
decays memory activeness with time since observation (HitMem~\cite{hitmem}), or
re-observes suspect regions to correct embodied memory (SPARK~\cite{spark}).
TWS-V instead binds visual evidence to dispatch, validates cached and replacement
proposals separately, and reports outcome, error and cost jointly.

Mind2Web~\cite{mind2web}, WebArena~\cite{webarena}, VisualWebArena~\cite{vwa},
AndroidWorld~\cite{androidworld} and OSWorld~\cite{osworldv1,osworld,osrepo}
provide web, mobile and desktop tasks with executable or native evaluation.
Offline visual benchmarks test a component; they cannot show success after an
intervention changes a trajectory, which is why we execute every policy.

\section{Decision-time evidence contract}
Let $z_t$ contain only evidence legitimately available before dispatch $t$:
observation content, source/context/target bindings, observation and readiness
intervals, the proposed action and the actor's own history. Evaluator state and
later feedback cannot enter $z_t$, and every numerical time comparison requires
a justified common clock. For observation interval $[o^-,o^+]$ and dispatch
interval $[d^-,d^+]$, the conservative age is
\begin{equation}
A_{\max}=d^+-o^-.
\label{eq:age}
\end{equation}
A check with readiness $[a^-,a^+]$ is usable only if $a^+\le d^-$; unknown clock
mappings are never replaced by zero. The validator checks expected bindings,
recomputes received-content hashes, tests availability and age against a
declared budget, and rejects malformed order. Expected hashes require an external
trust origin; a downstream digest does not attest a sensor.

\paragraph{Dispatch and bounded recovery.}
The result is eligible, invalid or unknown with reasons. Eligible means the
implemented contract passes, not that perception is correct. Because evidence
may expire during reasoning, revalidation happens at dispatch. The controller
reuses a supported cached proposal $q_0$ (Gate~1), requests fresh grounding
$q_1$ when Gate~1 fails, and allows at most one further observation/retry before
abstention. A replacement $q_1$ must satisfy its own bindings and support checks
before dispatch (Gate~2); passing Gate~1 or obtaining a newer screenshot cannot
validate that replacement.

The controlled visual check compares the source target region with the same
location in the current screenshot using fixed photometric normalization and
zero-mean normalized correlation (ZNCC). Its cutoff is fixed on valid
calibration actions; the shared resolver checks visible match quality and
competing locations. Exact preprocessing and working points are in the
supplement.

\subsection{Formal boundaries}
\paragraph{Proposition 1: indistinguishable evidence.}
Consider a fixed proposed action in two states with identical available evidence
$z$ but opposite validity. A randomized checker adopts with the same probability
$p(z)$ in both states. Its invalid-adoption probability on the invalid twin and
rejection probability on the valid twin therefore satisfy
\begin{equation}
e_0(z)+r_1(z)=p(z)+1-p(z)=1.
\label{eq:identifiability}
\end{equation}
Always abstaining removes invalid adoption at the cost of all valid adoption on
this pair. More informative observation or justified context can break the
indistinguishability; a new threshold on the same evidence cannot.

\paragraph{Proposition 2: temporal applicability.}
If true times lie within truthful, correctly mapped intervals, then
$o^+\le d^-$, $d^+-o^-\le B$ and $a^+\le d^-$ imply actual age in $[0,B]$ and
readiness before dispatch (by endpoint ordering). This guarantees the temporal
contract only, not semantic correctness or source authenticity.

\paragraph{Proposition 3: horizon exposure.}
For at most $H$ dispatches let $A_t$ mean adopt, $V_t$ action validity, and $F_t$
failure of the contract's factual assumptions. Suppose
$P(F_t)\le\delta_t$ and
$P(A_t=1,V_t=0,F_t^c)\le\epsilon_tP(A_t=1,F_t^c)$. Then
\begin{equation}
P(\exists t\!\le\! H:A_t=1,V_t=0)
\le\sum_{t=1}^{H}\!\left[\delta_t+\epsilon_tP(A_t=1,F_t^c)\right].
\label{eq:horizon}
\end{equation}
The proof splits each error by $F_t$ and applies the union bound; independence
is unnecessary. Marginal calibration error rates do not establish these
adaptive-history premises, so we do not claim a deployment bound
(Supp.~S27 gives the full assumptions).

\section{Evaluation protocol}
We separate three evidence classes: software contract controls, offline visual
validity and executed native task outcomes. An actor must pass a frozen
admission test (visible, absent and ambiguous targets) before entering TEST, and
failed actors are reported. TRAIN (fitting), VAL (model selection), CAL
(threshold calibration) and TEST (held-out evaluation) are disjoint, and old TEST
data are never reused for tuning.

Controlled MiniWoB++~\cite{liu2018miniwob} generators and native handlers are
pinned at commit 33c3b4d, with Playwright replacing the official runner~\cite{miniwob}.
Shared initialization supplies source target regions equally to every policy,
so these are validation/repair tasks rather than autonomous grounding. Native
OpenApps~\cite{openapps} studies use the unchanged native Task/CompositeTask
evaluators on adapted fictional fixtures. Model inputs exclude the DOM,
reference coordinates, native state and task outcomes.

Comparators are cached (no extra check), always re-ground, CAL-budgeted random
or fixed-interval refresh, ZNCC alone, RGB difference and a standalone
contract-AND-ZNCC conjunction. All arms share the actor, prompts, valid caches,
observations and recovery. A \emph{native-negative} action is an executed action
that the application's own evaluator scores as wrong. We report completion,
native-negative actions, abstention, captures, re-groundings, model
calls/tokens and timing jointly, including failed episodes. Paired uncertainty
resamples generator seeds for controlled fixtures and source videos for natural
components.

\section{Results}
\subsection{Selective repair on controlled GUI tasks}
\label{sec:expanded}
Admission accepts the expanded symbol actor (50/50 visible completions, 0/20
absent-target attempts, 10/10 ambiguous abstentions) and rejects the link actor
(46/50, 4/20, 5/10), which does not enter TEST. Twenty CAL seeds (180 cases) fix
the local cutoff and the random repair probability (0.4) before twenty new TEST
seeds; nine imposed conditions give 180 native episodes per policy.

\begin{table}[t]
\centering\scriptsize
\setlength{\tabcolsep}{2pt}
\begin{tabular}{lrrrr}
\toprule
Policy & Complete & Negative & Abstain & Calls\\
\midrule
Cached &100&60&0&0\\
Always re-ground &99&0&81&261\\
Selective (TWS-V gate) &\textbf{119}&20&41&104\\
CAL-budget random &95&40&34&110\\
\bottomrule
\end{tabular}
\caption{Controlled symbol TEST, 180 episodes per policy. Negative: native-negative
actions; abstentions are a subset of non-completions. All arms receive current
screenshots.}
\label{tab:qualified-budget}
\end{table}

\begin{figure*}[t]
\centering
\includegraphics[width=.78\textwidth]{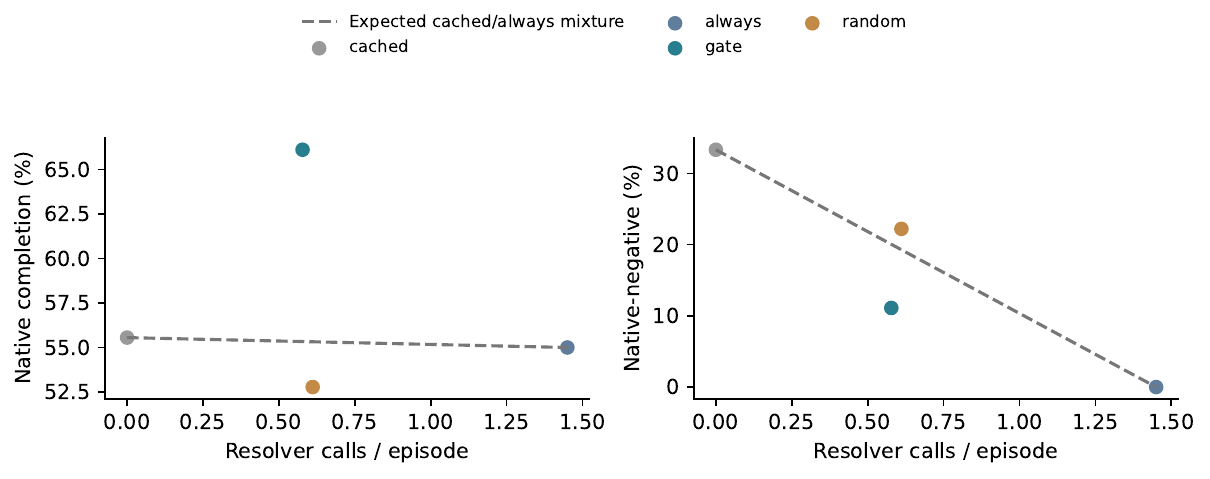}
\caption{\textbf{Outcome--cost trade-off on the controlled TEST.} Points are
executed policies; dashed lines are the analytical expected mixtures of cached
and always re-ground at the same resolver budget (not new runs). Left: native
completion. Right: native-negative actions. Selective repair improves both
completion and errors over CAL-budget-matched random repair at a similar call
budget; always re-grounding makes no native-negative actions, at higher call
cost and lower completion.}
\label{fig:costcurve}
\end{figure*}

Selective repair improves completion over the budget-matched random policy by
13.33 points [10.00, 16.67] and halves native-negative actions ($-$11.11 points
[$-$14.44, $-$7.22]) with a similar number of resolver calls (104 vs.\ 110).
Against always re-grounding it completes 20 more episodes with 2.5$\times$ fewer
calls, but it makes 20 native-negative actions where always re-grounding makes
none (Tab.~\ref{tab:qualified-budget}). Fig.~\ref{fig:costcurve} places each
executed policy against the expected cached/always mixture: the gate lies above
the mixture in completion and below it in errors, while random falls on the
wrong side in both. Fewer resolver calls do not imply fewer screenshot requests
or lower latency (421 vs.\ 414 requests; Supp.~S4).

\subsection{Where the errors come from}
Tab.~\ref{tab:failure} breaks the outcomes down by condition. On the four benign
conditions (static, dark mode, background change, benign style) every policy
completes 20/20. Relocation and target removal are the cases freshness checking
is designed for: the gate detects the change, re-grounds and completes 19/20
relocations, and abstains on all 40 absent-target episodes, whereas random
reuses stale proposals and makes 26 native-negative actions.

\begin{table}[t]
\centering\small
\setlength{\tabcolsep}{3.5pt}
\begin{tabular}{lcccc}
\toprule
Condition & Cached & Always & Gate & Random\\
\midrule
Relocation & 0/20/0 & 19/0/1 & 19/0/1 & 5/14/1\\
Absent (removed) & 0/20/0 & 0/0/20 & 0/0/20 & 0/12/8\\
Absent (covered) & 0/0/0 & 0/0/20 & 0/0/20 & 0/0/9\\
Ambiguity: clone & 20/0/0 & 0/0/20 & 20/0/0 & 10/0/10\\
Ambiguity: swap & 0/20/0 & 0/0/20 & 0/20/0 & 0/14/6\\
\bottomrule
\end{tabular}
\caption{Per-condition TEST outcomes, completion/native-negative/abstention out of
20. The four benign conditions (all 20/0/0) are omitted. Clone and swap have
byte-identical screenshots (Fig.~\ref{fig:identifiability}).}
\label{tab:failure}
\end{table}

The remaining difference lies entirely in the two ambiguity conditions, which
form the indistinguishable pair of Fig.~\ref{fig:identifiability}. The gate sees
identical pixels and reuses the cached click in both: this yields all 20 of its
extra completions over always re-grounding (clone) and all 20 of its
native-negative actions (swap). Always re-grounding abstains on both, paying 20
lost completions for zero errors. This is Prop.~1 executed: both deterministic
policies satisfy $e_0+r_1=1$ on this pair, at opposite ends of the trade-off. The gate's
error is thus not a calibration failure that a better threshold could remove;
reducing it requires evidence beyond the screenshot, such as a handler identity
or a post-action check. All 900 CAL/TEST rows and 1,956 image links are
recounted (Supp.~S4).

\subsection{Native application tasks}
\label{sec:native}
We next run the same comparison inside native OpenApps applications with a
fixed local actor (Qwen3-VL-2B goal parser with OCR re-grounding) and native
evaluators. A first study crosses four goals with three conditions and seven
policies (84 episodes, Supp.~S23). Under an imposed 76-pixel scroll, no-check,
fixed-interval refresh and RGB difference complete 1/4 goals and make unintended
checkbox changes, whereas ZNCC, the conjunction and TWS-V complete 4/4 with
three re-groundings versus ten for always re-grounding.

\begin{table}[t]
\centering\scriptsize
\setlength{\tabcolsep}{1.8pt}
\begin{tabular}{lrrrrr}
\toprule
Policy & Unch. & Scroll & Style & Clean/60 & Repairs\\
\midrule
CAL-freq.\ random & 20 & 5 & 20 & 45 & 3\\
ZNCC & 20 & 20 & 20 & 60 & 15\\
Contract AND ZNCC & 20 & 20 & 20 & 60 & 15\\
TWS-V (full trace) & 20 & 20 & 20 & 60 & 15\\
Always re-ground & 20 & 20 & 20 & 60 & 150\\
\bottomrule
\end{tabular}
\caption{Native OpenApps TEST with a common cap of six extra re-groundings and
sixteen steps: 20 goals $\times$ 3 conditions $\times$ 5 policies = 300 episodes.
Cells give native completions out of 20; Clean also requires no unintended
state change. All policies use 60 model calls and 10,767/1,692 tokens.}
\label{tab:native-cap}
\end{table}

A larger prospective study extends this to twenty new goals, including
two-operation compositions, under unchanged, scrolled and benign-style
conditions, with a common cap on repairs and steps (300 episodes,
Tab.~\ref{tab:native-cap}; Supp.~S31). The full-trace TWS-V instantiation and the
transparent conjunction both complete all 60 goals cleanly with 15 re-groundings,
one tenth of always re-grounding's 150, and capture 305 versus
300 screenshots. The random policy spends only three repairs but fails 15 scrolled
goals, each with unintended state changes. Evaluator-only hit tests confirm the
mechanism: in the scroll condition, fifteen initially invalid checkbox proposals
are re-grounded into valid ones and the corresponding goals complete, while the
60 initially valid proposals are reused.

In both studies TWS-V and the transparent contract-AND-ZNCC conjunction agree
on every paired scenario in outcomes, actions, re-groundings and tokens. This is
expected: with healthy metadata and identical predicates, recording a trace
cannot change behavior (Supp.~S27). The contract matters when bindings fail:
under a benign context-ID mismatch it forces four extra re-groundings without
changing completion, i.e., enforcement is visible but not automatically a
prevented error. Token use is identical across policies because OCR re-grounding
reuses the parsed intention, so savings appear in re-grounding calls and loop
time (346.7s vs.\ 437.5s for always), not tokens. The guard itself costs about
0.16\,ms per call.

A frozen repair-chain stress test then crosses 24 further related goals, four
conditions and eight policies (768 episodes; Supp.~S35). When a context mismatch
triggers re-grounding and the first fresh coordinate $q_1$ is deliberately
offset (R2 in Fig.~\ref{fig:overview}c), native fresh acceptance completes 6/24
episodes and dispatches 18 invalid checkbox actions; validating the replacement
at Gate~2 completes 24/24 with none. Across
all conditions this changes clean completion from 78/96 to 96/96. The same
scalar contract/OCR predicates again match TWS-V exactly. Replacement validation
uses 90 re-groundings versus 66 for native fresh acceptance (+36.4\%); a blind
extra grounding with the same old-evidence trigger also reaches 96/96 but uses
132; it is blind to replacement support, not to the refresh trigger. At the same
observed completion endpoint, replacement validation therefore uses 31.8\% fewer
re-groundings than blind extra grounding. This is controlled mechanism evidence,
not a natural error-rate or unique-framework advantage.

A further frozen target-context study crosses 12 new related goals, five imposed
conditions and nine policies (540 episodes; Supp.~S37). Checked context repair
completes 60/60 versus 51/60 for native replacement acceptance, using 48 versus
36 re-groundings. Every checked policy reaches 60/60, including the older
patch rule and the same-predicate scalar control: richer context supplies no
unique completion gain. Cost axes differ: full context checking uses 48
re-groundings and 51 extra OCR calls, versus 72/39 for trigger-matched blind
extra repair and 117/12 for always-checked repair. Recorded parser tokens are
identical across arms. Consequently, fewer refreshes alone do not establish
lower total computation or token cost. These two repair studies are executed
controlled cohorts; subsequent accounting audits are not new TEST episodes.

\subsection{Embodied action-chunk integration and failure analysis}
\label{sec:embodied-diagnostics}
\begin{figure*}[t]
  \centering
  \includegraphics[width=\textwidth]{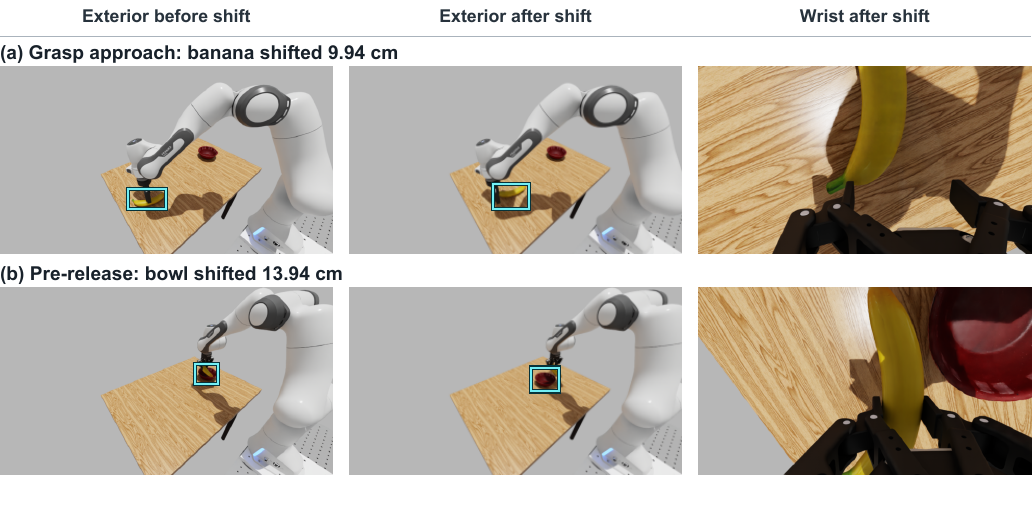}
  \caption{\textbf{Recorded RoboLab observations at two matched intervention
  forks.} (a) The banana moves 9.94\,cm during grasp approach (steps
  88--89). (b) The destination bowl moves 13.94\,cm before release
  (steps 177--178). Each row pairs the first exterior camera before and
  after the intervention with the post-shift wrist view. Both post-shift
  panels in each row are tiles from the FLUX 3 Action input composite.
  Boxes identify the moved objects for the reader. Policy outcomes are
  reported in Tab.~\ref{tab:robolab-main}.}
  \label{fig:robolab-same-state}
\end{figure*}

\paragraph{Offline dispatch bindings.}
We apply the contract to six 32-step proposals from a frozen FLUX 3
Action checkpoint~\cite{fluxaction}, generated from two Cosmos3-DROID
recordings~\cite{cosmosdroid}. In 24 paired replays, the gate admits all
six fresh, correctly bound proposals, blocks all 12 with an injected age
or chunk-binding fault, and withholds six as unknown when dispatch time
is missing. The bypass admits all 24. The unit of source replication
is the recording. This test checks timing and proposal provenance
under controlled faults (\suppindex{41}).

\paragraph{Matched closed-loop development cases.}
We use the same frozen FLUX 3 Action actor in the RoboLab~\cite{robolab}
banana-in-bowl task port. At each fork, all four policies share the same
pre-intervention scene state, camera views and cached action queue.
Figure~\ref{fig:robolab-same-state} shows the exterior and wrist evidence.
At the grasp-approach fork, shifting the banana by 9.94\,cm drops
correlation at the cached location from 0.940 to 0.087. The three refresh
policies receive the same replacement. TWS-V's exterior-template
recheck rejects it even though both refresh controls execute it and
complete (Tab.~\ref{tab:robolab-main}). The resulting hold times out.
A wrist-view check added after inspecting this failure admits the
same proposal in a repeat of the development scene.

\begin{table}[t]
\centering\small
\setlength{\tabcolsep}{3pt}
\begin{tabular}{@{}lrr@{}}
\toprule
Policy & \shortstack{Banana shift\\9.94\,cm} & \shortstack{Bowl shift\\13.94\,cm} \\
\midrule
Scheduled reuse & 196 & 267 \\
Unconditional refresh & 224 & 219 \\
Time + vision & 224 & 219 \\
TWS-V & 1{,}050 (timeout) & 219 \\
\bottomrule
\end{tabular}
\caption{RoboLab completion steps or timeout from episode start at two
matched intervention forks of one development seed.}
\label{tab:robolab-main}
\end{table}

At the pre-release fork, shifting the destination bowl by 13.94\,cm
leads all four policies to complete. All three refresh policies dispatch
identical actions and finish 48 steps sooner than scheduled reuse
(\suppindex{42}).

\subsection{Natural video components}
We test whether candidate re-grounding transfers to natural video. On
237 source-family CDTB~\cite{cdtbrecord} TEST pairs, a learned linear check does
not significantly improve on RGB difference (detection difference 0
[$-$0.156, 0.200]; Supp.~S16--S22). A new SAM3~\cite{sam3} study uses six
separately qualified videos and disjoint TRAIN/VAL/CAL/TEST source videos
(59/24/37/66 videos; 491 TEST cases). Correct proposals exist: the top-five pool
covers 141/152 unique visible TEST targets and top-1 is correct for 119/152.

\begin{table}[t]
\centering\small
\setlength{\tabcolsep}{3pt}
\begin{tabular}{lrrr}
\toprule
Rule & Correct & Unobs.\ error & Multiple error\\
\midrule
CAL product &48.05&2.55&8.48\\
Selected simple &52.30&2.55&9.38\\
Score linear &54.61&6.46&9.38\\
Selected MLP &50.35&5.78&8.48\\
\bottomrule
\end{tabular}
\caption{New-video SAM3 TEST: correct unique adoption and adoption errors (\%),
macro-averaged over videos. Rules/models selected in VAL, cutoffs in CAL.}
\label{tab:recovery-main}
\end{table}

Adoption, however, is the bottleneck. The VAL-selected MLP does not meet the
prespecified joint criterion against the selected simple rule: correct adoption
changes by $-$1.95 points [$-$13.54, 7.99] while unobserved-target error rises by
3.23 [0.47, 7.27] (Tab.~\ref{tab:recovery-main}; operating-point curves in
Supp.~S26). Candidate supply is therefore not adoption value, and the same joint
criterion that credits the gate in Sec.~\ref{sec:expanded} rejects this head.

A Human-1M natural-video mini-validation adds head-pose and hand-contact
cues to a pixel rule. The two policies make identical decisions in all
54 evaluable TEST rows. An author spot-check agrees with the AI labels
on cached-point validity in only 16/30 cases ($\kappa=.29$), so the
recorded rates are diagnostic and cannot establish natural-video
performance (\suppindex{40}).

\section{Implications for deployed agents}
\label{sec:practice}
Any agent whose perception and execution are separated by reasoning, queuing or
network delay faces this gap. Our results translate into four practices; each
rests on the evidence above and none assumes a particular model or actor.

\paragraph{Log evidence bindings with every action.} Recording the source,
context, target and observation/dispatch intervals of each action's evidence
costs about 0.16\,ms per guard call and does not change behavior when bindings
are healthy (Sec.~\ref{sec:native}). It buys an audit trail: after an incident,
an operator can tell whether an action was dispatched on evidence that was
stale, unbound or never checked, and an unknown clock mapping appears as
unknown rather than silently valid.

\paragraph{Check the repair, not only the staleness.} Accepting the first fresh
coordinate $q_1$ after a refresh trigger dispatched 18 imposed invalid checkbox
actions in our repair stress test; validating the replacement (Gate~2) removed
all of them, at 36.4\%
more re-groundings than unchecked acceptance and 31.8\% fewer than blind extra
grounding (Sec.~\ref{sec:native}). For actions that change application state,
a second check on the replacement used the fewest re-groundings among the
error-free policies we measured (tied with its scalar equivalent), although it
adds OCR calls that blind extra grounding does not (Supp.~S35).

\paragraph{Spend re-groundings where evidence changed, and name the cost axis.}
In native applications a transparent contract-plus-correlation rule matched
always re-grounding with 15 instead of 150 re-groundings and 346.7\,s instead of
437.5\,s of loop time, but with identical tokens and nearly the same screenshots
(305 vs.\ 300). Savings appeared in grounding calls and latency, not in model
tokens; a cost claim is only meaningful with its axis named.

\paragraph{Decide the residual trade-off explicitly.} When current pixels cannot
separate a valid from an invalid action, every checker trades wrong actions
against lost completions (Prop.~1): our gate made 20 such errors that always
re-grounding avoided only by abstaining. Which side to take depends on what a
wrong action costs in the application. Reducing both requires evidence beyond
the screenshot, such as handler identity or a post-action check. Before
deployment, a checker should be compared with budget-matched random refresh and
always re-grounding on the same actor and caches, reporting completion, wrong
actions, abstentions and every cost axis together.

\section{Limitations}
Our executed gains come from controlled tasks with supplied targets and from
adapted native applications; they do not yet establish natural long-horizon
transfer. On official OSWorld tasks, the 2B actor fails admission and a 4B actor
completes none of three tasks under any controller (Supp.~S28--S30), so the
intervention could not be tested there. We show no behavioral advantage of the
full specification over an identical-predicate scalar rule. Incomplete binding
checks can produce different decisions, but tracing alone cannot. Independent
closed-loop benefit remains to be tested with a qualified actor on natural
workflows against strong controls at prespecified error and cost limits
(Supp.~S36).
The repair studies' imposed faults identify mechanisms, not field failure
prevalence. Re-groundings, OCR, parser tokens and measured loop time are distinct
cost axes; empirical timing summaries do not certify deployment latency.
The robot replay measures packet-level contract behavior. The closed-loop
robot branches share one development seed, and the wrist-view recovery
was designed after inspecting that scene. The matched development cases
give no independent estimate of robot-task improvement.
Natural-video labels group by source
video rather than certified independent sites, and the SAM3 cohort is
underpowered for tight noninferiority. A signed false timestamp stays false, and
hardware attestation and physical-safety guarantees are outside this work.

\section{Conclusion}
Seeing a target and acting on it are separated in time, and freshness,
traceability and semantic action support are different questions. An executed
counterexample shows that byte-identical evidence can support opposite
outcomes, and it accounts exactly for the errors of an otherwise effective
selective repair rule. TWS-V makes these boundaries and their binding
assumptions explicit at dispatch. A selective instantiation improves completion
over a budget-matched baseline on controlled tasks, and a transparent
contract-plus-correlation instantiation reaches full native completion with a
tenth of the re-groundings of always refreshing. The central design lesson is to
validate both the cached proposal and its replacement: the second check improves
controlled outcomes over unchecked repair at additional refresh cost, and
matches blind extra repair with fewer re-groundings. Richer bookkeeping adds
auditable provenance and explicit unknowns, not a behavioral advantage over a
scalar rule implementing the same predicates, and neither comparison
establishes unique method superiority. In robot simulation, the same
separation makes a false replacement rejection visible, while the
destination-shift case isolates refresh timing from replacement checking.
The next step is evidence beyond the screenshot, such as action effects or handler identity, tested with the same
joint outcome--error--cost protocol on independent natural workflows that
include binding failures. Wearable platforms such as Aria Gen 2 already record
gaze, hand pose and metric head trajectories~\cite{kong2025aria2pilot}, but
their perception outputs do not state whether a stored observation still
supports a later action, which makes egocentric memory a natural next test. As
visual agents move from benchmarks into products, the question to ask of every
action is not only what the agent saw, but whether that evidence still holds
when it acts.

\section*{LLM Usage Statement}
LLM-based tools (OpenAI ChatGPT/Codex and Anthropic Claude) were used under the authors' direction for
implementation, experimentation support, figure preparation, writing and reviewer-style feedback.
The authors determined the research design and conclusions, verified all results, and take full responsibility for this preprint.

\clearpage
\section*{Appendix: Supplementary Evidence and Reproducibility}
\addcontentsline{toc}{section}{Appendix: Supplementary Evidence and Reproducibility}
\setcounter{figure}{0}
\setcounter{table}{0}
\setcounter{equation}{0}
\renewcommand{\thefigure}{S\arabic{figure}}
\renewcommand{\thetable}{S\arabic{table}}
\renewcommand{\theequation}{S\arabic{equation}}
\renewcommand{\theHfigure}{app.\arabic{figure}}
\renewcommand{\theHtable}{app.\arabic{table}}
\renewcommand{\theHequation}{app.\arabic{equation}}
\section*{Reader's guide}
\textbf{Key controlled results.} Selective repair completed 119/180 tasks
versus 95/180 for budget-matched random repair (\suppindex{4}). Checking
replacements increased clean completions from 78/96 to 96/96
(\suppindex{35}). In R3, checked adaptive repair and the same-predicate
scalar control each reached 60/60, with different refresh and OCR costs
(\suppindex{37}). See the \hyperref[tab:claim-evidence-map]{claim--evidence map}
and the \hyperref[tab:supp-index]{full section index} for direct links. In the
robot development case (\suppindex{42}), TWS-V withheld a replacement and
timed out at 1,050 steps while the other three branches succeeded.

\paragraph{Audit and scope conventions.}
Saved-output recounts verify recorded inputs, labels, decisions, and outcomes
within the scope stated in each section. The supplement reports new inference
and reruns explicitly. It retains failed qualification and unevaluable outcomes
with their original denominators. Repeated frames and policy branches share
their parent episode or video. Controlled and development results support
claims within their respective tasks, perturbations, and comparators.
Physical safety, natural prevalence, and token savings require direct
measurements. Selected asset, provenance, and process receipts are collected in
the \hyperref[tab:reproducibility-manifest]{reproducibility manifest}.

\begin{table*}[t]
\centering
\begingroup
\scriptsize
\setlength{\tabcolsep}{4pt}
\begin{tabular}{@{}p{.22\textwidth}p{.34\textwidth}p{.37\textwidth}@{}}
\toprule
Claim & Direct evidence & Boundary of the claim\\
\midrule
Selective repair improves the controlled operating point
& 119/180 completions versus 95/180 for CAL-budget-matched random repair;
104 versus 110 resolver calls; native-negative actions are halved
& Controlled TEST with supplied source targets; not natural GUI transfer,
token saving, or a safety result\\
Replacement evidence requires its own validation
& Checked repair changes 78/96 to 96/96 clean completions and prevents all
18 imposed invalid replacement dispatches
& Identical-predicate scalar checks reach the same endpoint; imposed faults
identify mechanism, not natural prevalence\\
Pixel-only evidence has an identifiability boundary
& Clone/swap states have byte-identical source and current screenshots but
opposite native outcomes; Prop.~1 gives the resulting error--rejection trade-off
& Does not imply that every visual error is unidentifiable; new observation or
justified context can separate the states\\
Natural-video component transfer is not established
& The SAM3 rule misses its joint criterion (\suppindex{26}); the published-source
MLP trails the simple rule by 28.52 points (\suppindex{38}); Human-1M binding ties
the pixel rule and its AI labels fail two author spot-checks (\suppindex{40})
& Human-1M error rates are diagnostic only; neither study establishes natural
end-to-end completion\\
Independent closed-loop benefit remains open
& Official OSWorld/Arena actors fail prespecified admission or completion gates
& No independent natural-workflow, token-saving, or physical-safety claim\\
\bottomrule
\end{tabular}
\caption{Claim--evidence--boundary map. Positive, tied, negative and
non-evaluable results remain separate; no row upgrades a component test into a
closed-loop or safety claim.}
\label{tab:claim-evidence-map}
\endgroup
\end{table*}

\begin{table*}[t]
\centering
\scriptsize
\renewcommand{\arraystretch}{0.96}
\begin{tabular}{@{}lll@{}}
\toprule
Section (page) & Evidence and linked claim & Outcome or scope \\
\midrule
\multicolumn{3}{@{}l}{\textbf{Core controlled and theoretical evidence}}\\
\suppindex{4} & Selective repair on a controlled symbol task & Gain over matched random repair \\
\suppindex{14} & Controlled native dispatch & Intervention pilot \\
\suppindex{15} & Native GUI instance and composition transfer & Prospective exploratory test \\
\suppindex{23} & Native joint controls and repair costs & Controlled outcomes and costs \\
\suppindex{31} & Native repair under a common cap & Expanded controlled study \\
\suppindex{35} & Cached applicability and replacement support & Checked replacement gains; scalar control ties \\
\suppindex{37} & R3 context repair with stronger controls & Completion ties; costs differ \\
\suppindex{5} & Contract and visual ablation & Separate endpoints \\
\suppindex{24} & Evidence-to-claim alignment & Scope of the reported results \\
\suppindex{27} & Conditional guarantees and identifiability & Assumptions and limits \\
\suppindex{36} & Independent closed-loop criterion & Evaluation conditions \\
\addlinespace
\multicolumn{3}{@{}l}{\textbf{Natural video and embodied evaluation}}\\
\suppindex{16} & Aligned natural component transfer & Primary point unevaluable \\
\suppindex{18} & Independent bounded-cache validity & Increment unestablished \\
\suppindex{20} & Joint operating points and video sensitivity & Paired operating-point analysis \\
\suppindex{21} & Prospective source provenance & Cohort eligibility accounting \\
\suppindex{22} & CDTB calibration transfer & Frozen transfer assessment \\
\suppindex{26} & Constrained SAM3 adoption on new videos & Prespecified criterion unmet \\
\suppindex{38} & Published-source candidate adoption & Frozen MLP trails simple rule by 28.52 points \\
\suppindex{39} & Frozen embodied-policy admission & Prespecified futility stop \\
\suppindex{40} & Human-1M component mini-validation & Binding tie and label reliability limits \\
\suppindex{41} & Offline FLUX action dispatch & Controlled contract replay \\
\suppindex{42} & RoboLab same-state closed-loop simulation & TWS-V timeout and separate refresh case \\
\addlinespace
\multicolumn{3}{@{}l}{\textbf{Development history and archived diagnostics}}\\
\suppindex{1} & Tracker qualification and validity replay & Development gate unmet \\
\suppindex{2} & Fixed actor, learned checker and refresh cost & Development controls \\
\suppindex{3} & Held-state spatial checks & Spatial diagnostic \\
\suppindex{6} & Adverse RGB replay & Heuristic gain unestablished \\
\suppindex{7} & Frozen linear and MLP visual heads & Head and ambiguity controls \\
\suppindex{8} & Retained-box transfer & Source-qualification probe \\
\suppindex{9} & Prospective DAVIS source qualification & Natural component development \\
\suppindex{10} & Visible and unobserved target qualification & Qualification gate unmet \\
\suppindex{11} & Stronger visual front end under strict prefixes & Full qualification gate unmet \\
\suppindex{12} & Fixed-front-end validity heads & Component comparison \\
\suppindex{13} & Native application integration & Bounded admission \\
\suppindex{17} & Calibration-only diagnosis & Source feasibility \\
\suppindex{19} & Cross-scene score diagnosis & Calibration only \\
\suppindex{25} & SAM3 candidate supply and feasibility & Exploratory qualification \\
\suppindex{28} & OSWorld task integration & Actor admission failed \\
\suppindex{29} & Three official native tasks & All 21 native outcomes scored zero \\
\suppindex{30} & Harness dispatch and authored workflows & Boundary and development checks \\
\suppindex{32} & Dense cached-evidence feasibility & All-reject and repair diagnostics \\
\suppindex{33} & Executor and image-input qualification & Failed or ineffective contrasts \\
\suppindex{34} & Fresh-proposal validity and complete repair & Development mechanism checks \\
\bottomrule
\end{tabular}
\caption{Evidence-ordered index. Each section identifier links to its start;
the compiled page is shown in parentheses.}
\label{tab:supp-index}
\end{table*}

\section*{S1. Retained causal-tracker qualification and replay diagnostics}
\label{suppidx:1}
\begin{figure*}[t]
\centering
\includegraphics[width=.98\textwidth]{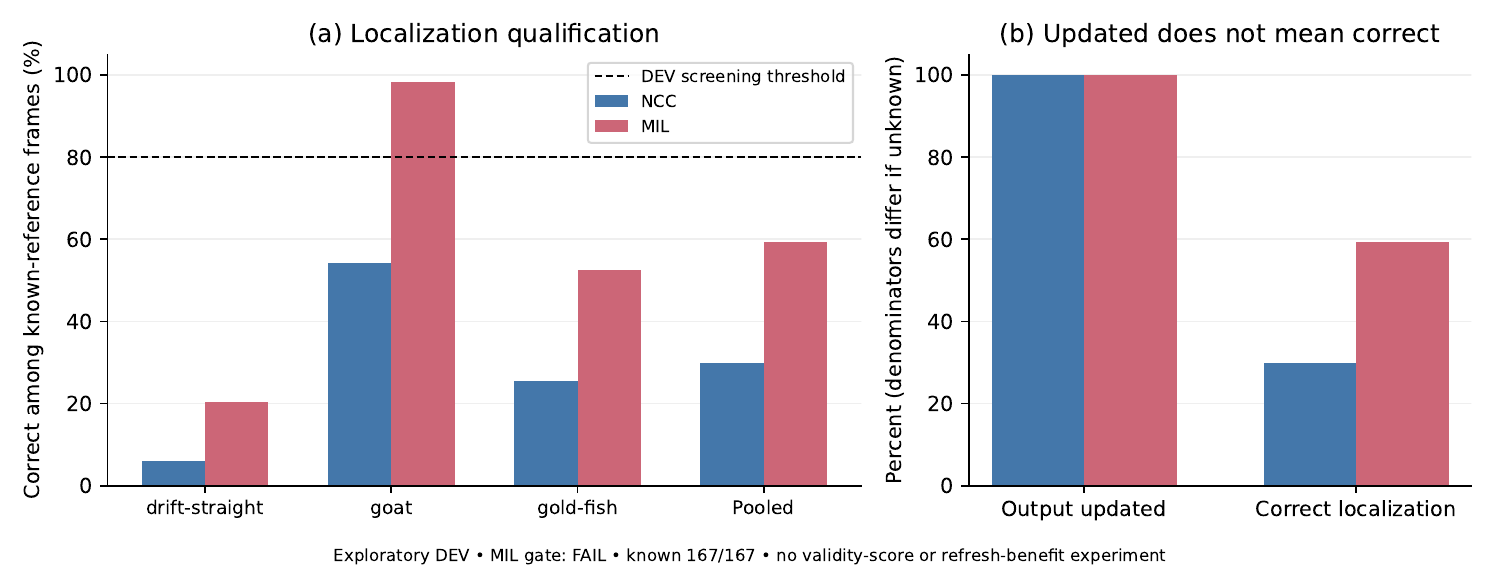}
\caption{Exploratory causal-tracker qualification. MIL improves on the weak
NCC reference but fails the frozen development gate; an updated output need
not be a correct localization. This run establishes neither validity-filter
efficacy nor refresh benefit.}
\end{figure*}
This exploratory negative control preceded the link/symbol intervention.

\paragraph{Adaptive tracker qualification on a third DEV slice.}
Five further alphabetically selected clips supply 282 frames, with 167 decisions
on three EVAL-DEV clips. A default OpenCV MIL tracker~\cite{opencv_mil}, initialized from the same
first-frame box and updated causally, improves localization from NCC's 50/167
to 99/167 correct boxes (IoU at least .5). It improves in all three clips but
fails the prespecified 80\% development screening gate. MIL reports successful
updates on every evaluated frame despite 68/167 invalid boxes: output recency
and API success do not establish spatial correctness. The gate is an engineering
choice, not a safety standard. No validity score is tuned on these outcomes.
MIL learns online; no offline model or external weights are trained or downloaded.
Synthetic checks exposed in-process nondeterminism despite RNG resetting; fresh
per-sequence processes restored the tested repeatability before new data fetch.
All 167 real EVAL estimates subsequently matched in fresh-process replay.

\paragraph{Fresh-clip flow follow-up.}
On the next five alphabetically selected clips (280 frames), fixed NCC and a
causal LK/partial-affine prototype were compared under locked DEV rules. Current
localization error was 77/177 for NCC and 89/177 for flow: flow did not qualify as
a stronger baseline. Its local-score macro AUROC rose from .445 to .502, but
only two of three evaluation clips contain both classes; no robust method claim
follows. Confidence alone remains a stronger pooled reference for both trackers.
Evaluator-only current/cache validity cells for flow are 85 both valid, 3 current
valid/cache invalid, 11 current invalid/cache valid, and 78 both invalid. Thus an
already-computed newer estimate can be worse than retained state in this setup.
These are descriptive replay labels, not an extra-observation or safety benefit.
Four synthetic flow qualification tests and an independent recount support the
implementation; no model weights were trained or selected on new outcomes.

\begin{figure*}[t]
\centering
\includegraphics[width=\textwidth]{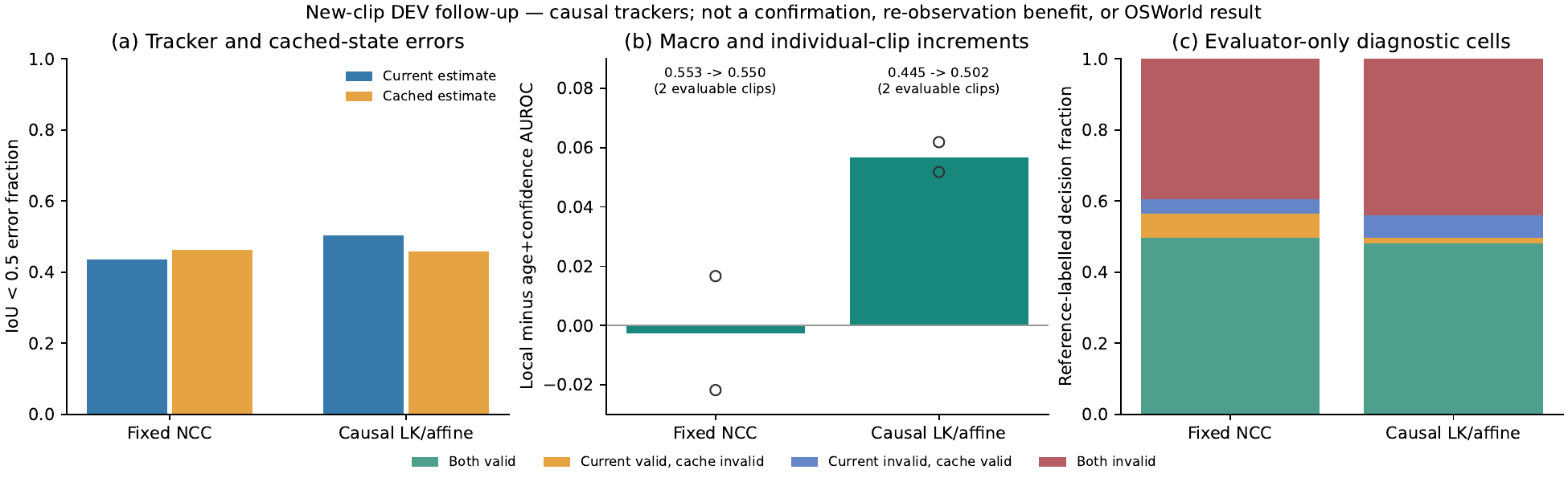}
\caption{Fresh-clip DEV follow-up. (a) Current-estimate accuracy must be qualified
before calling a tracker stronger. (b) Local increments are small and only two
clips have both classes. (c) Evaluator-only validity cells separate potential cache
updates from perception failure. Recency alone does not establish applicability;
the cells do not demonstrate an intervention benefit.}
\end{figure*}

\paragraph{Fixed-lag reference diagnostic (post hoc).}
On the same 120 current frames (40 per EVAL-DEV clip, indices at least 20),
retaining an earlier reference box produces 0/120, 0/120, 7/120, 29/120 and
52/120 invalid boxes at lags 0, 1, 5, 10 and 20 frames. Annotations are strictly
evaluator-only. The reference arm tests change from an idealized correct source
box, not a deployable tracker or measured temporal budget. Cows has no invalid
reference boxes at these lags, preserving the old-but-correct control.
Estimated-box errors are nonmonotone: NCC gives 63/120 errors at lag 0 versus
56/120 at lag 20; flow gives 77/120 versus 43/120. An initially incorrect box can
happen to overlap a later target. Thus reference-box error is not a guaranteed
lower bound, and differences cannot be interpreted as an additive causal
decomposition. No lag was optimized or converted to milliseconds; no intervention
gain is established. These are diagnostics on already examined DEV clips.

\begin{figure*}[t]
\centering
\includegraphics[width=\textwidth]{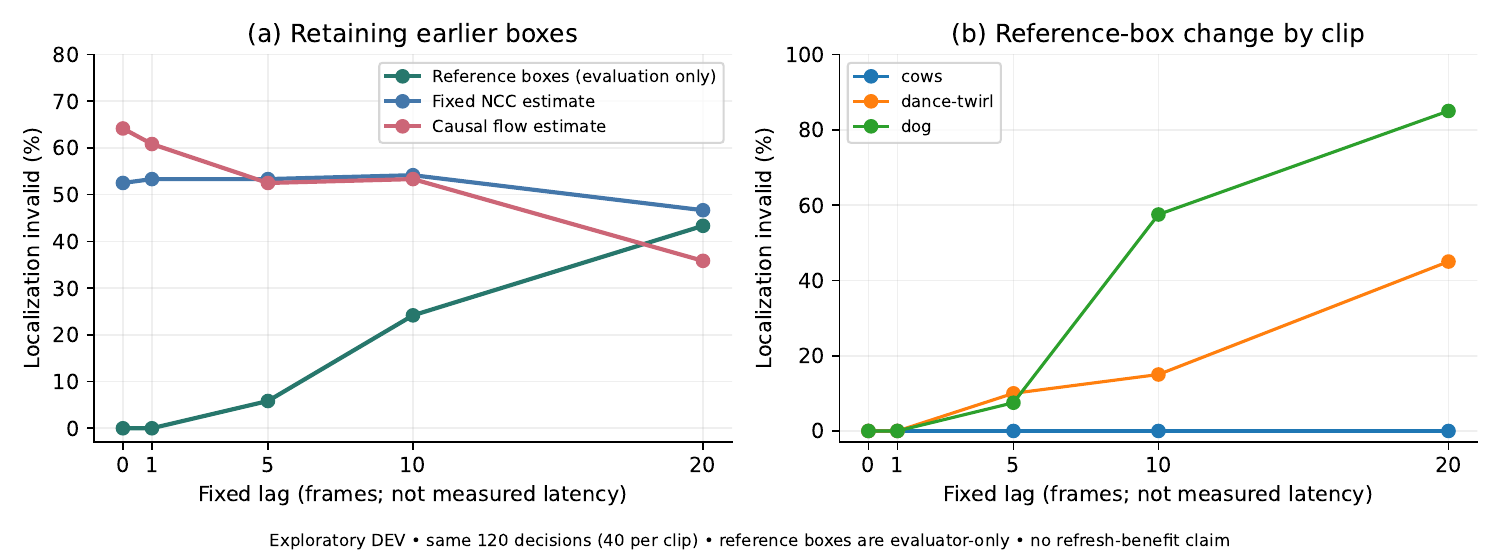}
\caption{Evaluator-only fixed-lag diagnostic, with the same 120 decisions in
every arm. (a) Retaining formerly correct boxes and retaining inaccurate visual
estimates answer different questions. (b) Reference-box change varies by clip.
Frame lag is not measured latency. No refresh, rejection or safety gain is shown.}
\end{figure*}

\paragraph{Pretrained causal baseline (reused DEV).}
A pinned OpenCV Zoo VitTrack ONNX model~\cite{vittrack} improves current localization
to 348/521 correct boxes, versus NCC's 231/521, on all nine previously examined
EVAL-DEV clips. Only four clips reach 80\%, below the prewritten requirement of
six, and pooled correctness is 66.8\%; the qualification gate therefore fails.
The plan's expected count of 511 was an arithmetic error (177+177+167=521),
corrected in a separate retained corrigendum without changing clips or criteria.
All 521 API updates succeed, despite 173 invalid boxes. Model-backed repeatability
and causal-prefix checks complement six adapter tests. Data reuse and unverified
pretraining-data overlap preclude an independent confirmation claim. No score
or refresh-policy search follows the failed qualification; a more capable model
alone is not a TWS-V efficacy result.

\section*{S2. Earlier fixed-actor and calibration controls}
\label{suppidx:2}
\paragraph{Executed colour-action policies (fresh DEV).}
An earlier fixed RGB/contour actor, supplied initial target boxes, used 70 CAL
and 420 TEST episodes. Always re-localizing and gated reuse each complete
100/140 tasks, versus cached execution's 40/140. Gating reduces re-localizer
calls from 200 to 160 but raises median policy process CPU time from 12.32 to
14.00 ms. Both abstain in 40 cases and use 60 extra observations; all 20 transient
occlusions recover, while persistent occlusion and a partially painted relocation
defeat the actor. Imposed conditions and identical per-seed outcomes limit
generalization; an observed tie is not equivalence. Zero LLM calls establish
neither token savings nor whole-stack efficiency. No RL/PPO policy is trained.

\paragraph{Learned-checker control (exploratory).}
Seed-disjoint TRAIN/VAL/CAL sets of 100/50/50 actions fit three CPU logistic
checks before 200 fresh-seed actions. At CAL-fixed cutoffs, RGB and learned
visual checks each reject 101/111 invalid actions, but falsely reject 0/89 and
4/89 valid actions respectively. Adding receipt age rejects 100/111 with 3/89
false rejections; age alone detects none. Observed FPRs are not matched.
Learning shows no increment over fixed RGB here; supplied boxes, imposed faults
and shadow decisions preclude natural-task or intervention claims.

A frozen follow-up on 280 new-seed actions exposes shift dependence. Dimming
preserves all 40 native actions, but RGB falsely rejects 34/40 versus learned
visual 0/40. Under blur, both fail (40/40 versus 39/40 false rejections). Overall
they detect 102/115 invalid actions at 74/165 versus 39/165 false rejections;
these are not matched-FPR comparisons or natural failure rates.

\begin{figure*}[t]
\centering
\includegraphics[width=.98\textwidth]{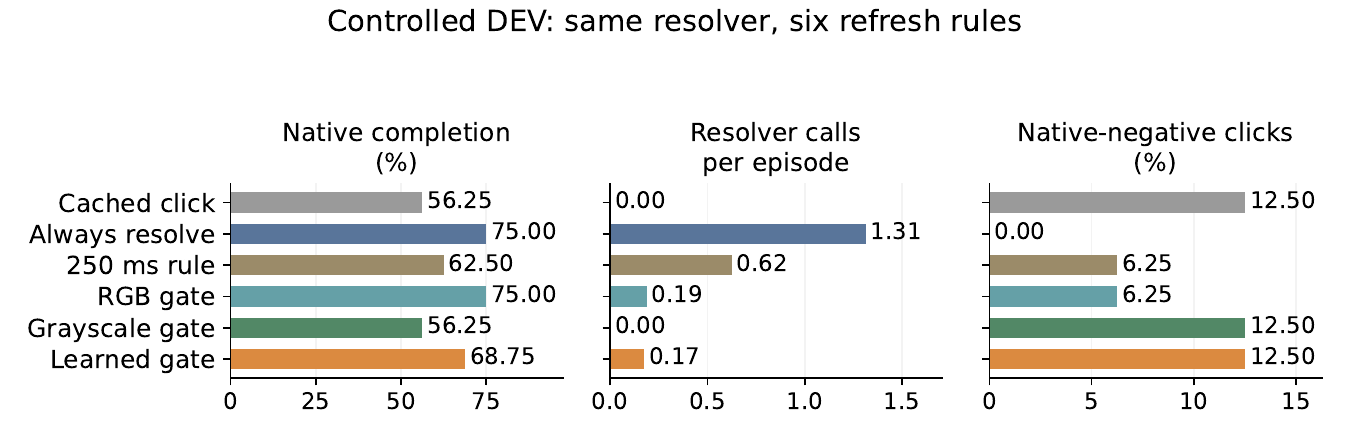}
\caption{Fresh six-rule fixed-actor DEV (160 CAL, 960 TEST): completion, resolver cost and native-negative
clicks must be considered jointly. RGB gating matches always re-localizing's
completion count with fewer calls, but makes ten native-negative clicks versus
zero. Supplied initialization and imposed changes limit scope. All policies
receive a current screenshot; no screenshot or token savings are measured.}
\label{fig:refresh-policy-dev}
\end{figure*}

\paragraph{Refresh costs with nuisance-aware calibration (fresh DEV).}
An additional 160 CAL and 960 TEST episodes compare six rules with the same
colour actor, eight conditions and two observation waits. Thresholds are locked
before ten new TEST seeds; dimming and blur are included in CAL. RGB gating and
always re-localizing each complete 120/160 tasks, using 30 versus 210 resolver
calls, but incur 10 versus zero native-negative clicks. The learned gate completes
110/160 with 28 calls; it does not improve over RGB. Cached, 250-ms and grayscale
rules complete 90, 100 and 90 of 160. Every paired initial image matches exactly.
In dim scenes, cached clicks complete 20/20 while always re-localizing abstains
20/20: re-localizer capability matters. Equal completion counts conceal different
action outcomes (Fig.~\ref{fig:refresh-policy-dev}). These are imposed fixed-actor
trials, not general-agent or safety gains.

\paragraph{Repair capability changes apparent gate value (exploratory).}
A fixed global exposure wrapper qualifies on 100 prior CAL cases (100 correct
versus 82 for raw RGB), then is tested with the unchanged gate on 450 new
episodes: ten seeds, nine conditions, five policies. Always-resolve completion
rises from 50/90 to 60/90; gated completion remains 75/90, narrowing its apparent
advantage from 27.8 to 16.7 percentage points. Both gated variants incur five
native-negative outcomes versus zero for always-resolve. Compensation fixes
moderate global dimming, but fails local and excessive dimming; these cases
remain included. One of 90 compensated-always image pairs differs; excluding
that entire seed preserves the interaction. The interaction is identical across
seeds under imposed conditions, limiting population inference. An independent
OCR screen reads only 7/10 TRAIN crops exactly, below its fixed 8/10 gate;
no text-task intervention follows.

\paragraph{Random refresh at a comparable expected budget (exploratory).}
A separately frozen 270-episode follow-up uses ten new seeds and the same nine
conditions. Random refresh probability is fixed at .125 from prior invocation
counts, not tuned on new outcomes. With the same compensated actor, visual gating
completes 76/90 versus random's 60/90, using 16 versus 18 actual resolver calls
and incurring four versus eight native-negative outcomes. The paired completion
difference is +17.8 points [14.4, 21.1] over ten seed clusters. Always resolving
completes 61/90 with 119 calls and zero native-negative outcomes. One random/always
image pair differs; excluding that whole seed retains a +17.3-point completion
contrast. Actual budgets are not exactly matched; one random realization and
imposed conditions limit inference. Thus selective repair can beat random repair
here, while completion alone still cannot decide the deployment trade-off.

\section*{S3. Held-state spatial diagnostics}
\label{suppidx:3}
A proprietary scene was used as a read-only reference. Our rigid target,
support and occluder are additions. The first streaming capture contains 180 reads
and 120 distinct native reference-time keys, not independently identified exposures.
A separate, fixed four-render held-pose protocol qualifies 16/16 positions within
5 mm; first reads qualify 0/16 and 15/15 with a predecessor match the previous pose.
This is a buffering/content diagnostic, not a moving-frame or physical exposure
synchronization result. All later rules share the four-pass capture baseline.

Three CAL and six TEST seeded simulator sessions yield 405/810 position decisions,
not independent real-world trials. CAL-fixed local age rejects 30/117 positions
with retained error above 20 mm, but also 30/693 valid positions (4.33\% false
rejection); global checks reject none, so FPR is not matched. A fixed local 100 ms
rule rejects all 117 invalid positions and 147 valid ones. Refresh-only accepts
current target measurements and abstains on 117 invalid plus 183 valid positions;
unknown is not a correct rejection. A motion-times-age score rejects 18 invalid
positions, all in one session where apparent pre-occlusion velocity arises while
the reference target is static. We retain this negative diagnostic rather than
claim robust motion information. A numeric-resolution amendment was frozen before
replacement collection and before reading session metrics; the unamended run is
archived, not used to discredit the clock baseline. These held-state tests establish
costs and failure conditions, not task-intervention or physical-safety benefits.

\begin{figure*}[t]
\centering
\includegraphics[width=.98\textwidth]{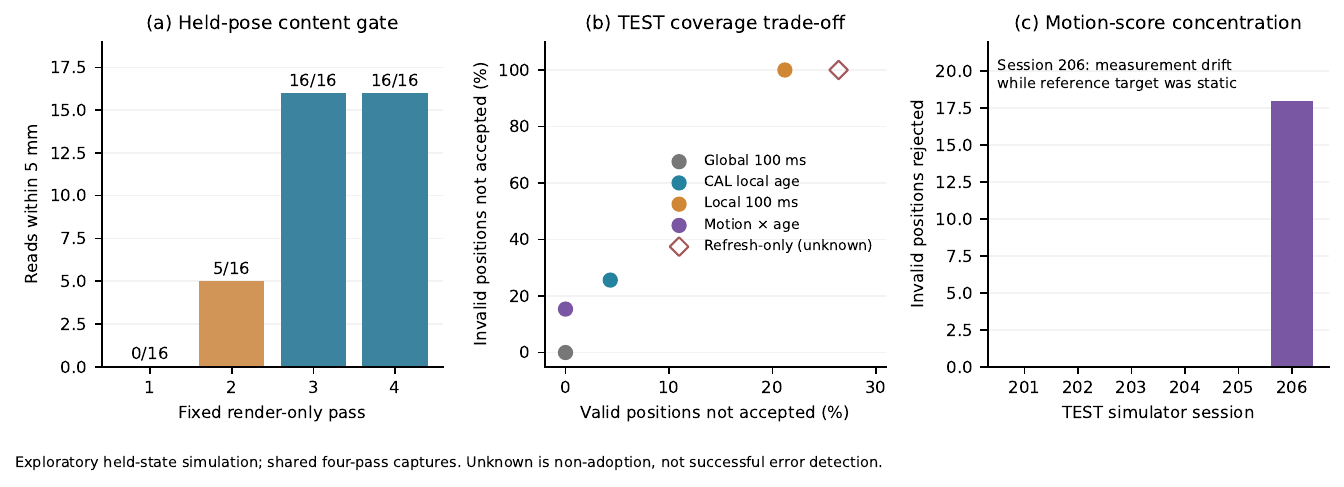}
\caption{Exploratory Half Cubic diagnostics. (a) Fixed-pass held-pose content
correspondence. (b) TEST non-adoption trade-off: refresh-only unknowns are shown
as abstention, not detection. (c) All motion-score rejections concentrate in one
session with stationary-reference measurement drift. All rules share capture
budget; no actions are controlled and no proprietary scene imagery is included.}
\label{fig:half-cubic-session}
\end{figure*}

\section*{S4. Expanded-qualified symbol intervention}
\label{suppidx:4}
The guarded symbol actor alone passes the expanded qualification; the link actor
is retained as failed. New seeds yield 180 CAL and 720 TEST episodes. The local
cutoff and random probability (.4) are frozen after CAL. Every policy uses the
same actor and receives current screenshots; source ROIs are supplied. All
changes, including inert-clone controls, are imposed. No natural-task, source
authentication, physical-safety or token-saving result is inferred.

\begin{table*}[t]
\centering\small
\begin{tabular}{lrrrr}
\toprule
Condition & Cached & Always & Gate & Random\\
\midrule
static & 20/0/0 & 20/0/0 & 20/0/0 & 20/0/0 \\
dark & 20/0/0 & 20/0/0 & 20/0/0 & 20/0/0 \\
background & 20/0/0 & 20/0/0 & 20/0/0 & 20/0/0 \\
relocation & 0/20/0 & 19/0/1 & 19/0/1 & 5/14/1 \\
benign\_style & 20/0/0 & 20/0/0 & 20/0/0 & 20/0/0 \\
absent\_remove & 0/20/0 & 0/0/20 & 0/0/20 & 0/12/8 \\
absent\_cover & 0/0/0 & 0/0/20 & 0/0/20 & 0/0/9 \\
ambiguity\_clone & 20/0/0 & 0/0/20 & 20/0/0 & 10/0/10 \\
ambiguity\_swap & 0/20/0 & 0/0/20 & 0/20/0 & 0/14/6 \\
\bottomrule
\end{tabular}
\caption{Each cell: completion/native-negative/abstention out of twenty TEST
trials. Abstention is a subset of no-completion; not every non-completion is
native-negative. Random p=.4 matches CAL-expected repair cost, not exact TEST cost.}
\end{table*}

\begin{figure*}[t]
\centering
\includegraphics[width=.98\textwidth]{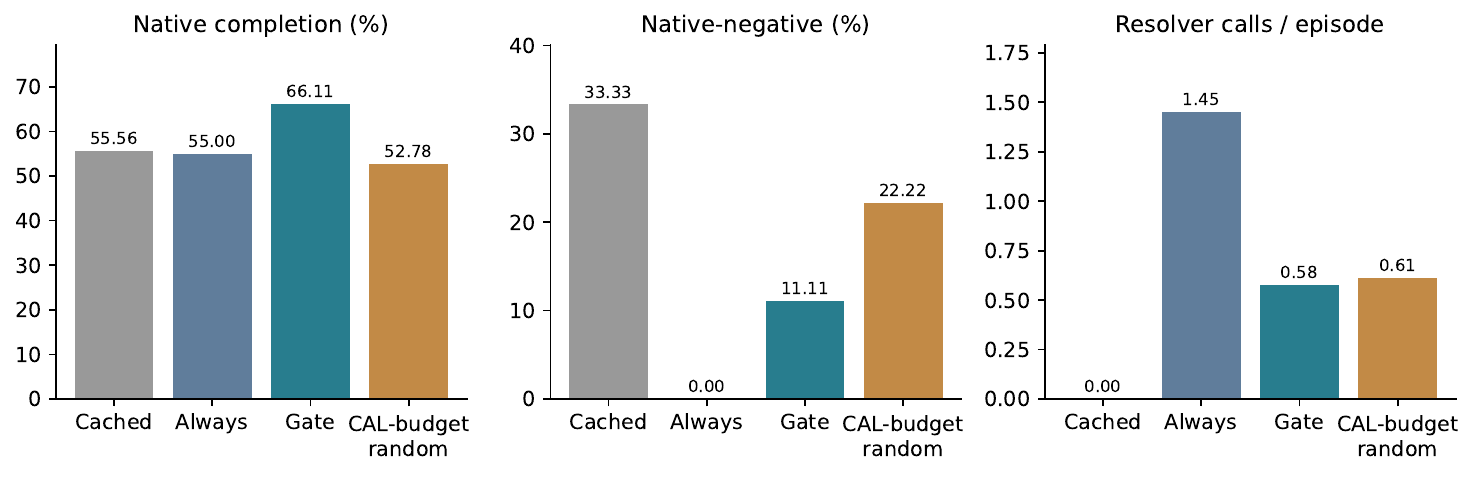}
\caption{All four policies on the new qualified-symbol TEST. Completion, native
negative outcomes and actual resolver invocations must be considered jointly.}
\end{figure*}

\begin{figure*}[t]
\centering
\includegraphics[width=.96\textwidth]{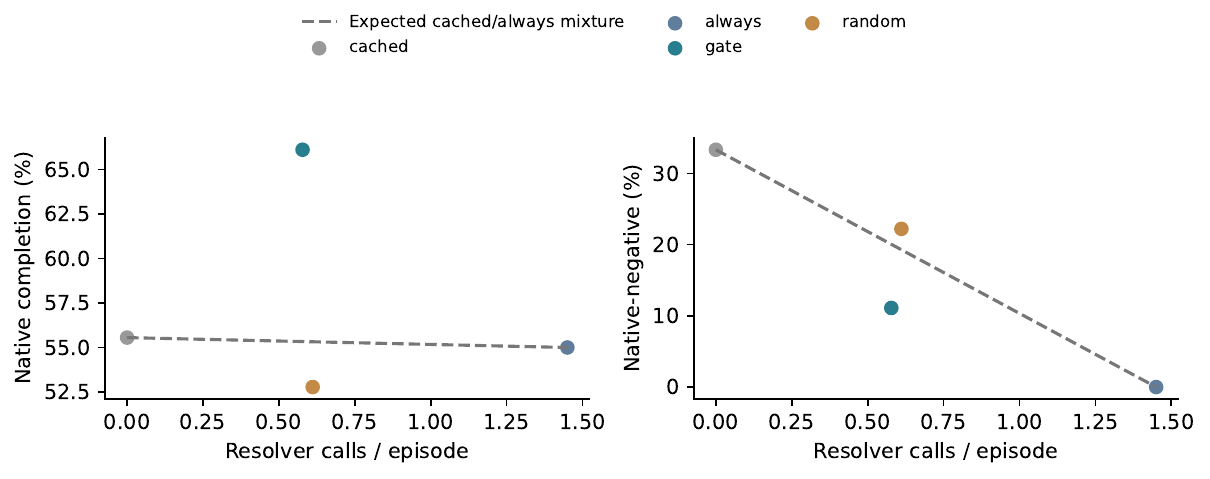}
\caption{Dashed curves are analytical expected cached/always mixtures, not new
executions or TEST-selected probabilities. Points are actually executed policies.
Paired image identity supports, but does not establish general deployment
exchangeability. No latency medians are interpolated.}
\end{figure*}

\paragraph{Uncertainty.}
Gate-minus-random completion is +13.33 points [10.00,16.67], native-negative
outcomes -11.11 [-14.44,-7.22], and calls -0.033 per episode [-0.167,+0.094].
Whole-seed bootstrap uses twenty clusters and 2,000 paired draws; intervals are
exploratory/unadjusted with one random realization. Gate-minus-always completion
and native-negative differences are each +11.11 points on every seed, producing
degenerate intervals that do not establish population certainty.

\paragraph{Capture accounting.}
The handler-swap fixture takes a preparatory clone screenshot before its final
current image, equally across policies. Frozen-code-path counts of requested
screenshots are 380/461/421/414 for cached/always/gate/random, including retries.
This post-run accounting clarification does not modify the run or its outcomes.
Host elapsed time includes that capture, imposed waits and PNG logging; browser
initialization is excluded. These API request counts are not camera exposures.
Gate has more screenshot requests and a larger median elapsed time than random,
despite fewer resolver calls. No whole-stack efficiency claim follows.

The first-seed real-image illustration appears in the main paper.

\section*{S5. Contract and visual ablation are different endpoints}
\label{suppidx:5}
One image pair per fresh qualification case is crossed with seven software
metadata/time variants and four rules. The resulting 4,480 evaluations share
160 cases; they are not independent trials. Source PNG hashes and all visual
scores are recomputed. The original task CAL cutoffs transfer unchanged;
no new semantic calibration guarantee is inferred.

Under healthy metadata, age/contract each admit 60 reference-invalid old points;
visual/full each admit one and withhold seven of 100 reference-valid points.
Injected binding faults are enforced by contract checks, including when the
old action remains reference-valid. Unknown clocks/readiness remain unknown.
The virtual timestamps and trusted manifest are declared software fixtures,
not measured source time or hardware identity. Old-point membership labels
are evaluator-only and not executed-click outcomes in this diagnostic.

\begin{figure*}[t]
\centering
\includegraphics[width=.98\textwidth]{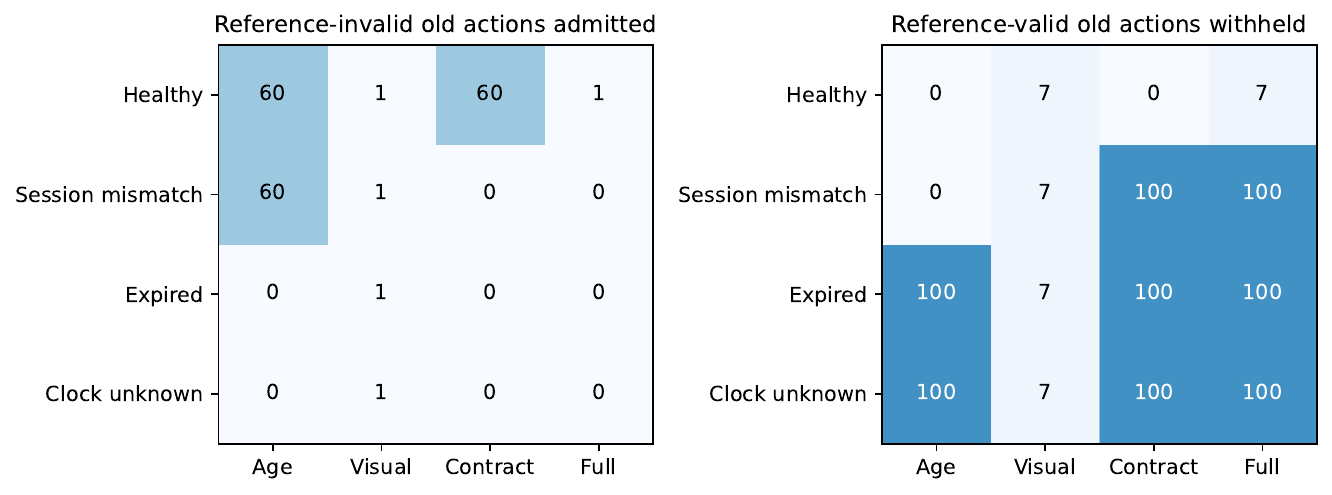}
\caption{Expected contract enforcement and semantic-reference screening must not
be conflated. With healthy metadata, full and visual-only coincide. Contract
withholding under injected faults can also withhold still-valid old actions.}
\end{figure*}

\section*{S6. Initial adverse RGB replay}
\label{suppidx:6}
\paragraph{Auxiliary RGB DEV pilot.}
Five alphabetically selected DAVIS validation clips~\cite{davis}, capped at 60
frames each, supply 290 images. The first two are CAL-DEV and the next three
EVAL-DEV; none constitute prospective confirmation. A fixed first-frame NCC
template generates causal tracks, with a shared five-frame cache refresh.
Later masks are evaluator-only. Cached-box IoU below .5 defines a localization
error, not physical pose error. Seven additional software checks validate bounded
retrieval, metric calculations and invariance to future-frame changes.

Among 177 EVAL-DEV decisions, 91 cached boxes are invalid and 86 valid. The local
heuristic improves pooled AUROC from .552 to .591 over age+confidence, but reduces
sequence-macro AUROC from .587 to .569 and is lower within every evaluated clip.
Confidence alone has pooled AUROC .649. These descriptive results do not establish
a consistent visual increment. CAL-fixed thresholds yield actual EVAL false
rejection 0 versus .023, so the extra local detections are not a matched-FPR gain.
No independent-frame confidence interval or closed-loop benefit is claimed.

\begin{figure*}[t]
\centering
\includegraphics[width=.96\textwidth]{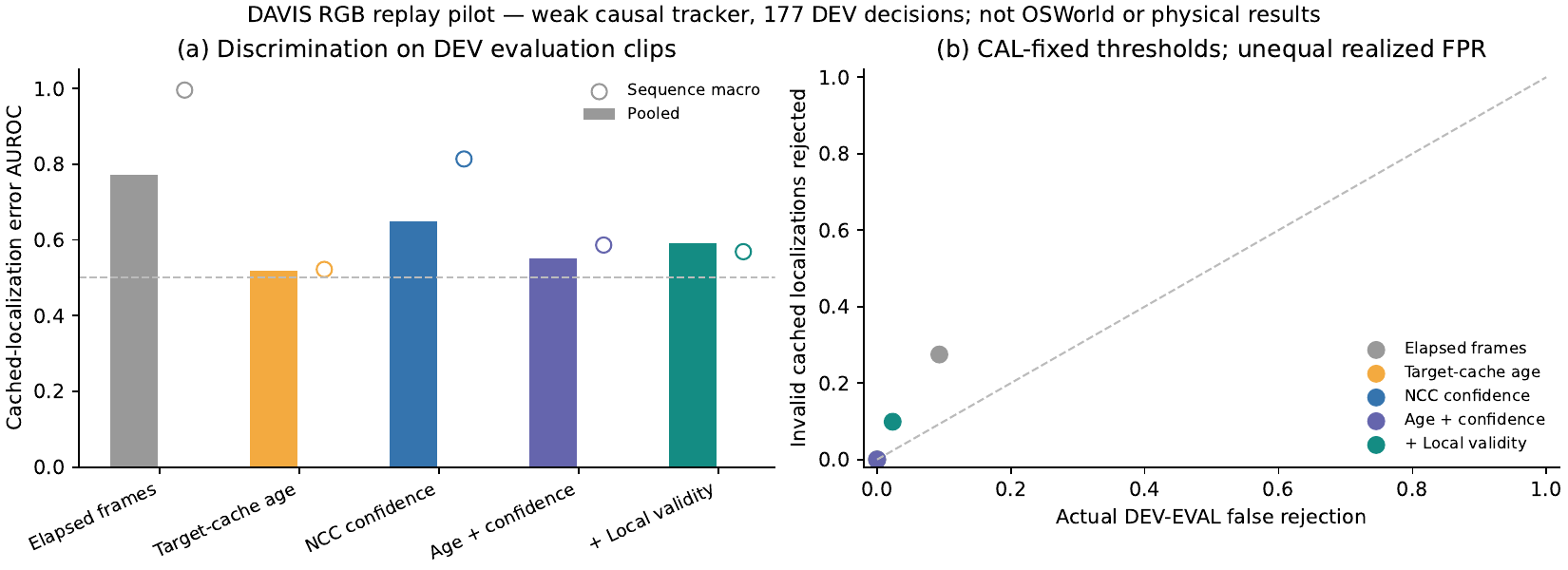}
\caption{Exploratory RGB replay with a weak tracker, not the completed TWS-V
method. Pooled and sequence-macro discrimination differ; CAL-fixed thresholds
produce unequal realized false rejection. The heuristic has no established gain.}
\end{figure*}

\section*{S7. Frozen visual features: head comparison and its boundary}
\label{suppidx:7}
This new exploratory experiment fixes the representation while changing only
the classifier head. It uses 1,340 fresh controlled MiniWoB cases: 560 TRAIN,
140 VAL, 280 CAL and 360 TEST. All conditions and both task families for a
generator seed share a split. The two families appeared in earlier development;
new seeds do not establish unseen-task or natural-workflow generalization.
All 2,680 PNG hashes and native/evaluator label agreements were recounted;
no exact source/current/box duplicate crosses splits. Recounting is not an
independent native rerun.

\paragraph{Representation and frozen choices.}
Cached ImageNet ResNet-18 weights~\cite{resnet} remain frozen. Four 512-D
embeddings represent whole source/current images and their crops at the
historical source box. Absolute differences and normalized historical box
coordinates give 3,076 features, standardized on TRAIN only. Neither current
oracle target coordinates nor task identity, condition, seed, file name or
native label enters the head. Linear logistic and MLP(64, ReLU, 1) heads use
Adam (learning rate .001, batch 128), two L2 choices and at most 100 epochs.
VAL binary cross-entropy selects checkpoints. Three initialization seeds are
all reported; seed 11 is designated before results. Task-specific CAL valid-action
95th percentiles fix rejection thresholds; TEST is never used to retune them.

\begin{table}[ht]
\centering\small
\begin{tabular}{lcc}
\toprule
Method & Links D/FR & Symbols D/FR\\
\midrule
NCC &60/60; 4/80&60/60; 4/80\\
RGB mean difference &20/60; 4/80&20/60; 2/80\\
ResNet + linear &60/60; 6/80&60/60; 5/80\\
ResNet + MLP 64 &60/60; 6/80&60/60; 5/80\\
\bottomrule
\end{tabular}
\caption{Primary initialization 11, core TEST. Detection uses invalid cached
actions; false rejection (FR) uses valid actions. Each task has 60 invalid and 80
valid cases under an imposed condition mix, not natural prevalence. Actual FR
is unequal despite the same calibration rule.}
\end{table}

\paragraph{Saturation is not a neural gain.}
Both neural heads have AUROC 1.000 and TPR 1.000 on both core tasks in all three
initializations; NCC also attains 1.000. Primary linear and MLP each adopt 149/280
core actions, with no observed adopted failure; NCC adopts 152/280 with none.
The three linear initializations have link FR 7.50--11.25\% and symbol
FR 5.00--8.75\%; MLP ranges are 5.00--7.50\% and 3.75--8.75\%.
Primary paired macro TPR difference is 0.000[0,0] and FR difference
0.000[$-.03125,.03125$], from 2,000 shared seed-cluster resamples. The degenerate
TPR interval reflects complete sample saturation, not equivalence or population
certainty. No preferred initialization is chosen using TEST.

\begin{figure}[h]
\centering
\includegraphics[width=\linewidth]{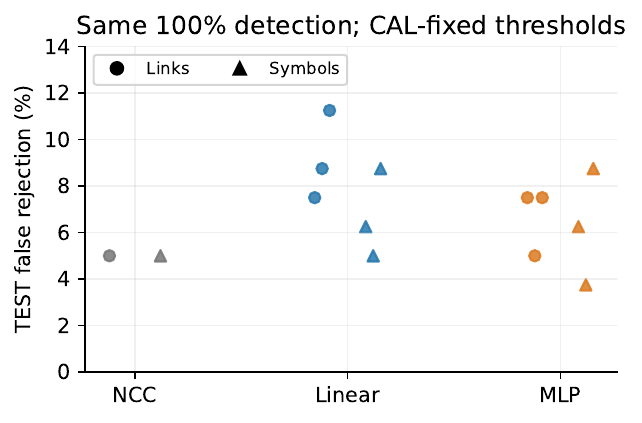}
\caption{Actual false rejection at CAL-fixed thresholds. All three neural
initializations are shown with slight horizontal offsets; NCC is fixed.
Every displayed method detects all 60 core invalid actions per task. Equal
calibration rules do not impose equal TEST false rejection. No learned
intervention benefit is established.}
\end{figure}

\paragraph{Ambiguity remains outside the core result.}
An additional 80 TEST cases form 20 clone/swap pairs per task, with opposite native
cached-action outcomes. Only the 20 symbol pairs have byte-identical source and
current images; link pairs differ in flow layout and are not an identical-pixel
control. Both primary heads adopt the same 8/20 symbol clones and 8/20 swaps,
thus rejecting 12 valid clones while admitting 8 invalid swaps. NCC admits all 20
of each. Tiny GPU batch-dependent score differences (at most $5.19\times 10^{-5}$
linear, $1.89\times 10^{-5}$ MLP) do not distinguish hidden state; paired primary
adoption decisions coincide. No threshold is tuned to exploit numerical differences.

\paragraph{Resources and scope.}
Feature extraction takes 8.72\,s for 1,340 cases on an RTX 5080 Laptop GPU. Linear
and MLP heads have 3,077 and 196,993 parameters, respectively; fitting both
regularizers for one initialization takes approximately 0.45--0.80\,s. These timings
exclude browser collection and are neither complete online decision latency nor
whole-system CPU, energy or token costs. Native intervention and natural-scene
efficacy remain separate, unestablished outcomes.

\section*{S8. Natural retained-box transfer probe}
\label{suppidx:8}
Five new, preselected DAVIS sequences provide 236 decisions. Horsejump-high and
india supply 108 CAL decisions; judo, kite-surf and lab-coat supply 128 EVAL
decisions. The frozen GUI heads retain frame 0's smallest-ID object box; later
masks supply evaluator-only IoU labels. This is neither native action validity
nor robot pose error. The initial lab-coat target is only $4\times 5$ pixels
(9 mask pixels) at $500\times 500$ resolution, contributing 46/58 pooled invalid
retentions. It is retained rather than removed after outcomes, exposing a
missing prospective source-target qualification gate.

Pooled AUROC is 0.565 for linear, 0.935 for MLP, 0.981 for NCC and RGB, and 0.697 for elapsed frame
index. CAL-fixed detections/58 invalids and false rejections/70 valid retentions
are 0/0, 37/1, 20/0, 27/0 and 2/0, respectively. The linear CAL cutoff saturates at 1.0;
the frozen strict tie rule rejects nothing and is not changed using EVAL.
MLP's extra detections therefore are not an established matched-FPR gain.
Judo is entirely valid and lab-coat entirely invalid, leaving only kite-surf
eligible for within-sequence AUROC: MLP 0.968 versus NCC, RGB and elapsed frame
index 1.000. No target becomes reference-absent. All 482 archive files and reference
counts were recounted, but this small, source-qualification-limited probe does
not establish natural generalization; frames are not independent replications.

\section*{S9. Prospectively source-qualified natural follow-up}
\label{suppidx:9}
A follow-up plan fixes source qualification before fetching ten disjoint DAVIS
sequences (alphabetical validation offset 20, first 60 frames). At frame 0, the
first positive object ID with box sides $\ge24$ px, mask area $\ge256$ px and
mask/box occupancy $\ge.10$ initializes a retained box. All ten targets pass;
none is removed or replaced. These are operational choices, not validated
localizability limits. The source mask is an initialization reference, not an
autonomous locator. Later masks only label retained-box IoU $<.5$.
The frozen GUI heads and normalization are unchanged; no new training or model
selection occurs. The first five sequences supply 254 CAL decisions, including
96 valid retentions; the last five supply 258 EVAL decisions, 42 valid and 216 invalid.
CAL fixes 95th-percentile cutoffs, with the same strict tie rule.

\begin{table}[ht]
\centering\small
\begin{tabular}{lrrr}
\toprule
EVAL sequence & Valid & Invalid & AUC eligible\\
\midrule
parkour &3&56&No\\
pigs &20&39&Yes\\
scooter-black &2&40&No\\
shooting &9&30&Yes\\
soapbox &8&51&Yes\\
\bottomrule
\end{tabular}
\caption{All five EVAL sequences remain reported. Per-sequence AUROC requires
at least 5 valid and 5 invalid retentions; only three qualify. No reference-absent
target occurs, so natural absence handling remains untested.}
\end{table}

\begin{table}[ht]
\centering\small
\begin{tabular}{@{}lrrrr@{}}
\toprule
Method & AUC & Macro & Detect/216 & FR/42\\
\midrule
Linear &.849&.830&0&0\\
MLP &.866&.894&50&0\\
NCC &.763&.831&162&18\\
RGB &.868&.846&102&2\\
Elapsed frames &.931&.981&73&0\\
Global fresh &.500&.500&0&0\\
\bottomrule
\end{tabular}
\caption{Descriptive pooled and eligible-clip macro AUROC, and actual EVAL
counts at CAL-fixed cutoffs. Unequal false rejection is not a matched-FPR gain;
no population interval is estimated from five clips.}
\end{table}

The elapsed-frame reference detects 73 invalid retentions versus MLP's 50, both
with 0/42 observed false rejections; this finite sample does not establish equal
population FPR or general superiority. Fixed-source retention can drift
monotonically, making elapsed time a necessary reference. The adopted-invalid
fraction is 166/208 (79.8\%) for MLP and 143/185 (77.3\%) for elapsed frames,
versus 216/258 (83.7\%) without rejection. NCC detects more but falsely rejects
18/42 valid retentions (42.9\%) despite the CAL target of 5\%. Linear's cutoff
again saturates at 1.0; it is not adjusted using EVAL. All source targets pass,
so differences from S8 cannot be attributed causally to qualification.

The bounded transfer downloads 71.2 MB and supplies 522 images and 522 masks.
Three eligibility boundary tests pass. An initial missing-code dependency fails
before prediction; after packaging is corrected, the unchanged runner finishes
in 40.33 s using two low-priority CPU threads. This batch time is not online
latency or CPU utilization. A separate recount verifies all 1,044 file hashes,
source eligibility, boxes, labels, thresholds, counts and per-clip AUROC; it is
not a second model or native-action run. This exploratory diagnostic supports
neither natural-agent efficacy nor a learned increment beyond simple references.
No model-API efficacy comparison is established.

\section*{S10. Natural visible/unobserved locator qualification}
\label{suppidx:10}
The next exploratory plan is fixed before acquisition: first 12 alphabetically
sorted sequences of the official DAVIS2017 train split, first 100 frame indices,
stride 5 including frame 0. All eligible source-frame objects enter, using the
same minimum size/occupancy rule as S9. These sequences are disjoint from all
five previous acquisition manifests. The 186 RGB images and 186 masks yield
22 supplied source objects and 340 object/frame cases. Three methods produce
1,020 decisions on the same cases, not independent episodes. This is DEV
qualification, not model training, official TEST or an efficacy estimate.

The fixed base searches the entire current image using grayscale zero-mean
normalized template correlation, fixed source size and peak threshold .80.
A single guarded candidate requires .90 and abstains if a separate non-overlapping
peak is at least .90 and within .05 of the best. There is no current-reference input,
appearance update, threshold search, neural training or model API call.
A retained source box supplies a simple reference. The source box is supplied
from annotation, not autonomously grounded.

A nonempty current instance mask counts as visible, including partial occlusion;
IoU$\ge.5$ defines a correct candidate. An empty mask without void pixels is
annotation-unobserved, not proof that the physical object ceases to exist.
Void label 255 is not an object; an empty target mask with any void would remain
reference-unknown. No such unknown occurs here. Frame 0 identity checks are
separate from later decisions. All methods see identical RGB pairs.

\begin{table}[ht]
\centering\small
\begin{tabular}{lrrr}
\toprule
Method & Source/22 & Visible/301 & Unobs./17\\
\midrule
ZNCC base &22&8&0\\
Guarded &22&1&0\\
Retained box &22&118&17\\
\bottomrule
\end{tabular}
\caption{Correct identity/visible localizations versus erroneous candidate
attempts on unobserved targets. The base and guard abstain on 293/301 and
300/301 visible cases, respectively; unknown output count is zero. Reporting
only their zero unobserved attempts would hide near-total visible refusal.}
\end{table}

The base and guard correctly localize only 2.66\% and .33\% of visible later
cases. Neither passes the fixed 80\% visible gate overall or in eight sequences.
All 22 source/current identity cases pass, as do five synthetic boundary tests;
these do not qualify moving-target localization. The source-box reference is
correct on 39.20\% of visible cases but always emits a box when the target is
unobserved. No actor proceeds to a new policy intervention.

The 17 unobserved cases occur only in bmx-bumps (10) and color-run (7), below
the fixed support requirement of 20 cases across three sequences. Zero attempts
therefore does not qualify natural absence handling, independently of the visible
failure. All 12 sequences remain reported; no clip or threshold is replaced.
Reference-initialized fixed templates, partial visibility and this particular
cohort limit interpretation; the run does not isolate a causal explanation for
failure or compare absolute performance with the different S8--S9 endpoints.

A separate recount verifies all 372 file hashes, all 1,020 row identities,
source eligibility, visibility, reference boxes, IoUs, shared peak scores,
decision counts and gate failures. It is not a second locator or native-agent
run. CPU batch evaluation takes 3.13s, excluding acquisition; it is not online
latency, utilization or token cost. Auditing all 763 masks of S8--S9 finds no
void pixels, so earlier counts need no correction. This failed branch is retained
rather than tuned until passage. The next efficacy step requires an independently
qualified locator and adequate natural unobserved support, not a more permissive
threshold on these examples. No natural-policy intervention benefit is established.

\section*{S11. Stronger natural front end under strict prefixes}
\label{suppidx:11}
After S10, a new plan selects the official pretrained SAM2.1 Hiera Tiny
baseline~\cite{sam2}, pinned to upstream commit 2b90b9f and a verified
156MB checkpoint. No weights are trained or tuned. Bounding-box prompts use
only the supplied source boxes; no later annotation prompts or label-based
reinitialization enter inference. Float32, logits$>0$ masks and disabled optional
postprocessing remain fixed. Each decision initializes a fresh state containing
only frames $0\ldots t$, then propagates forward to $t$. This excludes future
frames and full-clip length from the current input, at a reprocessing cost;
it is not an efficient streaming or real-time implementation.

The reused 12-sequence DEV cohort gives SAM 281/301 correct visible outputs
(93.4\%) versus VitTrack's 165/301. SAM passes the predeclared visible screen:
80\% overall, 80\% in at least eight sequences and all source identity checks.
Its 3/17 unobserved attempts do not qualify absence handling. The passed visible
screen permits only the separately specified estimator comparison, not a policy
intervention. Failed template and earlier model results remain unchanged.

The fixed next cohort takes 12 train-split sequences at alphabetical offset 12,
first 100 indices, stride 5, with all eligible initial objects. All sequences
are new to our development, but pretraining overlap is unverified. The 176 RGB
images and 176 masks supply 27 source objects and 429 object/frame cases;
four methods yield 1,716 decisions, not independent episodes. Source eligibility,
prompts, weights and cutoffs stay fixed. Images are canonicalized once to
$500\times500$ JPEG (quality 100, subsampling 0), and all methods use the same
decoded bytes; historical comparator scores are not substituted.

\begin{table}[ht]
\centering\small
\begin{tabular}{lrrr}
\toprule
Front end & DEV/301 & New/364 & Unobs./38\\
\midrule
SAM2.1 Tiny &281&320&6\\
VitTrack &165&183&38\\
Guarded ZNCC &1&0&0\\
Retained box &118&126&38\\
\bottomrule
\end{tabular}
\caption{Correct later visible localizations and erroneous attempts on new
unobserved targets. All 27 new source checks pass. Zero unobserved attempts
for guarded ZNCC coexist with refusal of all 364 visible cases. These compare
front ends, not a TWS-V checking increment.}
\end{table}

The primary, predeclared SAM-minus-VitTrack visible-correctness contrast is
$+0.376$ [$0.249,0.472$], using 2,000 paired whole-sequence bootstrap draws
(seed 20261112). The equal-sequence point is $+.372$. Intervals are exploratory
and unadjusted, with 12 sequence clusters; frames/objects are not independent
replications. Model architecture, history use, pretraining and GPU resources
also differ, so the contrast does not isolate one mechanism. Drone remains a
counterexample (25/58 correct); dog-agility has only four visible cases and
cannot count as a five-case sequence passage. SAM reaches the visible threshold
in 10 eligible sequences, with all 12 clips retained.

Full qualification still fails. SAM emits six boxes on 38 unobserved cases
(15.8\%), above the fixed 10\% limit, now with sufficient minimum support from
three sequences. Among visible cases it emits 336 boxes: 320 correct, 16 wrong;
28 cases abstain. Stronger perception and static identity therefore do not
establish decision validity. No gate is relaxed and no policy intervention runs.

A separate recount verifies all 352 source hashes, 176 canonical transformations,
1,716 row keys, prefix membership, predicted-mask boxes, references, counts and
the visible screen. It is not second inference or independent author replication.
The strict-prefix batch takes 303.13\,s including historical reprocessing and
baseline work; peak allocated CUDA is approximately 742\,MiB, not whole-GPU
reservation or end-to-end latency. Raw outputs remain on the experiment host. Additional
SAM3.1~\cite{sam3p1} UI feasibility checks are excluded from these natural performance claims.
The next test must hold the front end fixed and measure validation increment
with a separate fitting/calibration/new-test design.

\section*{S12. Fixed-front-end validity heads on new natural sequences}
\label{suppidx:12}
A further plan fixes the remaining 36 DAVIS train sequences (alphabetical
positions 24--59), allocated 12/6/6/12 to TRAIN/VAL/CAL/TEST before download.
First 100 indices at stride five provide 496 RGB images and 496 masks.
All sequences are new to our development, not proven unseen in pretraining.
Source eligibility leaves one VAL clip without an eligible target; it is kept
in the allocation, not replaced. No exact original or canonical RGB hash group
crosses splits; this is not perceptual duplicate clearance.

SAM2.1, source prompts and strict-prefix inference remain fixed as in S11.
The checker screens emitted boxes only. Its training target is visible-box
IoU below .5 or an output on an annotation-unobserved target. The latter means
unsupported by current visible annotation, not a proven wrong physical position.
An empty SAM mask remains front-end abstention, not successful checker rejection.

Cheap inputs are elapsed original-frame index, max mask logit, source-relative
area/aspect changes, center displacement, crop RGB difference and centered
grayscale correlation. The frozen ResNet-18~\cite{resnet} inputs add source/current
candidate-crop embeddings, absolute differences and cosine similarity: 1,544
shared features for the linear and 32-hidden-unit MLP heads. Only TRAIN fits
normalization and head parameters. Logistic $C\in\{.01,.1,1\}$ and the MLP's
checkpoint (150 Adam epochs) use VAL logloss. Linear-versus-MLP selection also
uses VAL; no TEST outcome enters this choice. Every cutoff is the higher 95th
percentile of CAL-valid risk, with strict rejection above it. Literal NCC/RGB
controls are fixed additively before TEST. Both CAL locks precede TEST admission;
TEST inference binds the main lock's hash.

TRAIN/VAL/CAL have 320/65/262 known emitted candidates, with only 20/1/14
composite-invalid examples. The single VAL error makes model selection fragile.
VAL selects linear $C=.01$ (loss .0916), versus MLP .1026 at epoch 49;
cheap-feature logistic scores .0651 and remains a comparator. This does not
establish a general architecture ranking.

TEST has 27 source cases and 354 later cases in 12 allocated sequences.
Of 348 visible cases, SAM emits 340 correct and eight wrong boxes; all six
unobserved cases yield empty masks. The six abstentions are SAM outputs, not
checker detections. This cohort neither qualifies absence handling nor replaces
the earlier 6/38 failure; its different clips cannot establish a paired
improvement from S11.

\begin{table}[ht]
\centering\small
\begin{tabular}{lrrr}
\toprule
Rule & Wrong/8 & Valid/340 & AUROC\\
\midrule
Keep all &0&0&.500\\
Max logit &0&11&.677\\
Elapsed frames &0&6&.306\\
Age + logit &1&7&.417\\
Cheap logistic &1&28&.804\\
Linear (selected) &0&5&.744\\
Same-feature MLP &0&11&.518\\
Literal NCC &3&41&.869\\
Literal RGB &3&31&.603\\
\bottomrule
\end{tabular}
\caption{Exploratory CAL-fixed TEST working points. Actual false rejection
is unequal: e.g., selected linear 1.47\%, age/logit 2.06\%, NCC 12.06\%.
Higher detection at a larger rejection cost is not a matched-FPR gain.}
\end{table}

The primary selected-minus-age/logit detection difference is $-.125$
[$-.500,.000$]; false-rejection difference is $-.00588$ [$-.02251,.00645$].
Paired whole-sequence bootstrap uses 2,000 draws (seed 20261113), with six
undefined detection draws reported. Intervals are exploratory/unadjusted and
conditional on the fitted models/CAL cutoffs. Eight errors in five sequences
fall below the fixed 20-error support rule: no neural increment or equivalence
is established. The selected head catches no error and withholds five correct
boxes; its .744 AUROC does not establish a useful calibrated working point.

A separate recount passes 992 source hashes, 1,129 saved SAM mask/reference
records and all method counts; 496 canonical re-encodings match exactly.
This is not independent human replication or a second model inference.
Loaded-encoder TEST feature extraction costs 6.43s on CPU for 348 pairs;
head fitting costs 1.84s. Initialization is excluded. TEST strict-prefix
inference takes 237.73s with historical reprocessing, not real-time latency.
No action, token-saving, whole-system resource or safety benefit follows.
The next efficacy test needs more independent error and missing-target support,
not changed thresholds or replaced results on this TEST.

\section*{S13. Native application integration and bounded qualification}
\label{suppidx:13}
A subsequent prototype uses an adapted OpenApps checkout~\cite{openapps},
pinned to 19c524a5, with unchanged local application handlers and native
Task/CompositeTask scoring. It is not an OSWorld result or an official
OpenApps aggregate benchmark score. The local Qwen3-VL-2B revision~\cite{qwen3vl}
89644892 supplies a goal plan; current-screenshot OCR and conservative
checkbox geometry supply a developer-adapted grounding baseline. No DOM,
reference selectors, task-target fields or native evaluator states enter the
actor. Model plans and OCR estimates are not labels. The shared actor's
capability must not be attributed to the TWS-V checker.

Adaptive development retains failures from direct point prediction, additional
model reasoning, OCR assistance and structured decoding. A later hybrid candidate
completes the three inspected development goals after separating goal parsing,
exact-label grounding, draft verification and execution. Pointer parking before
capture and a conditional field-ROI re-read are disclosed perception adaptations;
confidence and exact-match requirements are not relaxed to force passage.
These repeatedly inspected goals are not an independent success estimate.

The candidate is then frozen before a 13-case engineering qualification pack:
five atomic native goals, two new two-subgoal compositions of native todo
checks, three absent-name cases and three duplicate-name cases. Every case
uses a fresh local application state. Only duplicate-scene construction uses
evaluator-owned UI selectors; those selectors are not actor inputs. Negative
reference conditions are checked against the initial native state. Seeds and
name variants do not create independent task families.

\begin{table}[t]
\centering
\begin{tabular}{lrl}
\toprule
Case group & Pass & Criterion \\
\midrule
Atomic goals & 5/5 & Native goal \\
Two-step goals & 2/2 & Combined goal \\
Absent names & 3/3 & Abstain; unchanged \\
Duplicate names & 3/3 & Abstain; unchanged \\
\bottomrule
\end{tabular}
\caption{Frozen-candidate engineering qualification. Negative passage is not
 task completion. The small, overlapping case pack provides bounded prototype
 admission, not generalization, population success rates or TWS-V gains.}
\end{table}

All 13 case identities, saved screenshot hashes, action histories/parses,
call totals, official scores and negative reference counts pass a separate
recount. This is not independent inference or human replication. Per-case
loop times include capture preparation, OCR, planning, UI delivery and native
checking, but exclude shared model loading and server/fixture initialization;
they do not establish complete deployment cost or real-time performance.
No new weights are trained and no cloud-model calls occur.

This qualification permits a subsequent controlled comparison with the same
actor across no-extra-check, always-fresh grounding, budgeted refresh, simple
visual checks and TWS-V bounded repair. It does not execute that comparison.
Independent task-outcome improvement, natural-workflow transfer, long-horizon
benefit, cost savings and physical safety remain unestablished.

\section*{S14. Controlled native dispatch intervention pilot}
\label{suppidx:14}
After S13, a new DEV plan fixes three known goals (mark Visit grandma, Buy
groceries or Water plants done), two conditions and five policies. The shifted
condition imposes one 76-pixel downward page scroll after the first exact-label
checkbox proposal, before dispatch. This is an explicit one-row displacement
mechanism, not naturally sampled failure prevalence. The shift deliberately
invalidates old positions; this tests a controlled intervention mechanism, not
its utility under unselected changes. Each policy starts in a
fresh native app state with the same scenario seed; policy order is shuffled
with seed 20261112. Thirty episodes are paired within six scenarios over three
overlapping goal primitives, not 30 independent tasks.

All policies have the same source/dispatch capture opportunities and qualified
actor. No-extra-check executes the proposal; always-fresh re-grounds at dispatch;
random refresh uses fixed $p=.5$. The simple guard checks a $96\times64$ pixel
patch around the proposed click using grayscale NCC, requesting one recovery
below .90 or for an undefined score. Full TWS-V adds the unchanged software
source/time/content validator. The threshold, $p$ and 30-second loose host-GUI
age budget are engineering DEV constants, not CAL-selected operating points.
No source/time faults occur here, so the full and visual-only checks may agree.
No matched-FPR or matched-actual-budget claim follows from this pilot.

\begin{table}[t]
\centering
\begin{tabular}{lrrrr}
\toprule
Policy & All & U & S & Extra \\
\midrule
No extra check & 3/6 & 3/3 & 0/3 & 3/6 \\
Always fresh & 6/6 & 3/3 & 3/3 & 0/6 \\
Random $p=.5$ & 4/6 & 3/3 & 1/3 & 2/6 \\
Visual NCC & 6/6 & 3/3 & 3/3 & 0/6 \\
TWS-V & 6/6 & 3/3 & 3/3 & 0/6 \\
\bottomrule
\end{tabular}
\caption{Native DEV outcomes. U: unchanged interface; S: imposed scroll; Extra:
episodes with changes to other checkboxes. Native completion penalizes extra
changes even if the requested box eventually becomes checked. These six
related scenarios do not estimate population success or safety.}
\end{table}

Proposals were transactional. When a proposal was withheld, the runner rolled
back unexecuted progress before re-grounding on the current image. It retained
the parsed intention and allowed one recovery per dispatch, followed by
abstention when the recheck failed. Ordinary actor loops continue
up to 16 decisions; unchecked episodes are not terminated at their first failed
task check. Evaluator states and reference coordinates never enter the actor
or gate. The first implementation unnecessarily re-parses intention on some
refreshes; its 30 DEV episodes and code are retained. The stronger refresh
implementation reported here repeats all cells with cached intention shared
by every policy, without changing thresholds or native outcomes. The two
versions are not pooled as independent replications.

Native completion is 6/6 for TWS-V, 3/6 for no checking and
6/6 for both always-fresh and simple NCC. Full and NCC have identical
observed native outcomes, extra-change counts and recovery counts in all six
scenarios; this is not population equivalence or an incremental benefit of
source/time checks. Re-grounding counts are 3 for TWS-V versus 8 for
always-fresh, but frame counts are 19 versus 16. Every policy uses 6
local goal-parser calls: retaining intention removes the apparent model-call
saving. These are local calls/tokens, not billed API costs. More captures and
fewer OCR re-groundings are a resource trade-off, not established total savings.

A separate recount checks all saved proposal/image/action links, NCC and recheck
scores, gate enforcement, native final states and unintended checkbox changes.
11 logged software-contract records replay exactly using their saved source
PNG bytes and metadata. This is not a second model/OCR run, independent clock
validation or sensor attestation. Source timing is a host screenshot API interval;
unknown camera exposure or cross-device mapping is not supplied by these tests.
Loop times include capture preparation, OCR, planning and dispatch checking, but
exclude shared initialization and per-server setup; they are descriptive shared-host
measurements, not real-time, energy or deployment-startup benchmarks.

This establishes a controlled native intervention mechanism for the qualified
prototype. Independent CAL/TEST tasks, natural interface changes, broader
agents and long-horizon benefit remain open. The paper's general task-completion
claim stays a research objective; no physical-safety conclusion is drawn.

\section*{S15. Prospective unperturbed GUI instance transfer}
\label{suppidx:15}
After S14, we froze the eight new exact-name goals and compositions, actor,
runner, verifier and settings before observing outcomes. All 40 planned policy
episodes ran without imposed scrolling, delay, source swaps or other faults.
Four goals mark new existing todo names, two create new names, and two compose
mark/create or two marking operations. Names and combinations are new to the
prior actor probes; the application and the two operation primitives are known.
Initial-state names were inspected for setup, not selected by trial outcomes.
This tests instance/composition transfer, not unseen-family, real-scene or
long-horizon generalization. The study is prospective exploratory, without a
powered confirmatory test or new CAL tuning.

We used the frozen shared actor and V2 transactional recovery. We shuffled
policy order with seed 20261112 and used paired initialization seeds
9400--9407. NCC remained at 0.90, the host-GUI age budget at 30\,s, and random
refresh at $p=0.5$. No reference selectors, native state, task reward
or evaluator coordinates enter the actor or checker. The runner allowed ordinary actor-driven scrolling, with a cap of 16
decisions and one recovery per dispatch.
There are no missing episodes, infrastructure aborts or outcome-based changes.

\begin{table}[t]
\centering
\begin{tabular}{@{}lrrrr@{}}
\toprule
Policy & Goals & Re-ground & Calls & Tokens \\
\midrule
No check & 8/8 & 0 & 8 & 1601 \\
Always fresh & 8/8 & 23 & 8 & 1601 \\
Random $p=.5$ & 8/8 & 10 & 8 & 1601 \\
Visual NCC & 8/8 & 0 & 8 & 1601 \\
TWS-V & 8/8 & 0 & 8 & 1601 \\
\bottomrule
\end{tabular}
\caption{Unperturbed prospective GUI instances. Goals: native completions out
of eight related instances; calls: local goal-parser calls; tokens: observed
input plus output tokens. Forty policy episodes are not forty independent tasks.}
\end{table}

Every policy completes 8/8, with zero extra changes to other checkboxes.
TWS-V and NCC request no re-groundings, whereas always-fresh requests 23 and
random refresh requests 10. Every policy captures 46 frames and uses eight
model calls, totaling 1,411 input and 190 output tokens. Input and output counts
match on every paired instance, not merely in aggregate. This is no token
saving, no task-completion advantage, no population equivalence and no unique
benefit of the source/time checks. The successful transfer is restricted to
these new names/compositions. Stable-app controls may already succeed without
a guard; the S14 benefit under an imposed scroll is therefore conditional on
that controlled observation/action mismatch, not evidence of universal benefit.

Summed measured loop times are 50.58s (no-check), 65.74s (always-fresh),
58.55s (random), 52.71s (NCC) and 50.66s (TWS-V). They include capture
preparation, OCR, planning, checking, actions and native scoring, but exclude
per-server setup and shared initialization (1.30s model and .16s OCR startup).
These small shared-host measurements are descriptive, without a population
latency, deployment, energy or monetary-cost claim. Every policy receives a
dispatch screenshot for common instrumentation; the no-check timing is not
an uninstrumented deployment baseline.

The saved-output verifier checks all 40 native final states, proposal/screenshot
hashes, action histories, NCC values, rechecks and recovery enforcement.
All 23 logged TWS-V contract records replay exactly using the saved source
bytes and full metadata. This is software-event replay, not a second model/OCR
execution, independent physical time validation or sensor attestation.
The paired resource recount refuses incomplete or duplicate policy cells.
No old natural-video TEST, learned weights or thresholds are changed.
A time/source-plus-NCC conjunction remains a required comparator for a distinct
future contract-fault intervention study. Real captured-scene transfer remains
unestablished by this GUI experiment.

\section*{S16. Aligned natural visual-component transfer}
\label{suppidx:16}
A new prospective exploratory study uses the first 12 alphabetically sorted
VOT2018 long-term sequences~\cite{vot2018}, with four CAL and eight TEST
allocations fixed before acquisition. Observations are original indices
$0, 30, \ldots, 600$; this is sparse-prefix input, not native-frame-rate tracking
or an official VOT benchmark. All 252 sampled images are acquired, requiring
93.7MB; there are no exact original-RGB hash matches with prior acquisitions,
without clearing perceptual or pretrained overlap. Raw images remain on the experiment host.

We kept the SAM2.1 Tiny strict-prefix front end, ResNet18 features, all S12
heads, TRAIN normalizers, VAL-selected frozen-linear head and original cutoffs
fixed.
Only the source reference box initializes the target. Later independent VOT
rectangles evaluate box IoU, never prompts or features. This aligns the visual
component's candidate-validity endpoint; it does not turn box accuracy into
native agent completion. Original exposure, readiness and dispatch clocks are
unknown, not reconstructed from frame rate. The full time/source contract is
not certified by this offline component study.

\paragraph{Primary new-domain working point: not evaluable.}
The fixed minimum 24px source sides excludes three CAL sequences and two TEST
sequences, all retained in the accounting rather than replaced. Only bicycle
qualifies in CAL: 20 later frames yield 15 valid and two invalid candidates,
plus three front-end abstentions. Because fewer than 20 valid CAL candidates
remain, every new-domain 95th-percentile cutoff is explicitly unsupported.
We recorded the unevaluable primary comparison before TEST and retained the
support requirement, assigned sources, and frozen weights.

\paragraph{Secondary unchanged-cutoff check: limited support.}
Six of eight allocated TEST sources qualify and all six initial source checks
are correct. Their 120 later cases contain 112 visible cases, all correctly
localized, and eight annotation-absent cases. Four absent cases produce empty
masks and four produce wrong boxes. All four wrong candidates are from one
sequence; there are no wrong-visible candidates for that error-type analysis.
This is underpowered by the fixed class and sequence-support rules.

\begin{table}[t]
\centering
\footnotesize\setlength{\tabcolsep}{3pt}
\begin{tabular}{@{}lrrrr@{}}
\toprule
Rule & Bad rej. & Good rej. & Bad kept & AUC \\
\midrule
Keep all & 0 & 0 & 4 & 0.500 \\
Max logit & 4 & 12 & 0 & 1.000 \\
Elapsed index & 4 & 100 & 0 & 0.752 \\
Age + logit & 4 & 95 & 0 & 0.850 \\
Cheap logistic & 0 & 0 & 4 & 0.395 \\
Frozen linear & 0 & 0 & 4 & 0.567 \\
Frozen MLP & 0 & 0 & 4 & 0.868 \\
NCC & 4 & 0 & 0 & 1.000 \\
RGB difference & 3 & 1 & 1 & 0.944 \\
\bottomrule
\end{tabular}
\caption{Prespecified secondary VOT transfer at unchanged S12 cutoffs.
Bad rej.: rejected invalid candidates out of four; good rej.: rejected valid
candidates out of 112. All four errors come from one sequence. These descriptive
scores do not establish general superiority or an improved agent success rate.}
\end{table}

At its original cutoff, the selected frozen-linear head rejects none of the
four wrong candidates. NCC rejects all four without rejecting a valid candidate.
MLP has a higher descriptive AUC than linear here but likewise rejects none at
its original cutoff; it is not promoted post hoc. New-domain operating points
remain undefined; no matched-FPR or task-success increment is claimed.
The study exposes insufficient CAL/error support and a working-point transfer
limit, not a reason to relabel a negative or unevaluable result as a gain.

Saved-output verification passes all 264 source files, 147 prediction/reference
records, 116 TEST candidate scores, CAL cutoff support and TEST lock admission.
It recomputes saved masks' boxes, independent reference correctness and fixed
risks without another model/OCR inference. No old TEST, weights or cutoffs are
changed. Native intervention remains evaluated by native outcomes (S14--S15);
real-video component transfer remains a separate, here inconclusive, claim.

\section*{S17. Calibration-only diagnosis and source feasibility}
\label{suppidx:17}
The unevaluable S16 primary contrast motivates a development diagnosis rather
than threshold adjustment. Its split preceded source eligibility, leaving
only one eligible CAL sequence. Its current-locator endpoint also differs
from validity of a cached action-supporting position. Neither limitation is
resolved by reclassifying old TEST cases or changing their cutoffs.

A separate numerical audit uses only the 254 archived S9 CAL-Q cases
(96 valid, 158 invalid; five videos), not EVAL-Q or TEST. With the original
GUI head weights and TRAIN normalization, CPU float32 evaluation reproduces
the saved probabilities exactly. For linear, 168 distinct raw logits collapse
to probability 1.0. The other 86 distinct logits map to 66 distinct float32
probabilities below 1.0, for 67 distinct stored probabilities in total.
The valid-CAL 95th-percentile probability cutoff is 1.0, so the strict rejection
rule rejects none. The raw-logit cutoff is 31.893; it rejects four valid and
31 invalid CAL cases. MLP has no exact saturation and its two representations
reject the same 57 CAL cases. These adaptive CAL diagnostics are not new
efficacy estimates or evidence that logits add semantic information.
Archived evaluations remain unchanged.

Before any new inference, a source-only screen checks twelve fixed anchors
in each of the 23 unused VOT sequences. Source rectangles must have both
sides at least 24 pixels after scaling; at least six anchors must qualify.
Only eleven videos qualify, so the originally specified twelve-video
allocation is explicitly unsupported. A separate source-feasibility amendment,
made before later-reference scoring or model inference, assigns all eleven
eligible videos to four CAL and seven TEST videos without relaxing eligibility.
It fixes bounded cache horizons of 1, 5 and 15 original frames, unchanged
heads in probability and logit form, and ZNCC/RGB/cache-age references.
This is a prospective component-validity question, not a claim of native task,
timing or safety gains. Its completed independent result follows in S18.

\section*{S18. Independent bounded-cache validity test}
\label{suppidx:18}
The S17 source-feasibility amendment fixes all eleven eligible original videos
before inference: four CAL and seven TEST. The 101 qualifying anchors supply
108 CAL and 195 TEST pairs at 1, 5 and 15 original-frame horizons. Each anchor
is a separately supplied valid source initialization, not label-guided recovery.
We kept its outward-rounded rectangle fixed. Later independent VOT rectangles
were used after scoring to evaluate retained-box IoU $<0.5$, including explicit
absence.
The unchanged 3076-D GUI heads receive source/current images and the old box,
never the later reference. No weights are fitted. This is a cached-box component
study, not a native action or full TWS-V contract test.

All methods have 73 valid CAL cases across four videos, exceeding the fixed
20-case/three-video calibration floor. We saved the higher 95th-percentile
cutoffs before TEST. TEST has 149 valid and 46 invalid cases, with errors in six videos;
the predefined error-support floor is met, although seven independent videos
still limit precision. All references and scores are evaluable. Two TEST
references are absent, with 44 other invalid cached boxes on present targets.

\begin{table}[t]
\centering
\footnotesize\setlength{\tabcolsep}{3pt}
\begin{tabular}{@{}lrrr@{}}
\toprule
Rule & Bad rej./46 & \shortstack{Good rej.\\/149} & AUC\\
\midrule
Centered gray ZNCC &39&22&0.948\\
RGB difference &23&4&0.901\\
Cache frames &0&0&0.794\\
Cache metadata seconds &0&0&0.794\\
Global fresh constant &0&0&0.500\\
Linear logit &36&20&0.903\\
Linear probability &36&20&0.901\\
MLP logit/probability &20&7&0.881\\
\bottomrule
\end{tabular}
\caption{New independent cached-box TEST, with unchanged heads and new CAL-only
cutoffs. Bad/good rejection uses invalid/valid cached boxes, not action outcomes.
Same CAL targets do not guarantee matched TEST false rejection.}
\end{table}

\paragraph{Primary contrast: no increment established.}
Linear logit minus ZNCC detection is $-0.0652$,
95\% sequence-bootstrap CI $[-0.3878,0.1579]$; false-rejection difference is
$-0.0134$ $[-0.1667,0.1493]$. All 2,000 shared video resamples are evaluable;
intervals condition on the fixed CAL lock. Actual false rejection is 13.42\%
for linear and 14.77\% for ZNCC, above the CAL target. MLP rejects fewer valid
boxes (4.70\%) but detects only 20/46 invalid boxes; this different working
point cannot establish superiority at matched false rejection.

\paragraph{Numerical diagnosis versus independent benefit.}
Nineteen linear and two MLP TEST probabilities saturate at 1.0. Raw logits
restore limited linear ranking resolution, but newly calibrated probability
cutoffs lie below saturation: both forms make exactly the same TEST decisions
for each head. Their paired detection/false-rejection differences are zero.
This saved-test agreement is not population equivalence. Numerical hygiene
does not itself establish a new operating-point gain.

At five frames there are 15 invalid/50 valid cases: ZNCC, linear logit and MLP
logit AUCs are 0.913, 0.847 and 0.793. At 15 frames (31 invalid/34 valid), they
are 0.917, 0.850 and 0.809. Frame age is constant within either horizon and
has AUC 0.500; its pooled 0.794 reflects the cache-horizon mixture. The one-frame
set has 65 valid cases and no invalid case, so its AUC is not evaluable.
The one-frame AUC is reported as undefined.

Saved-output verification checks 415 files, 303 source/reference labels,
1,212 frozen-head scalars, CAL support/cutoffs and TEST lock admission. No exact
original-RGB hash overlap is found with other saved acquisition manifests;
perceptual and pretraining overlap remain unknown. The network-only retry
preserves the original frame list and both code freezes. The new design made
the primary comparison evaluable and found no learned-head advantage at the
tested working point.

\section*{S19. CAL-only cross-scene score diagnosis}
\label{suppidx:19}
The S18 outcome motivated a development diagnosis using its four CAL videos
and 108 pairs (73 valid, 35 invalid). Both GUI heads and TRAIN normalization
remained fixed. A source-relative candidate
subtracts the head's response to the valid source paired with itself:
\[
s_{\Delta}(S,C,b)=s(S,C,b)-s(S,S,b).
\]
The 3076-D source-self feature duplicates saved source/source-crop embeddings,
zeros both difference blocks and retains the supplied source box. It contains
no current embedding, later reference or newly fitted parameter. Source-self
values can reuse source embeddings; no extra encoder pass is performed here.
The candidate subtracts an anchor-dependent quantity and may alter ordering
across anchors; it is not guaranteed to remove only nuisance information.

We held out one entire CAL video at a time. The other three videos' valid
scores set the same higher 95th-percentile/strict-rejection rule for each method.
All 24 video-method folds meet the fixed support floor. These are diagnostic
fold cutoffs; the archived S18 calibration lock and TEST result are unchanged.

\begin{table}[t]
\centering
\setlength{\tabcolsep}{5pt}
\begin{tabular}{@{}lrr@{}}
\toprule
Rule & Bad rej./35 & Good rej./73\\
\midrule
ZNCC &28&6\\
RGB difference &24&6\\
Absolute linear &23&8\\
Source-relative linear &18&3\\
Absolute MLP &25&6\\
Source-relative MLP &15&6\\
\bottomrule
\end{tabular}
\caption{CAL-only leave-one-video-out diagnosis. Counts are pooled held-out
CAL cases, not a new independent TEST. Methods and all folds remain visible.}
\end{table}

The range across four videos' valid-score medians shrinks from 28.706 to
11.779 logit units for linear and 12.534 to 5.455 for MLP. These are within-head
unit comparisons, not comparable scales across methods or proof that causal
scene bias is removed. Linear's lower false rejection accompanies lower
detection. MLP detects ten fewer invalid cases at the same overall observed
false-rejection count. Thus a smaller median spread alone cannot qualify an
improved checker; useful signal can also be lost.

Source-self input isolation and exact reproduction of all 216 original CAL
head logits pass. A separate implementation recounts all 24 fold-method
thresholds, counts and median ranges without reading TEST features or running
the encoder. Four videos limit interpretation; no significance-based model
choice is made. Neither source-relative variant is promoted on this evidence.
Any later efficacy comparison needs a separately frozen new independent TEST,
with ZNCC/RGB retained and detection, rejection and adoption reported jointly.
No network complexity, old threshold or archived TEST result is changed.

\section*{S20. Joint operating points and video sensitivity}
\label{suppidx:20}
\begin{figure*}[t]
\centering
\includegraphics[width=.98\textwidth]{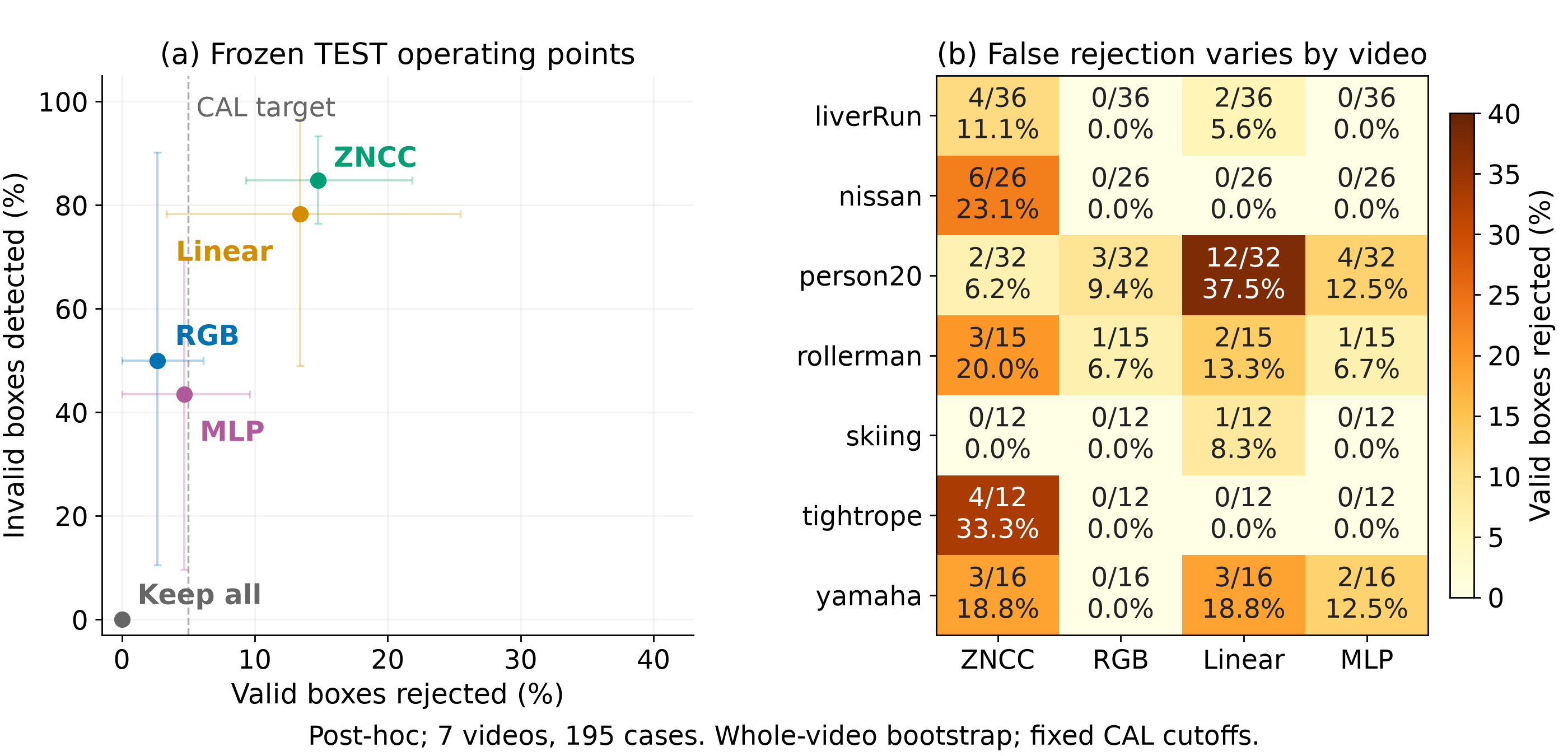}
\caption{Post-hoc frozen-output joint audit. Left: TEST operating points with
whole-video bootstrap intervals conditional on fixed CAL cutoffs; the dashed
line is a CAL target, not a TEST guarantee. Right: per-video false-rejection
counts/denominators. Seven videos are scene proxies, not new site/operator
replications. Neither view establishes natural-agent or safety efficacy.}
\label{fig:joint-operating-points}
\end{figure*}

This post-hoc audit reads only S18's unchanged saved aggregate outputs. No new
scores, thresholds, model fitting or raw TEST inputs are used. All nine methods
remain reported; probability/logit duplicates share displayed decisions.
Figure~\ref{fig:joint-operating-points} joins detection/rejection with per-video
false rejection rather than selecting a favorable scalar summary.

ZNCC's pooled false rejection is 14.77\%, but its worst video has 4/12 (33.33\%).
Linear's pooled rate is 13.42\%, yet its worst has 12/32 (37.50\%). RGB's
corresponding rates are 2.68\% and 3/32 (9.38\%); MLP's are 4.70\% and 12.50\%.
All seven videos have valid cases, while only six have invalid cases, including
two with just one and four. Zero-invalid detection is non-evaluable, not zero.
Removing one video at a time, without changing thresholds, moves linear minus
ZNCC detection difference from $-0.1935$ to $+0.0882$. This is influence
sensitivity, not a basis for excluding a video or revising the primary result.

For an explicitly illustrative error-cost ratio $\rho$, define
\[
L(\rho)=\frac{N_{\mathrm{missed\ invalid}}+
\rho N_{\mathrm{false\ rejection}}}{195}.
\]
Report all fixed ratios $0.1,0.25,0.5,1,2,4,10$ rather than choose one on TEST.
At $\rho=1$, sample count costs are 29/195 for ZNCC, 27/195 for RGB, 30/195
for linear and 33/195 for MLP. Linear minus RGB is $+0.0154$, with exploratory
95\% video-bootstrap CI $[-0.1012,0.1256]$; MLP minus RGB is $+0.0308$
$[-0.0053,0.0746]$. These new post-hoc shared resamples (seed 20261113) do not
replace S18's primary intervals. Seven videos and unadjusted comparisons limit
precision.

For these observed counts, the exact affine lower envelope switches from a
constructed reject-all reference to ZNCC at $\rho=7/127$, then RGB at $8/9$,
then keep-all-equivalent rules at $23/4$. Adjacent rules tie at boundaries.
Neither neural head minimizes this sample error cost at any nonnegative ratio.
This is not population dominance, deployment selection or measured ROI.
Linear is not individually Pareto-dominated, despite not appearing on this
affine envelope. Compute, refresh, waiting and actual action costs are absent.
Both ZNCC/RGB must remain strong references with their different trade-offs.

Independent aggregate recount verifies all nine methods, seven videos and 63
method-cost cells, paired loss differences and envelope intervals. No old
result or CAL-lock hash changes. Future independent validation must freeze its
working-point/utility specification before TEST; this audit cannot confirm it.

\section*{S21. Prospective source accounting}
\label{suppidx:21}
Further validation begins with source support, not threshold adjustment. A
frozen screen accounts for all 50 VOT2019-LT entries: 35 share nested color/
reference resource identities with the old VOT2018 collection; two repeat
previously used names, and two are conservatively reserved as unresolved
cross-collection aliases. Of eleven remaining entries, seven meet the unchanged
six-anchor/24px source rule, below the fixed floor of eight. No RGB or model
score is obtained on this pool, no split is allocated, and the floor is not
relaxed. Nested resource identities are checked explicitly; a different name
alone cannot clear duplicate provenance.

A separate prospective source screen uses the CDTB/VOT2019 RGBD
release~\cite{vot2019rgbd,cdtbrecord}. Its 80 entries are grouped into 21 fixed
name families before annotations, with only the lexical-first representative
per family. No easier replacement is selected after source qualification.
Twenty representatives qualify; the human representative has two eligible
anchors and remains excluded. The fixed twelve-family floor is met. The first
twelve qualified families are allocated by a separate frozen study plan to
four CAL and eight TEST families. Related variants remain reserved. The
resulting 120/237 planned pairs are design counts, not measured class support
or efficacy estimates.

Family names are conservative grouping proxies, not certificates of different
physical sites, cameras, operators or original sessions. The later study uses
only RGB and 2D rectangles, not depth-derived truth or measured sensor age.
The provider's record reports open access and CC BY 4.0; raw references and
RGB stay on the experiment host. A source-only recount verifies 132 VOT-LT and 252 CDTB
anchor records and earlier-manifest snapshots without parsing later-reference
lines or accessing images/models. These dispositions are retained even when
they prevent a proposed comparison; source selection is not based on method
performance.

\section*{S22. Prospective CDTB calibration transfer}
\label{suppidx:22}
The separately frozen follow-up in S21 evaluates RGB/2D cached-box validity,
not depth accuracy, sensor latency or native action success. Four source-name
families supply CAL (120 pairs, 107 valid); eight different families supply
TEST (237 pairs, 50 invalid and 187 valid). Invalid cases span seven families.
Supplied source boxes, 1/5/15-frame horizons, GUI-trained ResNet-18/linear/MLP
weights and TRAIN normalization remain fixed. No new training or TEST-based
threshold adjustment occurs. Each score is calibrated at the same nominal
95\% valid-CAL quantile, not a guaranteed equal TEST false-rejection rate.

The primary comparison changes only valid-CAL weighting: family-balanced
versus uniform ECDF, using the same quantile formula. For ZNCC, detection
decreases by 6.0 percentage points (95\% family-bootstrap CI
$[-12.5,-1.7]$), and false rejection decreases by 3.21 points
$[-6.51,-0.43]$. These original 2000 paired resamples (seed 20261114) are
conditional on CAL. This is a trade-off, not an overall improvement.
The legacy pooled-higher convention happens to give the same thresholds and
decisions as uniform ECDF for all nine scores; both remain reported.

\begin{table}[t]
\centering\small
\caption{Additional-source TEST: invalid boxes detected /50 and valid boxes
falsely rejected /187. U: uniform ECDF; B: family-balanced ECDF. All cutoffs
are CAL-only. Logit/probability variants have identical decisions; their
separate results and finite-precision AUROC ties remain in the record.}
\label{tab:cdtb-transfer}
\footnotesize\setlength{\tabcolsep}{2pt}
\begin{tabular}{@{}lrrrrr@{}}
\toprule
Score & \shortstack{U:\\det.} & \shortstack{U:\\false} & \shortstack{B:\\det.} & \shortstack{B:\\false} & AUROC\\
\midrule
ZNCC & 9 & 6 & 6 & 0 & .827\\
RGB & 20 & 5 & 15 & 1 & .858\\
Linear (logit/prob.) & 20 & 3 & 20 & 3 & .892\\
MLP (logit/prob.) & 14 & 6 & 17 & 6 & .880\\
Age (frames/nominal s) & 0 & 0 & 0 & 0 & .770\\
Global freshness & 0 & 0 & 0 & 0 & .500\\
\bottomrule
\end{tabular}
\end{table}

Linear has favorable secondary point estimates, but does not establish
superiority. After seeing the points, a separate explicitly post-hoc plan
fixes all nine scores versus both ZNCC/RGB under all three locked rules,
plus AUROC. Shared 2000 whole-family resamples (seed 20261115) give unadjusted
intervals without replacing the primary comparison. Under uniform ECDF,
linear minus ZNCC has detection difference $+.220$ $[-.026,.474]$ and
false-rejection difference $-.016$ $[-.052,.022]$; versus RGB these are
$0$ $[-.156,.200]$ and $-.011$ $[-.063,.030]$. AUROC differences are
$+.0648$ $[-.0751,.1577]$ versus ZNCC and $+.0338$ $[-.0411,.0940]$
versus RGB. All these intervals span zero; this is not equivalence.

The old VOT sample ranks ZNCC above linear (.948/.903), while this CDTB sample
reverses those point rankings (.827/.892). Different source populations and
score distributions prevent a single-factor explanation or a universal ranking
claim. Family names remain grouping proxies, not certified independent sites,
sessions or cameras. Source qualification restricts the evaluation population;
support floors are not power guarantees. Absence/identity qualification and
full-system natural-task success remain open.

All 237 TEST references and score sets are known. Verification recounts 488
saved files, 357 CAL/TEST rows, 1428 neural scalar scores and 27 method/rule
cells. Only 476 selected RGB frames and 12 reference files are acquired;
individual-file hashes are verified, not full archive checksums. No exact
frame-hash overlap is found in enumerated earlier manifests; perceptual overlap,
unobserved aliases and pretraining exposure are not excluded. Offline CAL/TEST
inference takes 2.028/3.566 seconds including loads/transforms/checks/encoder/
heads, not sensor or agent end-to-end latency. The old VOT lock/results remain
unchanged. The pre-inference dependency recovery and accounting amendment are
retained in the execution record.

\section*{S23. Joint native intervention and guard overhead}
\label{suppidx:23}
A prospective extension keeps the frozen actor, goal cache, OCR and native
application/evaluators fixed. Eight new adapted fixtures qualify before the
trial: four actual native completions and four missing/duplicate explicit
abstentions with unchanged full native state. Recount verifies 20 image hashes.
These are not eight successful tasks or natural absence qualification.
A synthetic preflight first exposes unverified clock admission; the host-GUI
adapter is corrected to pass unknown for any undeclared host mapping before
native execution. An initial receipt-filename failure stops before model/server
startup. Its log and original freeze are retained; neither recovery changes
the actor, tasks or thresholds after outcomes.

Four different TEST goals, including two two-step compositions, are crossed
with three conditions and seven policies, yielding all 84 planned episodes.
These are four related task instances, not 84 independent tasks. Evaluator-only
fixture setup changes fictional todo names through native APIs; the actor uses
only goal text, screenshots and its action history. No reference state, selector
or native label enters the actor/guard. Protocol, code and manifest are frozen;
no model training, cloud call or TEST-threshold adjustment occurs.

\begin{table*}[t]
\centering\small
\caption{Joint native TEST. Each cell gives completions /4 and re-groundings.
Natural variation is not inferred from imposed scrolling or metadata faults.
Fixed alternate refreshes every second click/type proposal. .90 ZNCC, .05 RGB
and 30s are frozen engineering working points, not matched TEST error/budgets.}
\label{tab:native-joint}
\begin{tabular}{lrrrrrr}
\toprule
&\multicolumn{2}{c}{Unchanged}&\multicolumn{2}{c}{Imposed scroll76}&\multicolumn{2}{c}{Benign context mismatch}\\
Policy & Complete & Re-ground & Complete & Re-ground & Complete & Re-ground\\
\midrule
No check &4&0&1&0&4&0\\
Always re-ground &4&10&4&10&4&10\\
Fixed alternate &4&2&1&5&4&2\\
ZNCC &4&0&4&3&4&0\\
RGB &4&0&1&0&4&0\\
Time/source/hash AND ZNCC &4&0&4&3&4&4\\
TWS-V &4&0&4&3&4&4\\
\bottomrule
\end{tabular}
\end{table*}

The primary full-versus-simple-conjunction contrast has zero discordant
completions in either direction for every condition. Actual actions, recovery,
capture, model-call and token counts match on all twelve paired scenarios.
This is observed agreement, not population equivalence or unique TWS-V efficacy.
On the imposed-scroll sample, full has a favorable completion/refresh point
versus fixed alternate (4/4 and three calls versus 1/4 and five), also achieved
by ZNCC and the conjunction. This is not a matched-budget population result.
Always re-ground completes 4/4 with ten calls; equal counts are not noninferiority.
Full captures 21 frames versus always's 20, so fewer re-groundings need not
mean fewer screenshots. No-check, fixed alternate and RGB each make unintended
checkbox changes in three scroll episodes; none are observed for the other
four policies. RGB's working point is retained rather than retuned on TEST.
One no-check scroll episode completes: imposed movement alone is not an
action-invalidity label.

All intended imposed interventions are reached. Producer expectations precede
the independent context-ID mutation. This condition is benign to the native
goal but invalid under the provenance contract: full/conjunction enforce it,
requiring four re-groundings without increasing completion. Contract rejection
therefore is not automatically a prevented task error. Each condition uses
713 input/108 output tokens and four parser calls per policy, with equality
on every paired goal. OCR re-grounding does not reparse unchanged intentions;
no token reduction is observed.

Full loop totals are 22.069/24.870/25.629s across the three conditions, versus
always 28.353/32.059/27.763s and conjunction 21.511/25.088/25.988s. These include
instrumented dispatch screenshots, OCR/planning/checking, forced settling,
logging, UI execution and native scoring; setup and shared loading are separate.
Shared-host timing and four related goals do not support population efficiency
or generalization claims. Failures remain in denominators and resource totals.

\paragraph{Isolated guard measurements.}
Twenty randomized/interleaved blocks use sixteen deterministically selected
saved source/current pairs from S15, with five repetitions per block (1,600
calls/guard). Includes producer hashing, local crops/arrays, validation and
audit-dictionary preparation; excludes PNG decoding, capture, OCR/model,
actions and storage I/O. Mean wall/CPU ms per call are .0224/.0224 for the
instrumented no-check wrapper, .0632/.0632 for ZNCC, .0680/.0680 for RGB,
.0850/.0849 for conjunction and .1558/.1558 for full. Full's block-mean wall
p95 is .1692ms, not per-request tail latency or a confidence interval.
The hypothetical 10 Hz CPU-duty projection (.1558\% of one core) is not measured
system CPU. Peak process RSS of 145,112 KiB includes libraries and 32 cached decoded
screenshots, not incremental guard memory. This local-patch workload does not
replace a broader feature-bundle profile. No whole-agent 1\% CPU, energy or billed-cost claim follows.

Saved-output verification covers all 84 native outcomes/cells, 219 dispatch
decisions, 455 image links and 72 initial/recheck contract records, recomputing
ZNCC/RGB, action histories, actual scrolls, expectations and bounded rechecks.
It is not second inference or independent author replication. The run/episode/
step/phase tuple scopes events; the standalone dispatch ID is not a globally
unique stream identity or anti-replay proof. All frozen sources remain unchanged.

\section*{S24. Evidence-to-claim alignment}
\label{suppidx:24}
Engineering validation, offline component validity and closed-loop task outcome
are different evidence classes. Engineering checks support enforcement of the
declared host-side contract and saved-event replay; they do not attest sensor
origin, truthful timestamps or semantic correctness. Offline DAVIS/VOT/CDTB
results concern supplied or predicted 2D boxes and source-qualified populations,
not task completion. Their cross-source point-ranking change does not establish
a universal learned advantage; negative and unsupported results are retained.

Closed-loop controlled tasks support only their fixed actor, scenarios and
actual outcome/resource counts. S23 adds strong simple controls and a favorable
observed outcome/refresh pair versus fixed alternate, but not a unique advantage
over ZNCC or the conjunction. New GUI names/compositions are not natural task
families. Combining natural-video component tests with imposed GUI interventions
does not establish natural end-to-end generalization. Visible qualification,
absence handling and persistent physical identity remain separate. Unnecessary
refresh, capture and complete loop cost accompany detection/completion counts.
Token saving, physical safety and general long-horizon deployment efficacy remain
unestablished. Nonsignificance or equal observed counts do not establish equivalence.

\section*{S25. Natural SAM3 concept qualification and candidate supply}
\label{suppidx:25}
We ran a bounded exploratory SA-Co SmartGlasses~\cite{saco} validation qualification using
SAM3~\cite{sam3}, not SAM3.1 or persistent physical-instance tracking. Eight
videos supplied 128 uniform frames; a reference-only frozen sampling rule
selected 20 annotated-absent, 12 uniquely visible and 20 multiple-instance
cases, with at most four cases/state/video. The required 20 cases/state across
at least three videos were not met for unique visibility. Combined confidence
$>.5$, mask probability $>.5$ and a 90\% correctness admission rule were fixed
before inference; unique correctness additionally requires mask IoU $\geq .5$.

Correct decisions were 20/20 absent, 5/12 unique and 13/20 multiple; three
multiple cases were falsely classified as unique. All seven remaining unique
cases produced no candidate. Empty front-end output is abstention, not checker
success. Nineteen absent cases are explicitly negative video--phrase pairs;
only one is an unobserved frame of a positive phrase. Thus this sample does not
qualify natural occlusion/reappearance. No new small head was trained and no
previous TEST or threshold changed. These small selected DEV counts do not
estimate general SAM3 performance or natural prevalence.

A separately frozen, post-hoc raw-proposal diagnostic reused the same twelve
unique DEV cases. It replayed native counts exactly. Without the confidence
cut, score-ranked top-1 masks meet IoU $\geq .5$ in 10/12 cases; the top-five
reference-assisted upper bound is also 10/12. Five of seven empty outputs
already contain a correct top-1 proposal. This motivates separating candidate
supply from presence confidence, but is not an adoption improvement: always
emitting top-1 also emits on absent queries, and oracle matching is unavailable
online. The saved legacy field \texttt{top1\_oracle\_correct} denotes score-ranked
top-1 correctness, not reference-based candidate selection.

Official code and checkpoint hashes, source dispositions, initial access/load
failures, recovery restrictions and saved-mask verification are recorded.
All active weights loaded; only 22 weights of a disabled interaction branch
were ignored. Raw media/masks remain on the experiment host. The shared archive
was streamed with a fixed prefix cap; only selected validation JPEGs were saved
or decoded, although skipped opaque bytes can contain other sources. No official
TEST annotation was opened. Candidate-based future comparisons require the same
raw proposal pool for native simple rules, linear and MLP heads, separately
calibrated thresholds and source-grouped independent evaluation. Concept
uniqueness and retained-action validity remain distinct endpoints.

\section*{S26. Constrained candidate recovery on new videos}
\label{suppidx:26}
A new exploratory study tests whether a small adoption head improves on
recalibrated SAM3 using an identical frozen raw top-five candidate pool. The
old 52 DEV cases and eight videos remain excluded. A separate six-video
qualification has 24 cases/state: top-five coverage is 24/24 unique targets,
score-ranked top-1 is correct for 20/24, and default correct adoption is 8/24.
This passes a separately frozen raw-supply gate, not the earlier native-adoption
gate. Verification recounts 360 masks before fitting.

Source-locked internal splits of official validation are TRAIN 59 videos/416
cases, VAL 24/172, CAL 37/276 and TEST 66/491. TEST has 152 unique cases from
47 video proxies, 142 bracketed-unobserved cases from 49 and 197 multiple cases
from 56. Positive-phrase annotation gaps have visibility before and after, but
the physical cause is unverified; multiplicity is not certified similar-instance
identity ambiguity. No official TEST annotation is opened. Video grouping does
not prove participant/take independence or exclude pretraining exposure.

The 26 score features comprise presence, five detector and five product scores,
five predicted-mask areas and ten pairwise mask overlaps. Another tier appends
1024 frozen global/crop ResNet18 features shared by linear and MLP heads.
IDs/hashes/source names are excluded. All rules select the same top-1 candidate;
references determine correctness only after prediction (unique and IoU $\geq .5$).
No reference-assisted candidate selection or learned reranking occurs.
TRAIN-only normalization and BCE fitting, VAL checkpoint/model selection and
CAL-only thresholds remain separate. Primary seed 20261112 is fixed; two other
seeds are sensitivity checks. CAL maximizes correct source-macro adoption under
separate empirical 5\% error constraints for unobserved, multiple and wrong
unique cases; these do not guarantee TEST error. VAL selects the strongest simple
rule (product times detector margin) and score MLP before TEST analysis.

\begin{table*}[t]
\centering\small
\caption{New-video TEST, locked CAL operating points. Percentages average video
rates within each reference state; raw correct counts have denominator 152.
Unobserved/multiple denominators are 142/197 cases. Different source weights
can reverse raw and macro differences. Native default uses its original rule.}
\label{tab:sam3-recovery}
\begin{tabular}{lrrrrr}
\toprule
Rule & Correct adoption & Unobserved error & Multiple error & Wrong unique & Raw correct\\
\midrule
Native default &48.05&2.72&11.16&1.06&78\\
CAL product/count &48.05&2.55&8.48&1.06&78\\
VAL-selected simple &52.30&2.55&9.38&1.60&80\\
Score linear &54.61&6.46&9.38&3.19&90\\
Score MLP (primary) &50.35&5.78&8.48&1.60&82\\
Score + visual linear &7.98&10.71&10.71&2.66&15\\
Score + visual MLP &9.22&1.70&7.14&1.06&17\\
\bottomrule
\end{tabular}
\end{table*}

\begin{figure*}[t]
\centering
\includegraphics[width=\linewidth]{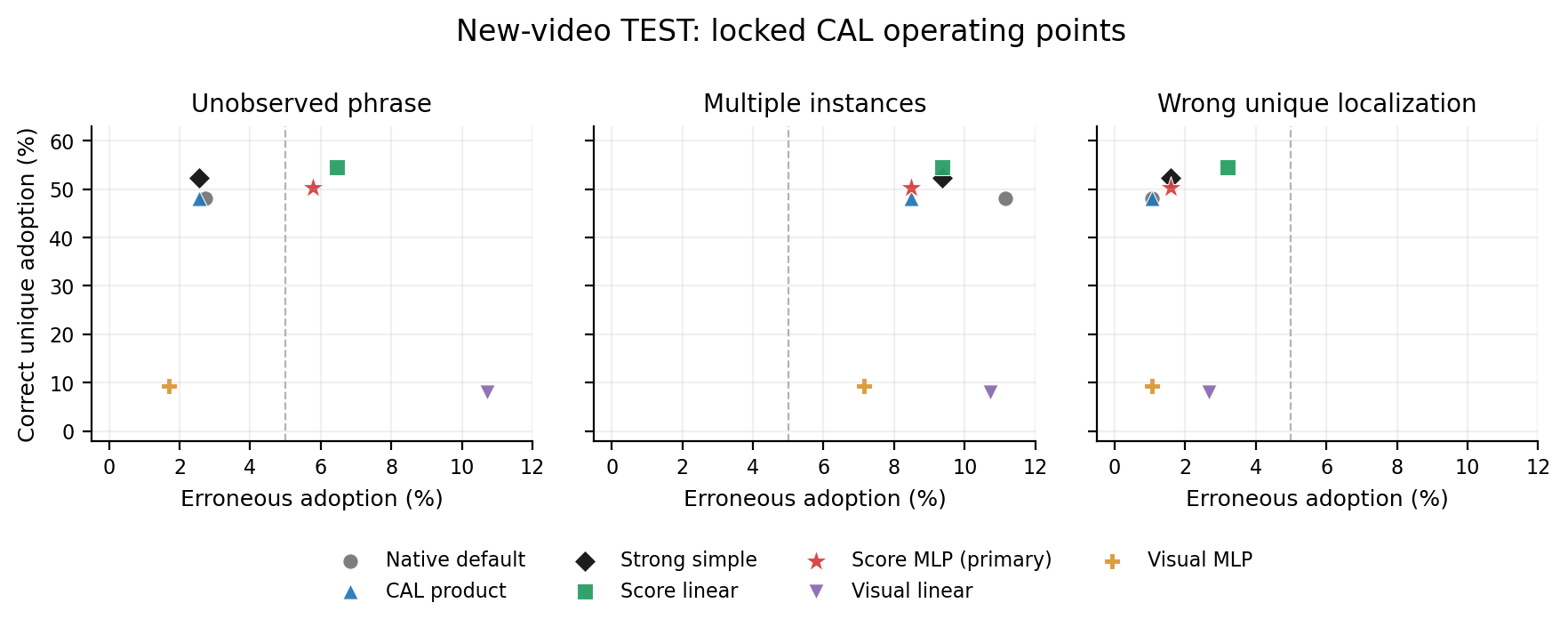}
\caption{Descriptive TEST operating points; no TEST threshold sweep. Each panel
pairs correct unique adoption with a different error endpoint. The dashed 5\%
line denotes the empirical CAL target, not a guaranteed TEST bound. Primary
paired uncertainty is reported in text; these points alone do not establish
superiority or noninferiority.}
\label{fig:sam3-recovery}
\end{figure*}

The frozen joint criterion is not met. Primary MLP minus selected simple correct
adoption is $-1.95$ percentage points, simultaneous interval $[-13.54,7.99]$;
unobserved error increases $3.23$ points $[.47,7.27]$. Multiple and wrong-unique
differences are $-.89$ $[-8.49,4.55]$ and $0$ $[-2.38,2.22]$ points. Ten-thousand
shared source resamples and conservative paired source-error bounds use the
predeclared multiplicity budget; the 2-point error-increase limits are not
established. The primary benefit point is negative, so this is not merely
insufficient noninferiority power. Raw MLP correct counts increase 80 to 82
(11 rescued, nine lost), whereas the prespecified macro point falls; both are
retained. TEST top-five coverage is 141/152 and top-1 correctness 119/152.
Candidate availability therefore does not establish constrained adoption value.

All three seeds remain in the record. Score-linear correct rates span
32.09--54.61\%, score-MLP 44.50--50.35\%; no best TEST seed is selected. Visual
features do not help under these fits/rules, not proof of general uselessness.
Primary heads have 27/897/1051/33665 parameters for score linear/MLP and visual
linear/MLP. The complete fit/selection/calibration/TEST/bootstrap stage takes
18.27s. TEST extraction and verification take 129.62s/12.52s including offline
loads, reference evaluation and I/O, not sensor-to-action latency or whole-agent
CPU. The fixed stream retains 1340 prescribed JPEGs while reading 31.80GB in
4360.8s; the full archive is not stored. Raw images/masks/features/weights stay
on the experiment host. No token, industrial ROI or task-completion gain follows.

Saved-output verification recounts 20 methods, all 491 TEST decisions, grouped
metrics, raw events, source isolation, rescue/loss counts and the final criterion;
all five split mask/feature verifications pass. Before real inference, a logged
A3 correction makes the conservative bound account for removed as well as new
errors; original freezes remain. Neither targets nor TEST thresholds change.
Further development requires a new independent evaluation, not retuning these
491 cases. This component study does not establish natural closed-loop efficacy.

\paragraph{Frozen predictive-feature control (exploratory).}
We also test whether the official VPP~\cite{vpp} one-step predictive representation improves this adoption problem without changing SAM3 proposals,
prompts, score features or top-1 selection. The released \texttt{svd-robot}
extractor (official commit \texttt{75cf37a}) passes a one-image qualification:
its $1{\times}16{\times}1280{\times}16{\times}16$ feature is finite and exactly
repeatable under a fixed seed, with 5.31~GB peak allocated GPU memory. We mean
pool full-frame and top-1-proposal-crop features, then give the same score inputs
to linear and 32-hidden-unit heads. TRAIN videos are hash-split into fit and
threshold-calibration groups; CAL and TEST are not opened. On held-out VAL, the
selected detector-score rule obtains .196 correct adoption for unique targets,
with false-adoption rates 0/.028 for unobserved/multiple cases. Primary-seed VPP
linear obtains .036 with .025/.014 errors; VPP MLP selects full rejection. VPP
linear minus the simple rule is $-.161$ correct adoption (paired video bootstrap
95\% interval $[-.333,-.017]$). Three seeds are retained and are unstable rather
than selected post hoc. Extraction averages about 0.15~s per case for the
two-image batch, excluding model load; this is offline readiness, not
sensor-to-action latency. Thus this adapter fails the frozen benefit criterion.
It does not show that predictive features are generally useless, and a predicted
feature remains derived evidence rather than a renewed observation.

\section*{S27. Conditional guarantees and interpretation boundaries}
\label{suppidx:27}
\paragraph{Identifiability with rejection or refresh.}
Let $Z$ be the complete evidence available to the checker, not the hidden
reference state. For a fixed action, suppose two possible states have the same
$Z=z$ but opposite binary validity $V$. A randomized checker whose randomness
contains no hidden-state information has the same adoption probability $p(z)$
on each. Thus $P(A=1\mid z,V=0)=p(z)$ and
$P(A=0\mid z,V=1)=1-p(z)$. Their sum is one. Returning unknown or requesting
refresh counts as not adopting at this dispatch. New evidence can change $Z$
and escape the pair; the argument does not preclude later recovery. Always
abstaining obtains no valid adoption coverage on the valid twin. The empirical
identical-pixel pairs demonstrate this boundary in imposed fixtures, not natural
prevalence or a universal impossibility of successful agents.

\paragraph{Temporal soundness is conditional.}
Assume $o\in[o^-,o^+]$, $a\in[a^-,a^+]$, $d\in[d^-,d^+]$ in one correctly
mapped clock and nonnegative declared budget $B$. From $o^+\le d^-$ follows
$o\le d$; $d^+-o^-\le B$ implies $d-o\le B$; $a^+\le d^-$ implies $a\le d$.
Therefore actual age is in $[0,B]$ and readiness precedes dispatch. The current
validator checks these conservative order/age conditions, but does not establish
that intervals or source assertions are truthful. A successful hash check is
content linkage under its declared trust origin, not source attestation.
No semantic completeness or physical correctness follows.

\paragraph{Horizon risk requires additional premises.}
At reached dispatch $t$, let $E_t=\{A_t=1,V_t=0\}$ and let $F_t$ be the event
that the contract's factual premises fail. If $P(F_t)\le\delta_t$ and
$P(E_t\cap F_t^c)\le\epsilon_tP(A_t=1,F_t^c)$, then
\begin{align}
P\!\left(\bigcup_{t=1}^{H}E_t\right)
&\le\sum_{t=1}^{H}P(E_t)\\
&\le\sum_{t=1}^{H}\left[\delta_t+
 \epsilon_tP(A_t=1,F_t^c)\right].
\end{align}
This is a union-bound argument, not a new concentration theorem. Independence
is unnecessary. If the premises hold conditionally for every reached history,
total expectation gives the stated marginal premises. Errors other than the
defined invalid adoption can still prevent success or cause harm. The bound
may be vacuous for a large horizon. Selected negative-state CAL error is not a
history-conditional adopted-action risk; class priors, dependence, selection and
state changes matter. We have not established deployment values of $\epsilon_t$
or $\delta_t$, and report no certified safety number from this expression.

\paragraph{Execution equivalence is a required control.}
Suppose two controllers receive identical observations and action proposals,
apply the same admission predicate, choose the same recovery transitions, and
use the same actor, initial state and coupled randomness. At the first step
their executed actions agree. If histories agree through step $t$, the same
inputs/transitions give the same action and next history, completing induction.
Trace recording alone cannot improve task completion under these premises.
Different timing or I/O can break the premises and must be measured, not assumed
beneficial. S23 retains the matching full/conjunction outputs rather than treating
a functional clone as a method that ought to lose.

\section*{S28. Official-task integration and failed actor admission}
\label{suppidx:28}
We pin OSWorld 2.1~\cite{osworldrelease} code, its official task and
asset revisions, provider-image hash and container digest. All 108 task sources
match the released hash manifest; all 1,084 asset files
pass size/hash checks. The official 82-ID offline list is a structural inventory,
not proof that every task dependency is executable. No cloud infrastructure,
live accounts or external websites are added by this admission.

The dedicated VM uses a read-only verified base, ephemeral overlay, an internal
network and host-loopback control relays. Two fresh boots expose a nonblank
desktop; actual typing into an owned test window is read back, and the temporary
marker disappears on the second boot. An earlier private-port configuration
failed, and an intermediate HTTP-only admission returned nearly black startup
frames. Both records remain; only the stronger desktop/input/reset check is
admitted. These are engineering properties, not perception or task gains.

Task 001, the first listed offline task, was fixed before model output. Its
instruction, setup and native evaluator are unchanged. A cached frozen
Qwen3-VL-2B-Instruct actor receives the current screenshot, instruction and its
last eight executed actions, without DOM, evaluator state or task source.
Generation is deterministic, with a 192-token cap and at most 786432 image
pixels. Only allowlisted GUI actions run inside the disposable VM; arbitrary
model-generated Python is never executed. This custom adapter is not a
reproduction of a published Qwen agent.

The first response has malformed JSON and executes no action; the native score
is 0. A new DEV run permits only quoting bare allowlisted keys, without changing
action values. It dispatches all 20 allowed actions through the official
environment, repeats the same click 20 times and again scores 0. Per-call guest
acknowledgements were not recorded in this pilot; dispatch count is not certified
low-level input acceptance. The run makes 20 model calls (49,145 input and 434
output tokens) in 61.02 seconds including setup/load/cleanup; these are actual
counts, not token savings. A subsequent four-position owned-button diagnostic
initially rejects fenced output; a syntax-only outer-fence adapter then yields
three native button hits in four cases. All four target locations are verified
by numeric rendering checks. The fixed 4/4 admission criterion is not met.
These selected DEV cases do not estimate general model grounding accuracy.

Thus the screenshot--actor--guest-dispatch--native-evaluation chain runs, but
this 2B actor is not admitted for an efficacy comparison. Raising the step
cap, changing the evaluator or pooling failed localization with task success
would not establish long-horizon benefit. A stronger actor requires new
qualification before a frozen paired intervention; the present result supplies
no TWS-V task advantage, independent natural generalization or safety claim.
Raw frames and gated task contents remain on the experiment host; only source,
aggregate receipts and hashes enter the repository.

\section*{S29. Prospective three-task official application comparison}
\label{suppidx:29}
A stronger frozen Qwen3-VL-4B-Instruct snapshot~\cite{qwen3vl} was acquired on the
experiment host and verified against pinned full-file hashes.
It passes the unchanged four-position owned-button check (4/4 native hits,
no syntax repair). This diagnostic does not qualify multi-step planning or
application competence. We use the same bounded custom screenshot actor,
not a published Qwen agent reproduction or a newly trained policy.

Before application outcomes, a hash-locked plan fixes task 001 (mail/calendar,
reused DEV), task 003 (image editing) and task 004 (presentation), seven
controllers, task-wise shuffled order, identical recovery capabilities and
unchanged native setup/evaluators. Tasks 003/004 provide prospective exploratory
new-task transfer; pretraining independence is unverified. The seven controllers
are no additional check, unconditional second proposal, alternating second
proposal, ZNCC, RGB difference, simple conjunction and TWS-V.
The no-check arm already receives a fresh screenshot each decision;
\emph{always fresh} means an unconditional \emph{second} proposal at dispatch.
ZNCC .9, RGB difference .05 and host request age 30 seconds are inherited DEV
settings, not choices optimized using these task outcomes. Each cell gets a
fresh disposable VM, a 500-action cap, a 2700-second wall cap and the same stop
after eight identical consecutive pointer actions. A large cap is not a
measurement of long-horizon competence when the actor stops early.

A retained first attempt has four task-001 zero-score cells due to malformed
curly quotation marks around one JSON key; a fifth cell is interrupted before
evaluation and remains unavailable. A syntax-only amendment, made before any
task-003/004 outcome, normalizes quotes around allowlisted keys outside typed
string values. It changes no coordinates, text values, prompts, thresholds,
model weights or evaluators; all four qualification actions replay unchanged.
These failed first attempts are not pooled with the amended batch.

All 21 cells complete native evaluation and saved-output verification; all
native scores are 0. There are 112 dispatches with successful guest API responses,
no unknown or unsuccessful API response, 142 model calls (338,758 input and 2,970
output tokens), and 21 re-groundings. Nine runs end with the actor's \texttt{done}
response despite native failure; twelve reach the common repeated-pointer stop.
These are observed software responses and task failures, not per-action semantic
labels. The seven image-editing runs end before any action. Eight syntax repairs
occur in the amended batch and remain counted.

\begin{table}[ht]
\centering\small
\resizebox{\columnwidth}{!}{%
\begin{tabular}{lrrr}
\toprule
Controller & Completed/3 & Calls & Re-groundings\\
\midrule
No additional check &0&19&0\\
Unconditional second proposal &0&24&11\\
Alternating second proposal &0&28&9\\
ZNCC &0&19&0\\
RGB difference &0&19&0\\
Simple conjunction &0&14&1\\
TWS-V &0&19&0\\
\bottomrule
\end{tabular}}
\caption{Frozen three-task official application subset, exploratory. Each of
21 task/controller cells receives a fresh VM and unchanged native scoring.
Zero completions for every controller do not establish equivalence, useful
call savings, or a success/cost advantage. Counts are actual and not matched
repair budgets. One run per selected task/controller does not estimate
population performance.}
\end{table}

This comparison exposes an actor-competence floor: four-point grounding admission
was insufficient for these application workflows. The result does not establish
that TWS-V improves or harms general task completion, nor that a stronger
published agent would fail. In particular, fewer calls on unsuccessful runs are
not evidence of useful token savings. Static contract and visual checks cannot
supply a missing correct action or certify the model's completion assertion.
A new actor or planning configuration requires separate development qualification
and a fresh frozen task sample; these tasks and thresholds are not retuned.

The screenshot adapter assumes a trusted guest captures within the host request
interval; it does not establish sensor exposure time, cross-device synchronization
or hardware attestation. Pointer checks use a local image patch; keyboard checks
use a whole-screen change proxy, not verified focus or semantic identity. Final
native scores do not label each action as valid, and re-grounding counts do not
identify unnecessary refreshes. The task subset is too small for population
superiority or equivalence claims. Passing saved-output verification is an
engineering/accounting result, not an independent repeat of model inference.
Raw frames and task contents remain on the experiment host.

\section*{S30. Harness-side dispatch and workflow development admission}
\label{suppidx:30}
A review of the pinned HarnessRouter/UHP protocol~\cite{harnessrouter}
finds server-owned tool execution (\href{https://github.com/HarnessRouter/harnessrouter/blob/f47d736d01d0bbd4b9489edd6615038d733eeef3/protocol/versions/2026-10-19/tasks.md}{pinned task contract}): reported tool events do not provide a
client-executed tool-return loop. Thus a dispatch gate belongs inside the tool
executor, not an after-the-fact event listener. Our local synchronous boundary
binds an action proposal to its frame, requires admission before execution and
refuses duplicate dispatch after ambiguous I/O. Nine focused boundary/output
checks pass; this is not live HarnessRouter conformance, empirical support for
multiple harnesses or a new semantic/safety guarantee.

We next freeze three independently authored application DEV workflows: replace
and save text, rescale/export an image, and edit/save slide text. These are not
released OSWorld tasks. A retains the frozen 4B, one-image, 192-output-token actor.
B adds screenshot-pair context, subgoal memory, bounded repeated-action replanning
and a second completion judgment at the same output cap. C uses B with a 768-token
cap. B is a bundled harness change, not a pure planning ablation. Saved outputs,
not model completion claims, supply the development predicates. The unchanged
admission requires all three workflow predicates and completed endings.

\begin{table}[ht]
\centering\scriptsize
\setlength{\tabcolsep}{3pt}
\begin{tabular}{p{.38\columnwidth}ccc}
\toprule
DEV profile & Text saved & Image exported & Calls\\
\midrule
A: retained actor &Yes&No&17\\
B: memory/review bundle &No&No&38\\
C: larger output cap &No&No&34\\
\bottomrule
\end{tabular}
\caption{One authored instance per workflow/profile, exploratory. Calls include
the original presentation diagnostic, which is not admitted for intended-task
comparison after the context audit below. These are resource counts, not
population success rates or useful token savings. No response reaches its cap.}
\end{table}

The first pilot's XML-only presentation scoring is retained as superseded.
A prospective amendment preserves output artifacts, supports both flat and
packaged ODF, and reports unparseable output as unavailable rather than zero.
A separate no-model replay then identifies the actual context problem: the
presentation fixture loads in Writer rather than Impress, despite a live
launcher. All original presentation cells remain, but are not admitted for the
intended task. This fixture issue does not rescore or explain the earlier
\emph{official} OSWorld batch. An XML declaration and explicit import filter
pass a new Impress-focus audit. A separate corrected-setup/clarified-wording
single-case diagnostic gives native failure for A, B and C (5, 10 and 10 calls).
Each verifies actual Impress focus before inference; none is a TWS-V treatment.

One further frozen post-hoc DEV baseline D supplies documentation-informed
editing/scaling/export procedures to A, retaining the same model and 192-token
output cap. It verifies the expected foreground application before inference.
Text saving passes, image export reaches the 30-action cap without output, and
slide editing terminates on malformed action JSON; the saved output does not
pass. D makes 39 calls (97,011 input and 812 output tokens). These are selected
workflow diagnostics, not an independently tested skill-learning advantage.
No profile passes the original three-workflow gate; failed cases are not removed.

These cases do not establish that planning, memory or larger models are useless.
They show that the tested changes do not by themselves qualify this actor.
Completion review remains a fallible model judgment. Saved-output re-evaluation
and log recounts verify accounting, not an independent model replication.
All weights remain frozen; no paid inference or TEST-threshold tuning occurs.
Larger intervention claims remain gated on a competent executor and independent
joint outcome--cost tests with the strongest simple controls.

\section*{S31. Expanded controlled native study with a common repair cap}
\label{suppidx:31}
\paragraph{Admission and protocol.}
A frozen extension retains the goal-text parser, local Qwen3-VL-2B cache,
OCR/coordinate executor and native OpenApps evaluators. Eight new positive
cases complete without extra state changes; eight missing/duplicate cases
explicitly abstain with the full state unchanged. These sixteen controlled
fixtures qualify this executor, not natural target absence. Twelve excluded
smoke episodes and 24 CAL episodes are recorded separately. A first excluded
smoke attempt stops on an uncalibrated random-policy call; its partial records
remain. Revision r1 omits that policy only from smoke, before CAL/TEST, without
changing the actor, scientific matrix or old thresholds.

Twenty constructed goals (single marks and two-operation compositions) cross
unchanged, imposed 76-pixel scroll and benign background-style conditions with
five policies: all 300 planned TEST episodes run once. Each policy has the same
maximum six extra re-groundings and sixteen steps. CAL fixes random-refresh
probability at $3/21=1/7$; this is frequency calibration, not equality of realized
TEST cost. ZNCC .90 and the 30s age working point remain unchanged. The native
server state is reset before each arm, a new browser context is used, and policy
order is shuffled within each goal/condition; condition order is fixed.
The actor receives goal text, screenshots and action history, never native
reference state, selectors or hit-test labels. All metadata bindings are healthy
in this study; the new perturbations are visual, not provenance attacks.

\begin{table*}[t]
\centering\small
\caption{New controlled TEST, twenty goals per condition. Condition cells give
native completions /20 and actual re-groundings in parentheses. Common maximum
allowance does not mean equal realized cost. Clean completion additionally
requires no unintended recorded native-state changes.}
\label{tab:native-common-cap}
\begin{tabular}{lrrrrr}
\toprule
Policy & Unchanged & Scroll76 & Benign style & Clean /60 & Wrong-state episodes\\
\midrule
CAL-frequency random & 20 (0) & 5 (3) & 20 (0) & 45 & 15\\
ZNCC & 20 (0) & 20 (15) & 20 (0) & 60 & 0\\
Contract AND ZNCC & 20 (0) & 20 (15) & 20 (0) & 60 & 0\\
TWS-V & 20 (0) & 20 (15) & 20 (0) & 60 & 0\\
Always re-ground & 20 (50) & 20 (50) & 20 (50) & 60 & 0\\
\bottomrule
\end{tabular}
\end{table*}

\paragraph{Primary contrast and actual budget.}
TWS-V minus the transparent contract--ZNCC conjunction is zero in native
completion: no discordance in either direction across all sixty paired
scenarios. Executed action trajectories and non-timing resource counts also
agree on every scenario, not just in their totals. ZNCC alone agrees likewise.
The planned 2,000 paired resamples of twenty goal blocks yield $[0,0]$ for this
empirical contrast; this is observed agreement, not population equivalence.
The blocks are related constructed instances, not independent task families.

TWS-V completes 60/60 versus random's 45/60 in the fixed three-condition mixture,
but uses fifteen versus three actual re-groundings. Thus this is not a
same-actual-cost success-rate advantage. Random's fifteen unsuccessful episodes
all have unintended state changes in the scroll condition. Relative to always
re-ground, full has equal observed completion with fifteen versus 150 repairs,
also achieved by ZNCC and the conjunction. Equal sample success is not a
noninferiority finding. This is one common-cap point, not a Pareto frontier.

\begin{table*}[t]
\centering\small
\caption{Complete TEST costs summed over sixty episodes per policy. Tokens are
actual parser inputs/outputs; repairs can reuse parsed intentions. Setup removes
first-arm cold-start asymmetry and equally amortizes the shared server start.
Shared model/OCR initialization is reported separately.}
\label{tab:native-common-cost}
\begin{tabular}{lrrrrrr}
\toprule
Policy & Repairs & Captures & Model calls & Input/output tokens & Loop s & Setup s\\
\midrule
CAL-frequency random & 3 & 355 & 60 & 10767/1692 & 386.114 & 106.144\\
ZNCC & 15 & 305 & 60 & 10767/1692 & 345.495 & 105.731\\
Contract AND ZNCC & 15 & 305 & 60 & 10767/1692 & 350.598 & 105.700\\
TWS-V & 15 & 305 & 60 & 10767/1692 & 346.675 & 105.706\\
Always re-ground & 150 & 300 & 60 & 10767/1692 & 437.502 & 105.565\\
\bottomrule
\end{tabular}
\end{table*}

Full captures 305 frames versus always's 300: fewer repairs need not mean fewer
screenshots. Every policy makes sixty model calls and uses 10,767 input/1,692
output tokens. No token saving is observed, because OCR re-grounding preserves
the parsed intention. The shared TEST model/OCR load is 0.836/0.129s.
Full/conjunction gate wall totals are 0.05390/0.04062s;
CPU totals are 0.08368/0.06538s.
Instrumented loops include screenshots, OCR/planning, forced settling, hit tests,
native scoring, logging and execution. Shared-host timing is descriptive,
not whole-agent CPU utilization, energy or production-only latency.

\paragraph{Mechanistic records and scope.}
Evaluator-only hit tests label actual checkbox dispatch positions. In full's
scroll condition, fifteen initially invalid checkbox proposals become valid
and the corresponding native goals complete. The first witnesses by the frozen
goal/condition/step order are Mesa Amber, Mesa Birch and Mesa Delta at step zero.
The complete labeled-event table remains with the raw records, including all
other policies. Full has 75 labeled initial checkbox proposals across conditions;
its sixty initially valid proposals are not refreshed. Other action types remain
unlabeled by this hit test; absence of a recorded wrong checkbox is not a general
semantic-validity or safety guarantee. These controlled witnesses are not
independent counterfactual proof.

Independent recount reconstructs the full planned matrix, all native outcomes,
extra state changes (including other recorded applications), action and resource
counts, image hashes, 320 contract checks and 170 random draws. Full native initial
states and first-source image bytes match across all five policies in every
scenario. All fourteen frozen source/plan files and the validity core retain
their recorded hashes. Analysis implementation is tracked separately and was
hardened while TEST ran; these outcomes remain exploratory.

Goal IDs/text are disjoint across stages, but smoke and the first two TEST goals
reuse background seeds 10300/10301 with different names. No unseen-background
claim follows. These twenty goals reuse one application's primitives; they do
not establish natural workflow, long-horizon or unseen-family transfer. Composed
success is the native final-state predicate, not independent verification of
ordered execution. Perturbations precede the final dispatch check, so concurrent
changes after that check remain untested. No physical-safety, source-attestation,
token-saving or unique efficacy claim is added by this experiment.

\paragraph{Historical evidence is not renewed by retrieval.}
A separate TRAIN-only feasibility audit pairs cached regions within video and
category. Among 45 pairs retaining a unique reference identity, eight cached
regions remain correct at IoU $\geq0.5$ (six videos), whereas 37 do not
(22 videos); 34 of the latter were correct in the source frame. These sparsely
selected pairs diagnose potential reuse failure, not its natural prevalence.
Frame-index gaps are not measured dispatch latency, and category multiplicity
does not establish identity ambiguity. We therefore do not train a deployment
controller or claim memory-related efficacy from this audit. A bounded prototype
keeps observation and availability times separate and returns unknown for
expired or conflicting evidence; its software checks are not task-success tests.

\paragraph{Candidate ranking: lower loss need not recover more targets.}
A separate development-only diagnostic trains shared candidate scorers on 116
unique-visible TRAIN cases (42 videos) and selects checkpoints on 45 VAL cases
(14 videos). A nine-parameter linear head and a 161-parameter MLP receive the
same eight score/area/overlap/rank features. No CAL or TEST is read. With a
3,000-epoch cap and the prespecified primary seed, native top-1 and the MLP each
select 42/45 correct regions; the linear head selects 39/45. Raw top-five supply
is 44/45. The MLP lowers candidate BCE (0.205 versus 0.288 for linear) without
rescuing a native error; all three MLP sensitivity seeds match native correctness.
The initial 300-epoch run is retained: all checkpoints selected its final epoch,
motivating the separate longer-budget diagnostic. No converged optimum is claimed
for every linear fit. Unique-visible forced ranking does not test rejection of
absent or ambiguous targets, independent generalization or task completion.
Changing both the fit and checkpoint criterion to pairwise ranking gives 42/45
for every linear and MLP seed; a regularized convex linear reference also gives
42/45. This is objective-alignment development, not an isolated causal effect
of the loss or an improvement over native ranking. No native miss is rescued.

\section*{S32. Dense cached-evidence feasibility and repair boundaries}
\label{suppidx:32}
\label{sec:dense-memory}

\paragraph{Question and source isolation.}
This exploratory follow-up asks whether a small checker can identify reusable
cached regions beyond strong simple rules. It is separate from the locked
candidate-recovery TEST and does not change its models or cutoffs. From the
SA-Co SmartGlasses public validation release, the study fixes 32 TRAIN, 12 VAL,
16 CAL and 30 TEST source videos, with no video shared between splits. CAL and
TEST sources were unused in the preceding candidate allocation. Participant
independence is not certified. At most two eight-index clips per video/category
are selected: a hash-seeded unique-start clip and, when available, a
transition-enriched clip. Thus this is not a natural-prevalence sample.
The acquisition contains 1,202 distinct frames and 1,162 source--current pairs.
Indices give ordering, not measured exposure-to-dispatch age.

\paragraph{Fixed inputs, controls and selection.}
Frozen SAM3 supplies the source top-1 region. Reuse is correct only when the
current reference contains a unique matching annotation identity and the cached
region has IoU $\geq0.5$. The mask stays at fixed image coordinates: no
camera-motion compensation or 3D reprojection is tested. Reference-unobserved is not certified physical absence;
category multiplicity is not certified identity ambiguity. The declared consumer
contract rejects category-multiple cases. References never enter online inputs.
A 1,035-parameter linear head and a 33,153-parameter MLP receive identical
1,034-dimensional inputs: ten source/photometric/index scalars and frozen
ResNet-18 global/crop feature differences. Three seed fits use source-weighted
TRAIN loss and VAL checkpoint selection; the primary MLP seed is fixed in
advance. Fixed-model evaluation on TRAIN is recorded separately from update-time
loss. Controls include confidence, ZNCC, RGB, combinations, linear heads and
regularized convex linear fits. A separately locked addendum includes depth
1--3 decision trees on the same inputs. CAL fixes separate empirical 5\% error
targets for incorrect unique, reference-unobserved and category-multiple cases.
This is an empirical working point, not a deployment risk guarantee.

\begin{table}[h]
\centering\footnotesize\setlength{\tabcolsep}{3pt}
\begin{tabular}{@{}lrrr@{}}
\toprule
Fixed rule & \shortstack{Cached\\accepts} & Correct & Invalid\\
\midrule
\shortstack[l]{VAL-selected primary\\simple (RGB)} & 1 & 1 & 0\\
Primary MLP & 0 & 0 & 0\\
\shortstack[l]{VAL-selected addendum\\(depth-2 tree)} & 0 & 0 & 0\\
\bottomrule
\end{tabular}
\caption{Dense TEST: 399 pairs from 30 new source videos. Zero invalid accepts
with zero coverage is abstention, not demonstrated useful reliability. The
addendum uses the same cohort, not an independent replication.}
\label{tab:dense-memory}
\end{table}

\paragraph{Outcome and supply diagnosis.}
The primary MLP CAL cutoff is infinite: it rejects every TEST cache proposal.
RGB correctly accepts one pair; its source-macro correct-adoption rate is 0.285\%.
MLP minus RGB is $-0.285$ percentage points (paired 95\% interval
$[-0.923,0]$). The shallow-tree addendum also rejects every pair. Degenerate
zero-error differences do not demonstrate equivalence or noninferiority.
An independent recount verifies all split identities, feature receipts and
stored decisions. In TEST, source top-1 is initially correct for 35/57 clips.
Only 9/399 cached regions remain correct, across five videos; even among the
245 pairs with initially correct source regions, only nine remain reusable.
CAL has just 11 reusable pairs from two videos. Conditioning on source-reference
correctness is a post-hoc diagnosis, not an admissible online oracle. The trial
therefore provides no learned adoption advantage; its scarce, rapidly changing
positive supply cannot support a broad efficacy or information-limit conclusion.

\paragraph{Rejection does not validate the repair branch.}
A post-hoc static-frame composition audits the following explicit repair rule:
reuse when the fixed checker accepts, otherwise rerun frozen SAM3 and adopt
only when its default output has exactly one mask. Let $R$ denote reuse,
$A_f$ fresh adoption, and $E_o,E_f$ invalid old/fresh regions. The final invalid
adoption event splits into disjoint branches:
\[
 P(E)=P(R\cap E_o)+P(R^c\cap A_f\cap E_f).
\]
This is event bookkeeping, not a novel safety theorem. A cache-branch constraint
does not bound the second term. Always refreshing and MLP rejection both yield
91 adopted regions: 48 reference-correct and 43 invalid under the declared
contract (eight incorrect unique regions, three reference-unobserved, and 32
category-multiple cases). RGB skips one refresh but has the same totals.
These are offline region decisions, not executed actions, physical failures or
task-success scores. Fresh default acceptance is not an optimized repair policy.
Summed prerecorded SAM stage times cannot establish online cost savings.

\paragraph{Resource and audit scope.}
The full frozen visual frontend reaches approximately 4.27\,GiB peak CUDA
allocation; a tiny adoption head does not make the entire stack a 1\%-CPU
system. Resource samples separate waiting and active stages, and summed process
RSS is not private memory. Raw frames, masks, weights and learning curves remain
on the experiment host; repository summaries retain hashes and split receipts.
Two pre-inference wrapper revisions are retained as engineering history, not
additional scientific trials. The current evidence motivates qualifying both
the cached source and fresh repair path before another task-benefit study.

\section*{S33. Bounded executor qualification and an ineffective input contrast}
\label{suppidx:33}
\label{sec:executor-input}

This development-only follow-up reuses the authored text, image and slide
workflows from S30. It is not an official OSWorld task score or an independent
generalization test. All four arms use the same frozen Qwen3-VL-4B-Instruct
checkpoint, procedural guidance, repaired exact F2 action support, application
setup and saved-output predicates. Each has a 192-token response cap, 30 action
steps per case and a 900-second per-case wall cap. The admission gate requires
all three native outputs to be correct and an explicit completion response.
No model self-report substitutes for a saved-output label.

The intended $2\times2$ contrast changes configured pixel caps
\{786432,1572864\} and constrained JSON decoding \{off,on\}. All arms finish
with clean teardown; text passes in four arms, image and slide in none.
Constrained decoding avoids the two unconstrained slide JSON parsing failures,
but its saved slides are still incorrect. There are 119 model calls in total
and no response reaches the output cap. The four arms are not independent
task samples and must not be pooled into a benchmark success-rate estimate.
Independent saved-artifact scoring and every source-frame hash agree with the
recorded reports. No actor passes admission; no nominal TWS-V efficacy batch
is run on these unqualified configurations.

Crucially, both configured pixel caps produce the same actual visual grid,
$[1,68,120]$. The intended resolution treatment therefore fails its manipulation
check. This does not show that resolution is ineffective. Configuration flags
alone are not evidence that a model received a different input. The initial
launcher also failed before inference because an optional dependency path
shadowed the host's binary image library; that attempt is retained as an
infrastructure failure, not counted as a task zero. The corrected launcher
isolates optional worker dependencies and leaves all task predicates unchanged.
These diagnostics localize implementation and actor-competence gaps; they add
no natural end-to-end, safety or token-saving claim.

\paragraph{Prospective correction, retained negative result.}
A no-model preprocessing check identifies the effective fast-processor size
field. A separately frozen A3 changes that field and records actual input grids;
it preserves the model, prompt, action space, budgets and native predicates.
The two caps now produce $[1,40,72]$ and $[1,58,104]$, verifying an actual input
contrast. All four corrected variants still pass text and fail image/slide.
Their 117 model calls consume 194,152 input and 2,295 output tokens, with 595.9s
summed profile wall time. These are qualification costs, not useful token savings.
All saved-output scores and frame hashes pass the independent recount, and
teardown is clean. Neither profile selection nor TWS-V efficacy is justified:
the actor remains unqualified on the original three-workflow admission gate.
The reused DEV instances and four configurations are not independent task trials.

\paragraph{Later excluded desktop transport smoke.}
A separate frontier-model transport smoke on excluded OSWorld task 052 is not
an admitted qualification trial. Retrospective inspection of the saved initial
browser state finds a task-site error page; later blank or unrelated pages do
not establish task readiness. Its pointer labels therefore cannot establish
natural task-relevant unsupported-action opportunities. The retained later model
timeout does not turn this infrastructure-ineligible attempt into a clean-actor
failure. The original report is preserved alongside the fixture audit, and no
20-task qualification or intervention comparison is inferred from this smoke.
A passive bracket-capture follow-up also encountered an error-page fixture;
its timing measurements do not calibrate a task-validity threshold. These
failures are distinct from the completed simulator qualification episodes in
S39, whose recorded infrastructure was audited separately.

\section*{S34. Fresh-proposal validity and complete-repair development}
\label{suppidx:34}

\paragraph{Scope and positive supply.}
Following S32, this exploratory extension opens only the existing TRAIN and VAL
sources, without reading CAL/TEST labels or features. The new-proposal label has
126 positives from 21 TRAIN videos and 59 from ten VAL videos, compared with
32/17 reusable cached regions. All 553 freshly recomputed top-1 masks and native
candidate counts match the prior extraction exactly. Grouping remains by video;
reference-unobserved and category-multiple retain the limitations stated in S32.
These are component labels, not native task-success outcomes.

\paragraph{Features and training.}
Fresh confidence, candidate count/margin/geometry, source-to-fresh crop RGB/ZNCC
and frozen ResNet18 differences enter only \emph{after} re-grounding. They are
not available to the preceding refresh decision. Same-input linear and width-32
MLP heads use 1,039 features and fixed source-weighted or source/class-balanced
TRAIN loss. VAL keeps the same source-weighted BCE checkpoint criterion.
Three regularized convex linear fits remain strong controls. The fixed
600-epoch/60-patience budget and primary seed are not extended after outcomes.
VAL also sets descriptive separate 5\% error working points; this reuse of VAL
does not constitute held-out calibration or evaluation.

\begin{table}[h]
\centering\footnotesize\setlength{\tabcolsep}{3pt}
\begin{tabular}{@{}lrrr@{}}
\toprule
VAL rule & Correct & \shortstack{Unobs.\\error} & \shortstack{Multiple\\error}\\
\midrule
Native default &61.40&14.29&30.27\\
Confidence + uniqueness &14.47&0.00&4.42\\
Full-feature MLP &2.31&1.10&0.00\\
Class-balanced MLP &7.35&4.76&0.00\\
\shortstack[l]{Scalar-only MLP\\(follow-up)} &0.77&0.00&0.00\\
\bottomrule
\end{tabular}
\caption{147 VAL pairs from 12 videos, source-macro percentages within each
reference state. Wrong-unique error is 16.67\% for native default and zero for
these selected working points. Checkpoints and cutoffs use this same VAL;
no interval or independent-superiority claim is made.}
\end{table}

The primary learned head does not exceed the simple confidence/uniqueness
conjunction. A separately declared post-hoc 15-scalar tier also retains linear,
MLP and three ridge fits. Its MLP lowers VAL BCE from 0.631 to 0.404 while
correct adoption falls from 2.31\% to 0.77\% at the constrained working point.
Lower average loss therefore does not establish better constrained adoption.
Input dimension and parameter count both change in this tier; it is not an
isolated causal representation effect. No model or threshold is promoted to
deployment or rescored on the old TEST.

\paragraph{Fresh rejection has a measurable acceptance cost.}
A post-training VAL-only routing replay allows one fresh request and assumes
healthy source/time/action contracts for the component comparison; actual
exposure, readiness and dispatch times are unknown. Old RGB followed by default
fresh acceptance yields 49 correct and 16 invalid region adoptions. Requiring
the selected simple fresh check yields 21 correct and three invalid adoptions;
the MLP yields 13 and two. The three remaining simple-chain errors include one
old-branch and two fresh-branch errors. These are acceptance/error trade-offs,
not demonstrated task-success improvements. All fresh outputs were precomputed,
so request counts cannot establish measured online or token savings.

A bounded state-machine implementation requires separate contract and support
checks for each fresh proposal, preserves unknowns and refuses premature fresh
results or reused observation identities. Eight focused software tests pass.
Independent Boolean replay agrees on all six policy/count tables. Independent
model reloads reproduce all seven primary and five scalar fit outputs; label,
pair identity, source-disjointness and scalar-score recounts pass. Frozen visual
embeddings are not independently re-extracted. A complete native efficacy study
still requires qualified repair/execution and new calibration/test sources.

\paragraph{Separating refresh selection from fresh acceptance.}
A post-hoc $2\times2$ VAL diagnostic crosses the old-RGB adaptive refresh
trigger with an independent random trigger, and native acceptance with the
selected simple fresh check. The random probability is $136/147$, matched to
the adaptive request count on this VAL \emph{in expectation}; this is not
independent CAL-budget matching. Non-refreshed proposals keep the same old
support gate in all arms. Exact integration over the random branches avoids
selecting favorable random seeds.

\begin{table}[h]
\centering\small
\begin{tabular}{llrrr}
\toprule
Trigger & Fresh rule & Correct & Invalid & Abstain\\
\midrule
Adaptive & Native &49&16&82\\
Random & Native &45.16&13.95&87.89\\
Adaptive & Simple &21&3&123\\
Random & Simple &16.48&1.93&128.60\\
\bottomrule
\end{tabular}
\caption{Exploratory component replay on the same 147 VAL pairs. Each arm
requests 136 fresh outputs in expectation. Random rows are expected counts,
not executed episodes. The shared candidates, cutoffs and old support gate
are fixed; no new fitting or CAL/TEST reading is performed.}
\end{table}

Adaptive selection increases both correct and invalid adoptions relative to
random refresh. The result does not establish an error-matched advantage.
Before adoption gating, 65 unique-visible VAL pairs contain 17 cases where
both candidates are correct, 42 where only the fresh candidate is correct,
and six where neither is correct. TRAIN additionally contains one case where
refreshing changes a correct candidate to an incorrect one. Freshness therefore
does not guarantee semantic validity, and a failed single fresh proposal does
not establish that further recovery is impossible. Independent scalar branch
enumeration reproduces all six policies, including always-fresh controls.
The next efficacy evaluation must calibrate the complete routed policy rather
than infer its error bound from separate per-head working points. This analysis
measures neither task completion nor online compute/token savings.

\paragraph{Independent support cannot be manufactured from repeated frames.}
A subsequent planning audit aggregates the \emph{complete route}: whether a
video has any invalid adoption within a reference state. This conservative
group-level endpoint differs from frame-level error and selective risk.
For a fixed policy and independent Bernoulli groups, zero observed errors give
the one-sided exact upper limit $U=1-\alpha^{1/n}$. At $\alpha=.05$, at least
59 zero-error groups are required for $U\leq .05$; a four-policy/two-state
Bonferroni allocation instead requires 99 per state. This is a sample-support
diagnostic, not a new theorem or a guarantee for our VAL-selected policies.
Current VAL contains only seven videos in each of the reference-unobserved and
category-multiple strata. Video independence beyond the grouping assumption is
not established. Nine focused software checks and eight independent numerical
comparisons pass, without adding empirical samples.

A metadata-only inventory reserves 96 remaining videos from the same official
source table, split into 48 CAL and 48 TEST, disjoint from 238 previously
allocated videos. Subsequent reference-only construction under the frozen
size/frame rules yields no unique-visible CAL cases and no eligible TEST cases;
only two annotated-gap and thirteen category-multiple CAL cases remain.
No new media acquisition or model inference is performed for this failed
reservation, and it cannot support an efficacy comparison. Source dispositions
are retained rather than relaxing eligibility to manufacture positives.
Even 48 zero-error independent CAL videos would not certify the conservative
5\% group-error target above. A separate, reference-enriched cohort from the
published official TEST release is therefore a new descriptive component study,
not a continuation that silently replaces these unavailable TEST outcomes;
it does not establish natural prevalence or participant independence.

\paragraph{LIF spiking-head follow-up.}
A separately frozen pilot compares ReLU, leaky
integrate-and-fire (LIF) and non-spiking leaky heads using the general
surrogate-gradient formulation~\cite{snntorch}. Each tier has identical inputs,
width 32, paired initial weights and trainable parameter counts (545 for 15
inputs; 33,313 for 1,039). Fixed LIF settings are $\beta=.9$, threshold 1,
subtractive reset, fast-sigmoid surrogate slope 25 and eight computational steps.
The same static vector is supplied throughout those steps; these are not eight
video frames. State resets for each example. Three paired seeds use the same
training/checkpoint budget; they are not independent test samples.

On the primary 15-input run, LIF correctly adopts three VAL pairs (source-macro
2.37\%) versus ReLU's one (0.77\%), with no observed invalid adoption for either.
A simple confidence/crop-similarity rule adopts 17 correctly with no observed
invalid adoption on the same VAL. The other two scalar LIF seeds increase
correct adoption over paired ReLU but also admit category-multiple cases.
The primary 1,039-input LIF and ReLU each adopt three correct and one invalid
pair; LIF has no stable advantage across seeds/tiers. All 18 checkpoint outputs
and state resets replay, and primary ReLU reproduces the earlier ANN fits.
Six focused neuron/gradient/initialization tests pass.

Warm CPU batch-one model-forward medians are 0.099\,ms for scalar LIF versus
0.009\,ms for scalar ReLU on this host, excluding the visual frontend. Spikes
do not establish energy savings; no power or neuromorphic measurement is made.
The 15-input tier still includes frozen-encoder cosine features and therefore
does not eliminate ResNet inference. No independent generalization, useful token
saving or native-task benefit follows from this development pilot.

\section*{S35. Separating cached applicability from replacement support}
\label{suppidx:35}

\paragraph{Two checks, different available evidence.}
Let $z_0$ denote inputs ready before a refresh decision and $z_j$ inputs ready
after fresh grounding $j$. Fresh-candidate scores cannot enter the earlier
refresh trigger. For one action decision, let $D_j$ indicate that candidate $j$
is actually dispatched ($j=0$ denotes reuse), and let $Y_j$ denote its
evaluator-only validity. At most one candidate is dispatched, so the exact
invalid-dispatch and fresh-grounding counts are
\begin{equation}
 E=\sum_{j=0}^{K}D_j(1-Y_j),\qquad
 C_{\mathrm{ground}}=\sum_{j=1}^{K}R_j,
\end{equation}
where $R_j$ indicates that fresh grounding $j$ was requested, including rejected
or terminal outputs. Grounding, screenshots, extra OCR, model calls/tokens and
wall time are separate costs. Two required evidentiary claims do not require
two neural networks or duplicated feature extraction. A recorder-verified
byte-identical current image can reuse already-ready OCR; that is a host cache
optimization, not proof of sensor authenticity or correct action semantics.

Neither a rejected cached proposal nor a newly captured image establishes
$Y_j=1$. Separate per-head empirical cutoffs also do not bound the complete
adaptive route without additional assumptions. Reference labels above are used
only for evaluation, never to select an online target or repair. For multi-action
tasks, interventions change subsequent states; we therefore execute each policy
from its reset rather than infer task success by adding offline candidate labels.

\paragraph{Prospective controlled design.}
The fixed local Qwen3-VL-2B intention parser, current-image OCR and native
OpenApps evaluators are retained. The actor is a hybrid system, not an autonomous
frontier visual policy. A separate admission stage requires eight new positive
goals to complete without unintended changes and eight missing/duplicate goals
to abstain with the full native state unchanged. Failed admission blocks CAL
and TEST. A source/code lock precedes these stages; old results and thresholds
are not overwritten.

The R2 TEST crosses 24 new related goal instances, four imposed conditions and
eight controllers (768 executed episodes). Conditions are unchanged, a 76-pixel
scroll, a received context-ID mismatch, and that mismatch plus a first-fresh
coordinate offset. The latter changes the resolver's first fresh checkbox
proposal by 95 normalized vertical units before producer sealing. It is a
controlled faulty resolver adapter, not an observed natural hallucination.
A context mismatch is contract ineligibility, not a semantic wrong-action label.
The old scroll miss found in excluded smoke is retained rather than tuning its
ZNCC cutoff. All arms share six extra groundings, two per decision and sixteen
steps; actual repair costs need not match despite equal caps. A CAL-derived
random trigger matches initial request frequency, not realized TEST cost.

Comparators cross adaptive/random refresh with native/checked fresh acceptance,
and include no check, unconditional fresh checking, a blind extra grounding and
the same contract/OCR predicates evaluated by a scalar implementation. The fresh
check asks whether the actor's declared target is uniquely supported by current
OCR/checkbox geometry within 12 normalized coordinate units, followed by contract
revalidation. It cannot verify the parsed intention or hidden native handler.
Missing/ambiguous grounding abstentions in the shared actor are not weakened.

Clean completion requires the unchanged native goal evaluator to pass, no
unintended native state change, and no evaluator-confirmed invalid checkbox
dispatch. Native completion and erroneous actions are also reported separately.
Paired exploratory intervals resample related goal instances, not independent
application families. Equal observed outputs do not establish equivalence.

\paragraph{Complete-batch result.}
All 768 planned episodes completed, and an independent saved-evidence recount
replayed 2,251 contract checks and 419 semantic decisions. Initial native states
and source-image hashes matched across all policy arms. Table~\ref{tab:double-repair}
aggregates the 24 goal instances for each policy across four conditions. In the
combined context-mismatch and faulty-fresh condition, native fresh acceptance
completed only 6/24 episodes and made 18 invalid checkbox dispatches. Checking
the replacement candidate completed 24/24 with no invalid dispatch, even though
the unchanged cached action would also have completed all 24. Thus a repair
trigger can make an otherwise successful route worse when replacement support
is not checked.

Across all conditions, checked adaptive repair completed 96/96 episodes versus
78/96 for adaptive repair with native fresh acceptance, an absolute difference
of .188 (related-goal bootstrap 95\% descriptive interval [.146,.229]). The same
scalar contract/OCR predicates also completed 96/96 with the same number of
groundings and OCR checks. Blindly requesting one additional candidate likewise
completed 96/96, but required 1.375 rather than .938 fresh groundings per
episode; always-checked repair required 2.000. The parser was unchanged, so every
arm used one model call per episode and had identical recorded input/output token
counts. Checked adaptive repair measured 139\,ms of gate wall time and
449.5\,ms of process CPU time per episode. The CPU timer accumulates time
across threads, including the multithreaded OCR runtime. The scalar
implementation also measured 139\,ms of gate wall time per episode. These
episode-level measurements include OCR support checks when invoked. S23's
0.16\,ms per-call figure measures the isolated guard without OCR. The
identical-predicate scalar result bounds the incremental accuracy claim.

\begin{table}[t]
\centering
\caption{R2 controlled repair chain, aggregated over four conditions. Costs are
per episode. ``Invalid'' counts evaluator-confirmed invalid checkbox dispatches.}
\label{tab:double-repair}
\footnotesize
\setlength{\tabcolsep}{3pt}
\begin{tabular}{@{}lrrrr@{}}
\toprule
\parbox[t]{.39\linewidth}{\raggedright Policy} & Clean & Invalid & \shortstack{Fresh\\groundings} & \shortstack{Extra\\OCR}\\
\midrule
\parbox[t]{.39\linewidth}{\raggedright No check} & 78/96 & 18 & 0.000 & 0.000\\
\parbox[t]{.39\linewidth}{\raggedright Adaptive / native fresh} & 78/96 & 18 & 0.688 & 0.000\\
\parbox[t]{.39\linewidth}{\raggedright Adaptive, checked fresh} & \textbf{96/96} & 0 & 0.938 & 0.188\\
\parbox[t]{.39\linewidth}{\raggedright Same predicates, scalar} & \textbf{96/96} & 0 & 0.938 & 0.188\\
\parbox[t]{.39\linewidth}{\raggedright Blind extra grounding} & \textbf{96/96} & 0 & 1.375 & 0.000\\
\parbox[t]{.39\linewidth}{\raggedright Always checked} & \textbf{96/96} & 0 & 2.000 & 0.250\\
\bottomrule
\end{tabular}
\end{table}

\begin{figure*}[t]
\centering
\includegraphics[width=\textwidth]{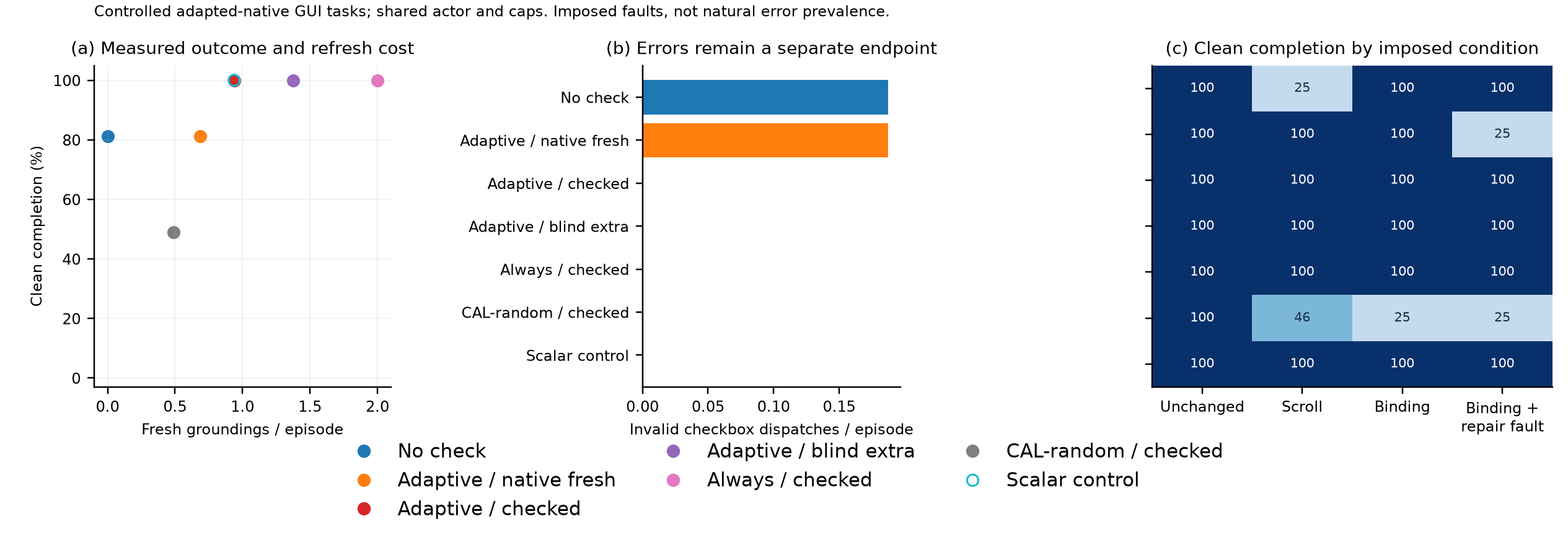}
\caption{Executed R2 outcomes and measured repair cost. Panel (a) shows that
checked adaptive repair, the same-predicate scalar control and less selective
repair policies all reach the same clean-completion endpoint at different
grounding costs. Panel (b) keeps invalid dispatches separate from completion.
Panel (c) follows the same policy order as panel (b) and localizes failures:
the cached scroll condition defeats no-check,
whereas accepting the controlled faulty fresh proposal defeats native repair.
Conditions are imposed mechanisms, not estimates of natural prevalence.}
\label{fig:double-repair-r2}
\end{figure*}

The recount did not re-execute the OCR recognizer; it verified saved recognizer
outputs, evidence hashes, candidate geometry, native state outcomes, controlled
offsets and actual dispatch routes. Related goal instances are not independent
application families, and zero observed errors is not an equivalence or safety
result. No running TEST outcome was used to alter a threshold or policy.

\paragraph{Comparator-specific costs (post-hoc recount).}
The blind-extra control shares the old-evidence trigger; it is blind to fresh
candidate support, not a fully signal-independent refresh policy. Across R2,
adaptive / native fresh, complete, same-predicate, blind-extra and always-checked arms use
66, 90, 90, 132 and 192 re-groundings, respectively. Thus complete repair uses
36.4\% more groundings than adaptive / native fresh, while using 31.8\% fewer than blind-extra
at the same observed completion endpoint. Equal budget caps do not imply equal
realized costs. These observations do not meet the later S36 completion-gain
criterion against the identical-predicate control, nor its +10\% grounding
margin against adaptive / native fresh. They do not retroactively change the R2 protocol.

\section*{S36. Prespecified independent closed-loop criterion}
\label{suppidx:36}
\label{sec:independent-closed-loop}

The controlled repair studies above identify mechanisms; they do not establish
an intervention benefit on independent workflows.  Before collecting the next
workflow cohort, we therefore choose one primary objective and do not switch to
a cost-saving objective after observing TEST:
\begin{quote}
At comparable realized complete re-grounding cost, improve paired clean task
completion while keeping absolute invalid dispatch below a prespecified limit.
\end{quote}
An actor--environment pair is admitted only if a separate qualification set
shows at least 70\% clean completion, enough naturally occurring unsupported
proposals to make intervention identifiable, deterministic reset or an
independently checkable outcome, complete dispatch/cost logging, and no fixture
failure counted as agent failure.  Failure of admission stops the efficacy
study rather than triggering actor or task replacement.

The four frozen arms use the same actor, instructions, context, repair cap and
wall-time cap: blind fresh-observation/retry, the strongest TEST-blind simple
validator, the complete TWS-V cached-and-replacement chain, and always-fresh as
a cost anchor.  Task/reset blocks randomize arm order.  The primary benefit is
the paired clean-completion difference against the strongest eligible
comparator.  Success additionally requires a one-sided 95\% upper confidence
bound of at most 5\% on TWS-V's absolute invalid-dispatch rate and no more than
10\% greater realized complete re-groundings per episode than that comparator.
Captures, OCR, model calls/tokens and latency remain separate costs.

The 70\%, 5\% and 10\% values are development working points, not safety or
industry standards.  A task owner may replace them once, before the new cohort,
using an explicit action-error and refresh-cost specification; they cannot be
changed after qualification or TEST.  Cohort size is fixed from a minimally
relevant paired-completion effect and the harm-bound precision before TEST.  For
orientation, even zero invalid dispatches require at least 59 independent
blocks for the one-sided exact 95\% upper bound to fall below 5\%; dependence or
observed errors require more.  New sessions or task families are used, natural
and imposed changes are separate strata, and an inconclusive harm bound is
reported as inconclusive rather than safe.

\section*{S37. Target-context repair and strong controls (R3)}
\label{suppidx:37}
\label{sec:context-repair-r3}

R2 showed that an unchecked fresh replacement can be wrong. R3 asks whether a
richer target-context rule supplies an outcome advantage beyond the strongest
simple controls. Before TEST, we froze 12 new goal instances, five imposed
conditions and nine policies. All policies share the actor, action and model-call
budgets; TEST contains 540 episodes. An independent saved-evidence recount
replays native outcomes, 420 semantic decisions and 573 fresh events. It does
not re-execute the recognizer, so this is an outcome and accounting check rather
than an independent perception replication.

\begin{table}[t]
\centering
\small
\setlength{\tabcolsep}{2.5pt}
\caption{Frozen R3 TEST totals. ``Checked'' means that the replacement is
validated before dispatch. All policies use 60 model calls and 12,675 tokens;
no token saving is inferred.}
\label{tab:context-repair-r3}
\begin{tabular}{@{}lrrrrr@{}}
\toprule
\parbox[t]{.36\linewidth}{\raggedright Policy} & Clean & Invalid & \shortstack{Re-\\fresh} & OCR & \shortstack{Time\\(s)}\\
\midrule
\parbox[t]{.36\linewidth}{\raggedright Adaptive / checked} & 60/60 & 0 & 45 & 9 & 373.2\\
\parbox[t]{.36\linewidth}{\raggedright Context / native} & 51/60 & 9 & 36 & 48 & 418.1\\
\parbox[t]{.36\linewidth}{\raggedright Context / checked} & 60/60 & 0 & 48 & 51 & 409.1\\
\parbox[t]{.36\linewidth}{\raggedright Same-predicate scalar} & 60/60 & 0 & 48 & 51 & 411.7\\
\parbox[t]{.36\linewidth}{\raggedright CAL-NCC / native} & 51/60 & 9 & 45 & 0 & 389.4\\
\parbox[t]{.36\linewidth}{\raggedright CAL-NCC / checked} & 60/60 & 0 & 57 & 9 & 387.6\\
\parbox[t]{.36\linewidth}{\raggedright Hash change / checked} & 60/60 & 0 & 105 & 12 & 432.3\\
\parbox[t]{.36\linewidth}{\raggedright Always / checked} & 60/60 & 0 & 117 & 12 & 441.0\\
\parbox[t]{.36\linewidth}{\raggedright Context / blind extra} & 60/60 & 0 & 72 & 39 & 425.5\\
\bottomrule
\end{tabular}
\end{table}

\begin{figure*}[t]
\centering
\includegraphics[width=\textwidth]{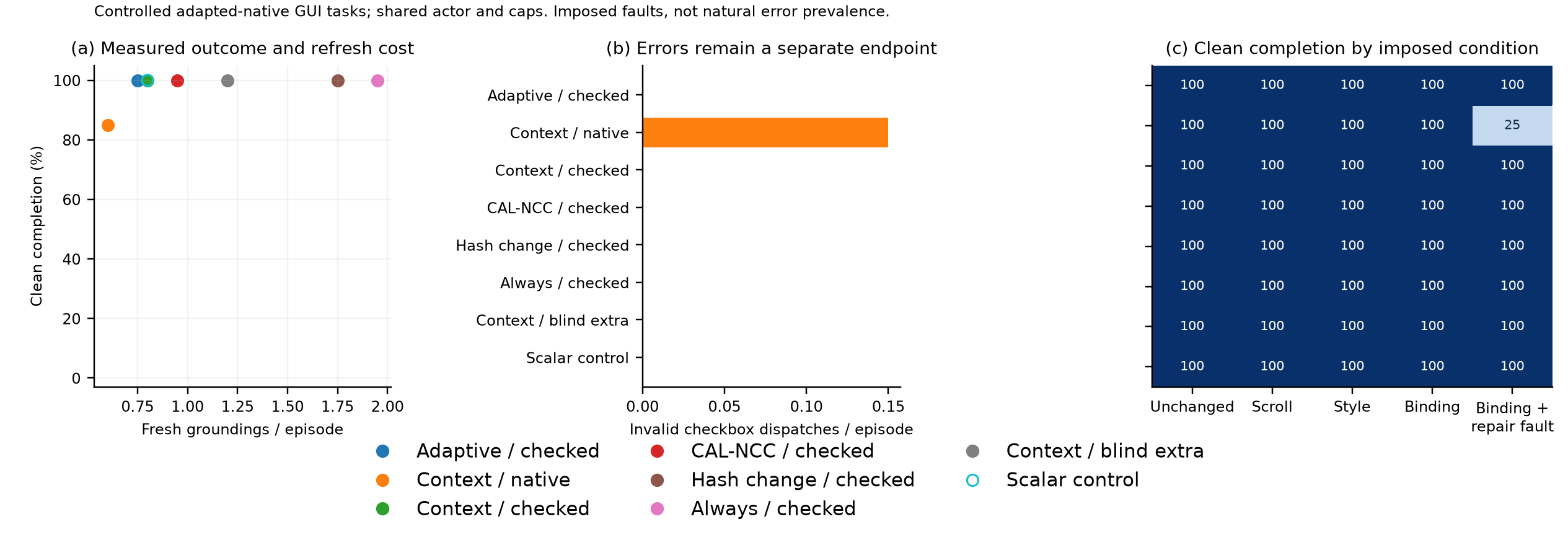}
\caption{\textbf{Outcome and cost must be read together.} (a) Several checked
policies reach the same clean-completion endpoint at different refresh costs.
(b) Native replacement acceptance leaves nine invalid dispatches in the
repair-fault condition; checked variants prevent them. (c) Rows follow the
policy order in (b). The largest difference occurs under the imposed
binding-plus-repair fault. Conditions reuse related goal instances.}
\label{fig:context-repair-r3}
\end{figure*}

Under the imposed repair fault, checking the fresh target-context proposal
changes clean completion from 51/60 to 60/60 and invalid checkbox dispatches
from nine to zero. The same change appears for calibrated NCC. Thus the
measured benefit belongs to the second validation boundary---checking the
replacement---rather than uniquely to the richer context score. Every checked
policy reaches 60/60 clean completion with no invalid dispatch. In particular,
the full context rule and an identical-predicate scalar rule produce the same
actions, while the older adaptive checked policy reaches the same endpoint
with fewer extra OCR calls. Hash-change and always-check controls are also
correct but refresh substantially more often.

The result strengthens the repair-chain claim while narrowing the method claim:
cached-evidence invalidation and replacement validation are distinct tests, but
this controlled cohort does not show unique target-context accuracy, natural
generalization, token savings or noninferiority. Zero observed discordance is
not an equivalence result, and the imposed conditions do not estimate field
failure frequency.

\paragraph{Separate cost axes.}
A post-hoc complete-grid recount retains all nine controllers and five conditions.
Full context validation uses 48 re-groundings and 51 extra OCR calls, versus
72 and 39 for trigger-matched blind-extra repair and 117 and 12 for always-checked
repair. Fewer re-groundings therefore need not mean fewer OCR operations.
Recorded parser tokens are identical across these arms. The corresponding
empirical episode-loop 95th percentiles are 10.68, 10.82 and 12.13 seconds;
these are descriptive timings on related controlled fixtures, not deployment
latency guarantees. The same-predicate scalar rule uses the same 48 groundings
and 51 OCR calls as full validation. No unique method or token-saving advantage
follows from this cost comparison.

\section*{S38. Frozen published-source candidate adoption}
\label{suppidx:38}
\label{sec:published-candidate-test}

We add a prospectively locked, source-name-held-out component test using
published SACo-VEval media. This is an additional population, not a replacement
for the 491-case study in Sec.~S26. Before opening these TEST outputs, VAL chose
\texttt{product\_margin} as the strong transparent comparator and
\texttt{mlp\_set\_ce\_2026111403\_adoption} as the learned head. The split,
models, thresholds, source list and a joint outcome--error criterion were then
locked. TEST contains 458 cases from 64 videos and 432 distinct frames: 185
unique-visible, 134 annotation-unobserved and 139 multiple-category cases.
The latter two names do not certify physical absence or same-instance identity
ambiguity.

The first archive stream ended in an external timeout. The retained recovery
replayed the gzip archive from byte zero and reused only 43 prior files whose
size and SHA-256 matched. Before inference, an independent transport reader
verified all 432 prescribed paths, JPEG start markers,
per-file hashes and the locked case/plan hashes. The subsequent saved-output
reader independently recounts 2,290 masks, 38 policies, IoU-derived labels,
source-macro rates and rescue/loss events; it does not re-execute SAM.

\begin{table}[t]
\centering\scriptsize
\setlength{\tabcolsep}{2.6pt}
\caption{Frozen published-source TEST. Percentages are macro-averaged over
source videos within each named stratum. The primary learned head is compared
only with the preselected transparent rule.}
\label{tab:published-candidate-test}
\begin{tabular}{@{}lrrrr@{}}
\toprule
Rule & Correct & \shortstack{Unique\\err.} & \shortstack{Unobs.\\err.} & \shortstack{Multiple\\err.}\\
\midrule
Product--margin &50.65&0.39&3.33&7.39\\
Frozen MLP &22.14&3.13&0.00&17.80\\
\bottomrule
\end{tabular}
\end{table}

The transparent rule yields 103 correct and 17 invalid adoptions; the MLP
yields 48 and 29. The paired correct-adoption difference (MLP minus simple) is
$-28.52$ percentage points, with simultaneous descriptive 95\% interval
[$-40.43,-17.38$]. The MLP rescues four correct adoptions and loses 59. Its
lower annotation-unobserved error does not compensate for the losses and higher
unique-visible and multiple-category errors: the prespecified gain gate,
simultaneous error-margin gate and conservative grouped bounds all fail.

Candidate supply is not the principal bottleneck on this subset. A correct
region appears in the frozen top five for 179/185 unique-visible cases
(96.8\%), but the learned head leaves 131 available correct candidates unused.
Thus the negative result localizes to candidate adoption rather than proposal
supply. It rejects a general advantage claim for this selected head; it is not
an official benchmark score, participant-independent evidence, a closed-loop
agent result, natural failure prevalence or a deployment/safety certification.

\section*{S39. Frozen embodied-policy admission: a negative result}
\label{suppidx:39}
\label{supp:arena-admission}
An additional clean-policy qualification used Isaac Lab-Arena
0.3.1~\cite{arena} with the GR00T-N1.6 DROID checkpoint~\cite{grootdroid}
and its native 32-action blocks. The plan fixed twenty seeds, interleaving four object--scene
configurations, a 70\,s simulated episode limit, official task predicates,
and a fresh seeded policy server per episode. It required at least fourteen
successful episodes; seven completed failures made that threshold unreachable
and triggered a prespecified futility stop. No TWS-V check, perturbation,
manual correction, or failed-episode retry was applied.

The run stopped after ten completed episodes: three succeeded and seven
failed. Ten planned episodes were not run and are not counted as failures;
even if all ten had succeeded, the maximum would have been thirteen.
Consequently, the policy did not pass admission, and no intervention-efficacy
comparison was started. A separate audit script verified the frozen file hashes,
seed order, one official outcome per completed episode, predicate records,
and the stopping decision. All seven unsuccessful episodes reached 1,050
steps (70\,s). The retained records showed no policy-server timeout or
out-of-memory termination. The manifest records process and video checks.
The stopped prefix documents the admission decision.

\begin{table}[!htbp]
\centering
\small
\begin{tabular}{@{}lrr@{}}
\toprule
Object--scene configuration & Completed & Successful\\
\midrule
Blue block / home office & 3 & 2\\
Orange / workshop & 3 & 0\\
Lemon / lounge & 2 & 1\\
Mustard bottle / interior & 2 & 0\\
\bottomrule
\end{tabular}
\caption{Completed episodes and native successes by fixed object--scene
configuration in the stopped GR00T-N1.6 DROID admission run.}
\label{tab:arena-admission-by-scene}
\end{table}
These descriptive counts are confounded with their fixed scene configurations
and do not support object-specific comparisons or selecting a favorable
subset for efficacy testing. The official placement predicate required one
satisfied step, not sustained placement stability.

\paragraph{Timing and transfer boundary.}
The pinned runner blocks on policy inference before advancing physics. Host
inference latency therefore cannot be interpreted as an equal interval of
unobserved simulated motion, or as a test of asynchronous world-action
execution. Simulator time and host time must be recorded separately. Buffered
actions refer to earlier observations, but expected robot motion alone does
not establish that an action is unsupported. Qualifying such opportunities
requires a separate, prespecified task-support criterion. Neither this
simulation admission result nor source-bound execution logs establish
physical safety or real-robot transfer.

\section*{S40. Frozen Human-1M natural-video mini-validation}
\label{suppidx:40}
\label{supp:human1m-minival}

\paragraph{Scope and frozen protocol.}
We use the NVIDIA PhysicalAI Human-1M-100K release~\cite{human1m} for an
exploratory natural-video component test, not a closed-loop task, safety,
prevalence or token-saving claim. Before decoding outcomes, 200 single-action
episodes were grouped by complete source-video container into 50 CAL and 150
TEST episodes. The storage-aware sample covers 81 video containers. A single AI
annotator labelled CAL and a frozen 100-episode TEST subset (one TEST episode
carries an earlier author label). The annotator knew the study design but saw no
policy output, feature, threshold or score; it viewed enlarged review sheets and
used SAM3 mask geometry as an aid to visual judgement. A frozen 30-episode
single-author spot-check then tested annotation reliability. Raw media remained
on the evaluation host. The manifest lists the frozen reference, audit
manifest, and evaluator identifiers.

The AI label file marks all 150 action texts as naming a visible source target.
Sixty-three rows lack a determinate cached-point reference, leaving 33
evaluable CAL and 54 evaluable TEST rows. Missing candidates or features remain
unknown rather than being discarded or mapped to a class. The 150 reviewed
post-event states include 128 visible, 18 ambiguous and four occluded cases.
These counts describe the storage-aware audit subset, not the Human-1M
population.

\paragraph{Frozen comparison.}
CAL selects a crop-ZNCC/RGB rule. At a matched CAL intervention count, the TWS-V
binding adds delivered head-pose displacement and an automatic hand--target
contact proxy; missing inputs force unknown. Always-refresh is a cost endpoint,
not an oracle. Thresholds are selected on CAL once and then applied to TEST.
Confidence intervals cluster-resample complete source-video containers.

\begin{table}[ht]
\centering
\begingroup
\scriptsize
\setlength{\tabcolsep}{2.2pt}
\begin{tabular}{@{}lccc@{}}
\toprule
Policy & \shortstack{False\\accept} & \shortstack{False\\reject} & Intervene\\
\midrule
Pixel ZNCC/RGB &5/37 (13.5\%)&11/17 (64.7\%)&43/54 (79.6\%)\\
TWS-V binding &5/37 (13.5\%)&11/17 (64.7\%)&43/54 (79.6\%)\\
Always refresh &0/37 (0\%)&17/17 (100\%)&54/54 (100\%)\\
\bottomrule
\end{tabular}
\caption{Frozen AI-labelled Human-1M TEST subset. For always-refresh,
``false reject'' denotes intervention on a valid cached point; it is a cost
quantity, not evidence that the replacement is correct. The rates are retained
for auditability but are not reliable performance estimates.}
\label{tab:human1m-minival}
\endgroup
\end{table}

The binding and pixel policies make identical TEST decisions: binding-minus-
pixel differences in false acceptance, false rejection and intervention are all
0.0, each with a video-cluster bootstrap 95\% interval of [0.0, 0.0]. The
prespecified primary success condition is therefore not established. More
importantly, the frozen author spot-check agrees with the AI on cached-point
validity in only 16/30 cases ($\kappa=.29$). A post-hoc relabel of the same 30
episodes under a clearer written rubric, by the author and a fresh AI annotator,
does not repair this boundary: agreement falls to 6/30 ($\kappa=0$), while the
author assigns all 30 cached points to the positive class; exact agreement across
all six recorded fields is only 4/30. Thus these AI labels are not reliable
enough for primary false-accept / false-reject claims; the table is preserved as
a diagnostic output only. CAL also contains only four valid cached
points among 33 evaluable rows, so its zero false-reject rate transfers to 11/17
false rejections on TEST. These observations diagnose annotation and
operating-point instability, not a natural-scene failure rate.

Refreshing also requires validation. Among the 100 source-admitted TEST rows, the
top post-event SAM3 proposal is the AI-labelled same target in 33, a different
ranked proposal is labelled correct in 25, and no same-identity proposal is
labelled in 42. Thus always obtaining a new proposal does not establish that a
repair succeeded. Under these unverified AI labels, the proposal counts are
consistent with, but do not establish, the need to validate replacements
separately; they provide no natural-scene advantage for the tested pose/contact
binding over the strong pixel rule. Two exact reruns match for the aggregate
summary, TEST decisions and CAL frontier hashes. An earlier run used the review
page export (\texttt{d9b85912}), in which 27 TEST rows carry the author's
spot-check or later edits instead of the frozen AI labels; it also gave identical
binding and pixel decisions.

\section*{S41. Offline FLUX 3 Action dispatch-contract integration}
\label{suppidx:41}
\label{supp:flux-action-contract}

We generate six 32-step action chunks with the frozen FLUX 3 Action~\cite{fluxaction}
DROID \texttt{sd-fp8r} checkpoint from two public Cosmos3-DROID~\cite{cosmosdroid} recordings,
using three fixed observation-selection seeds per recording.
Each step contains seven joint targets and a gripper value.
The model and recordings are identified by the
\href{https://github.com/black-forest-labs/flux-action/tree/aea7863fbbbda1e282fa7bde2054e290bf09b3f8}{official code revision},
\href{https://huggingface.co/black-forest-labs/flux-3-action-droid/tree/7305ef6d95de95ca4100c2e117c23ce580307335}{checkpoint revision}, and
\href{https://huggingface.co/datasets/nvidia/Cosmos3-DROID/tree/5c11a20accb11497270a5247a7f1e66ad04c956c}{dataset revision}.
A frozen local receipt binds each observation digest to its action digest.
The bypass and TWS-V conditions receive the same inputs in each paired replay.

\begin{table}[ht]
\centering
\small
\begin{tabular}{@{}lrr@{}}
\toprule
Controlled replay condition & Bypass & TWS-V \\
\midrule
Fresh, matching binding & 6/6 & 6/6 \\
Unchanged image, age over budget & 6/6 & 0/6 \\
Unchanged image, substituted chunk & 6/6 & 0/6 \\
Dispatch timestamp unknown & 6/6 & 0/6 \\
\bottomrule
\end{tabular}
\caption{Proposals allowed by each dispatch condition. The 24 pairs are
constructed from six observations nested within two source recordings.}
\label{tab:flux-action-contract}
\end{table}

Replay timestamps are injected with a fixed 30\,s engineering budget.
The substituted chunk is a real prediction from another observation in the
same recording. TWS-V reports the 12 over-budget or mismatched proposals
as invalid and the six missing-time proposals as unknown. The bypass applies
no contract checks. All pairs reuse the saved action outputs, including
when a chunk is substituted identically for both policies.
This diagnostic checks integration with a different action generator.
Visual target validity, replacement quality, and task completion require
separate evaluation. No robot motion is issued.

\section*{S42. Same-state closed-loop simulation with FLUX 3 Action}
\label{suppidx:42}
\label{supp:robolab-same-state}
We ran the RoboLab~\cite{robolab} DROID banana-in-bowl task port
with the frozen FLUX checkpoint from S41 on Isaac Lab-Arena~\cite{arena}. An ARM64 Jetson AGX Thor
served inference using a local SM110 compatibility patch and the NATTEN
Flex Attention fallback. The simulator applied 16 actions before each
new query. The clean development episode (seed 1127) succeeded at step
183. All 36 transmitted RGB views and 183 consumed action rows matched
their cross-host records. Median model inference was 4.70\,s per query,
and median observation-to-action-chunk readiness was 5.61\,s. Physics
paused during inference.

Figure~\ref{fig:robolab-same-state} shows the recorded exterior views and
post-shift wrist inputs for both matched interventions.

We reconstructed step 88 by replaying the exact clean prefix.
All four branches matched the pre-intervention RGB, recorded scene state
and pending actions. We then shifted the banana by the same amount in
every branch and advanced one common hold step to obtain new RGB.
The three arms that refreshed at the fork received the same replacement
proposal for the same frames and inference seed.

\begin{table}[htbp]
\centering\footnotesize
\begin{tabular}{@{}lrr@{}}
\toprule
Policy at the fork & Outcome & Steps \\
\midrule
Reuse, then scheduled replanning & Success & 196 \\
Unconditional refresh at the fork & Success & 224 \\
Time and local RGB rule & Success & 224 \\
TWS-V with exterior-template recheck & Timeout & 1,050 \\
\bottomrule
\end{tabular}
\caption{Matched intervention branches from one development seed.
Steps are measured from episode start. The TWS-V branch abstained at
the fork and held until timeout.}
\label{tab:robolab-same-state}
\end{table}

The old-location normalized cross-correlation fell from 0.940 before
the shift to 0.087 after it. The refreshed proposal passed timing
and content-binding checks, but the best exterior-template match
was 0.519 against the development threshold of 0.75. The visual
recheck therefore withheld the replacement executed by the two
refresh controls. Scheduled replanning recovered in the reuse arm.
The case isolates a visual replacement rejection and provides no
estimate of rescue frequency. After observing this rejection, we added a
wrist-RGB presence check with a minimum connected-region area of
1,000 pixels and retested the same development scene. The new check
admitted the identical replacement. The initial retest ended after a
transport error. A repeat completed at step 224 with the same dispatched
actions as the unconditional-refresh control. This post-hoc result is
specific to the development scene.

\paragraph{Pre-release destination displacement (development).}
At step 177 of the same development seed, we restored the common state
immediately before the cached opening command and shifted the bowl by a
measured 13.94\,cm before dispatch. All four branches succeeded by the
native task predicate. Scheduled reuse completed at step 267 after releasing
14.57\,cm from the shifted bowl center and recovering through later replanning.
Unconditional refresh, the time-and-vision rule, and TWS-V dispatched identical
actions and completed at step 219, 48 steps earlier than scheduled reuse in
this development episode.

\noindent\begin{minipage}{\columnwidth}
\section*{Reproducibility manifest}
Table~\ref{tab:reproducibility-manifest} summarizes the pinned provenance and
process receipts.

\medskip
\centering
\scriptsize
\begin{tabular}{@{}lp{.74\linewidth}@{}}
\toprule
Section & \raggedright Pinned artifact or audit receipt \tabularnewline
\midrule
\suppindex{1} & \raggedright OpenCV Zoo VitTrack ONNX payload, 714,726 bytes, at
the revision identified in the model citation~\cite{vittrack}.\tabularnewline
\suppindex{28} & \raggedright OSWorld 2.1~\cite{osworldrelease}, 108 task
sources matched to the release manifest and 1,084 assets verified,
4,920,057,674 bytes.\tabularnewline
\suppindex{29} & \raggedright Qwen3-VL-4B-Instruct snapshot
\texttt{ebb281ec70b05090aa6165b016eac8ec08e71b17}; 13 pinned files,
8,887,291,213 bytes.\tabularnewline
\suppindex{30} & \raggedright HarnessRouter/UHP task protocol 2026-10-19,
pinned commit \texttt{f47d736d01d0bbd4b9489edd6615038d733eeef3}.\tabularnewline
\suppindex{38} & \raggedright Transport recount verified 432 image paths and
per-file hashes, 163,416,828 bytes. The restart reused only 43 files that
matched size and SHA-256.\tabularnewline
\suppindex{39} & \raggedright The simulator completed normally for all ten
completed attempts; the policy server returned exit code 143 on explicit
cleanup. Thirty camera videos decoded successfully.\tabularnewline
\suppindex{40} & \raggedright SHA-256 prefixes for the frozen reference,
manifest, and evaluator, respectively: \texttt{7b54c5c5}, \texttt{8c9cd61f},
\texttt{0cb5765e}.\tabularnewline
\bottomrule
\end{tabular}
\captionsetup{hypcap=false}
\captionof{table}{Selected provenance and infrastructure receipts. Each section
describes its experimental design, denominators, and outcome checks.}
\label{tab:reproducibility-manifest}
\end{minipage}\par

\end{document}